\documentclass[11pt]{article} 
\usepackage{times} 
\usepackage[margin=1in]{geometry} 
\usepackage{algorithm}
\usepackage{algorithmic}
\usepackage[svgnames]{xcolor}

\usepackage{amsmath,amsfonts,bm}

\def\eqref#1{equation~\ref{#1}}

\def\1{\bm{1}}

\DeclareMathAlphabet{\mathsfit}{\encodingdefault}{\sfdefault}{m}{sl}
\SetMathAlphabet{\mathsfit}{bold}{\encodingdefault}{\sfdefault}{bx}{n}

\usepackage{graphicx}
\usepackage{subcaption}
\usepackage{hyperref}
\usepackage{url}
\usepackage{booktabs}
\usepackage{array}
\usepackage{graphicx} 
\usepackage{wrapfig}  
\usepackage{natbib}
\usepackage{float}
\newcolumntype{C}[1]{>{\centering\arraybackslash}p{#1}}

\title{SAC Flow: Sample-Efficient Reinforcement Learning of Flow-Based Policies via Velocity-Reparameterized Sequential Modeling}

\author{
    Yixian Zhang\textsuperscript{1$\ast$}, 
    Shu'ang Yu\textsuperscript{1, 4$\ast$}, 
    Tonghe Zhang\textsuperscript{2}, 
    Mo Guang\textsuperscript{3}, 
    Haojia Hui\textsuperscript{3}, 
    Kaiwen Long\textsuperscript{3}, \\
    Yu Wang\textsuperscript{1}, 
    Chao Yu\textsuperscript{1$\dagger$}, 
    Wenbo Ding\textsuperscript{1$\dagger$} \\[1em]
    \textsuperscript{1}Tsinghua University \quad 
    \textsuperscript{2}Carnegie Mellon University \quad 
    \textsuperscript{3}Li Auto \\
    \textsuperscript{4}Shanghai AI Laboratory \\[0.5em]
    \textsuperscript{$\ast$}Equal contribution \quad
    \textsuperscript{$\dagger$}Corresponding Authors \\
    \texttt{\{zoeyuchao@gmail.com, ding.wenbo@sz.tsinghua.edu.cn\}}
}
\date{}
\begin{document}

\maketitle
\begin{figure}[H]
    \centering
    \includegraphics[width=1.0\linewidth]{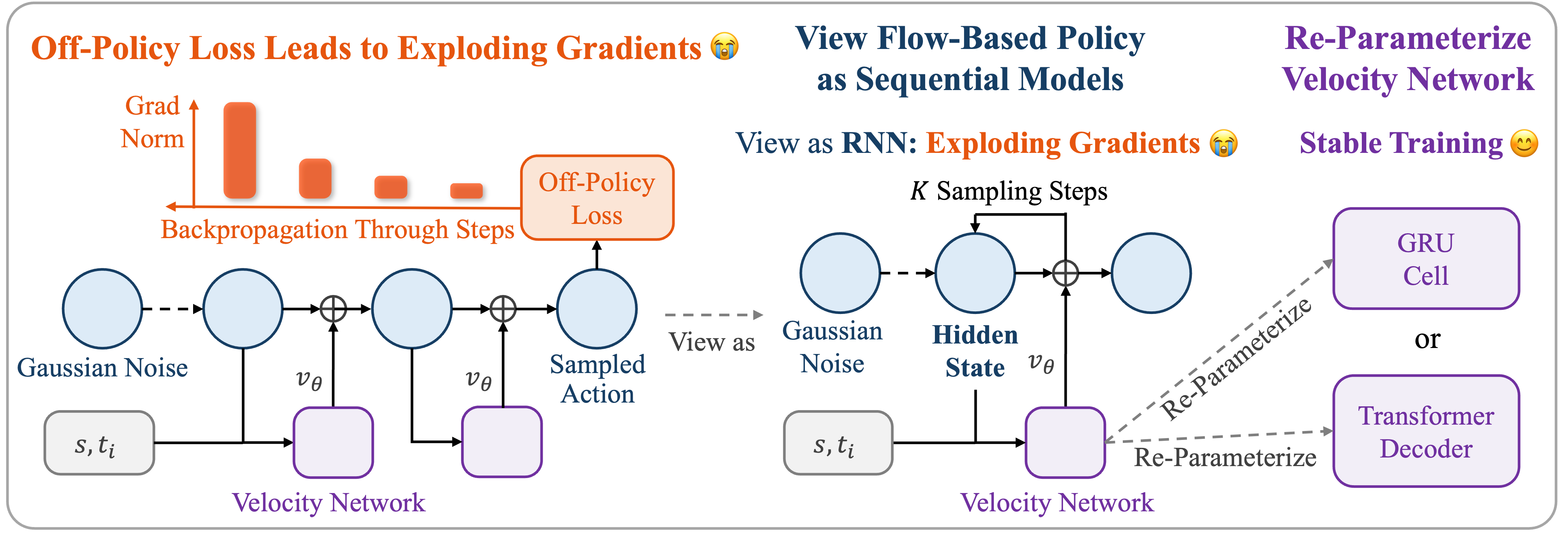}
    \caption{An Overview of SAC Flow. The multi-step sampling process of flow-based policies frequently causes exploding gradients during off-policy RL updates. Our key insight is to treat the flow-based policy as a sequential model, for which we first demonstrate an algebraic equivalence to an RNN. We then reparameterize the flow's velocity network using modern sequential architectures (e.g., GRU, Transformer). Our approach stabilizes off-policy RL training and achieves state-of-the-art performance.}
    \label{fig:overview}
\end{figure}
\begin{abstract}
Training expressive flow-based policies with off-policy reinforcement learning is notoriously unstable due to gradient pathologies in the multi-step action sampling process. We trace this instability to a fundamental connection: the flow rollout is algebraically equivalent to a residual recurrent computation, making it susceptible to the same vanishing and exploding gradients as RNNs. To address this, we reparameterize the velocity network using principles from modern sequential models, introducing two stable architectures: Flow-G, which incorporates a gated velocity, and Flow-T, which utilizes a decoded velocity. We then develop a practical SAC-based algorithm, enabled by a noise-augmented rollout, that facilitates direct end-to-end training of these policies. Our approach supports both from-scratch and offline-to-online learning and achieves state-of-the-art performance on continuous control and robotic manipulation benchmarks, eliminating the need for common workarounds like policy distillation or surrogate objectives. Code is available at \url{https://github.com/Elessar123/SAC-FLOW}.
\end{abstract}.

\section{Introduction}

Flow-based policies have shown strong potential on challenging continuous-control tasks, including robot manipulation, due to their ability to represent rich, multimodal action distributions \citep{pi0,Flow1,flow_robot1}. Early successes predominantly arose in imitation learning, where a flow-based policy is trained to reproduce expert behavior from static datasets \citep{flow_robot2,flow_robot3}. However, pure behavior cloning is fundamentally limited: dataset coverage is often sparse and of mixed quality \citep{IM_problem1,IM_problem2}, and the lack of environment interaction prevents exploration, making it difficult to exceed demonstrator performance on hard tasks \citep{IM_problem3,IM_problem4}.

A natural next step is to train flow-based policies with reinforcement learning. On-policy variants of PPO adapted to flows have demonstrated strong returns, yet they remain sample-inefficient \citep{PPO,ReinFlow}. Off-policy methods promise much higher data efficiency and early integrations with flow-based policies on MuJoCo and DeepMind Control show encouraging results \citep{Mujoco,DM_control,FlowRL,FlowQ}. However, these successes typically come with design compromises that leave a central issue unresolved. Either the update relies on surrogate objectives that avoid differentiating through the rollout of the original flow, or the flow is distilled into a simpler one-step actor that can be optimized with standard off-policy losses. Both strategies reduce gradient stress but decouple optimization from the expressive generator and tend to blunt the benefits of multimodal flow-based policies \citep{FlowQ,FlowRL}.

We propose a different viewpoint: treat the flow-based policy as a sequential model. Concretely, we show that the Euler integration used to generate actions in the flow-based policy is algebraically identical to the recurrent computation of a residual RNN. This observation explains the instability observed with off-policy training: the same vanishing or exploding gradients known to affect RNNs also afflict the flow rollout. Building on this link, we reparameterize the vanilla velocity network with the cell of modern sequential models that are designed to stabilize deep recurrent computations. We introduce two such novel designs of the flow-based policy: Flow-G, which incorporates a GRU-style gated velocity to regulate gradient flow across rollout steps, and Flow-T, which utilizes a Transformer-style decoded velocity to refine the action-time token via state-only cross-attention and a residual feed-forward network.

Our main contributions are summarized as follows:
\begin{itemize}
    \item \textbf{A sequential model perspective for stable flow-based policies.} We formalize the $K$-step flow rollout as a residual RNN computation, providing a clear theoretical explanation for the gradient pathologies that cause instability in off-policy training. This insight allows us to reparameterize the velocity network with modern sequential architectures, leading to two novel, stable designs: \textbf{Flow-G} (GRU-gated) and \textbf{Flow-T} (Transformer-decoded). Our approach resolves critical gradient pathologies, enabling direct end-to-end optimization and eliminating the need for surrogate objectives or policy distillation.

    \item \textbf{A practical and sample-efficient SAC framework for flow policies.} We develop SAC Flow, a robust off-policy algorithm built upon our stabilized architectures. By introducing a noise-augmented rollout, we enable tractable likelihood computation for the SAC objective, a key technical hurdle. This approach yields two robust training procedures: (i) a stable from-scratch trainer for dense-reward tasks and (ii) a unified offline-to-online pipeline for sparse-reward tasks.
    
    \item \textbf{State-of-the-art sample efficiency and performance.} Our proposed methods, SAC Flow-G and SAC Flow-T, establish new state-of-the-art results across a comprehensive suite of benchmarks. In from-scratch training on challenging MuJoCo tasks, our approach delivers performance gains of up to $130\%$ over strong baselines. Furthermore, in complex offline-to-online manipulation tasks on OGBench, it achieves up to a $60\%$ higher success rate. These results empirically validate the superior \textbf{sample efficiency} of our direct off-policy training approach, with ablation studies further confirming the robustness of our designs.
\end{itemize}

\section{Preliminaries}
\label{Preliminaries}

\subsection{Reinforcement Learning}

We consider policy optimization in an infinite-horizon Markov decision process $\langle\mathcal{S}, \mathcal{A}, p, r, \rho\rangle$ with continuous state and action spaces. The transition function $p: \mathcal{S} \times \mathcal{A} \times \mathcal{S} \rightarrow [0, \infty)$ specifies the transition probability density, and rewards are $r_h = r(s_h, a_h) \in [r_{\min}, r_{\max}]$, where $a_h$ is sampled from the policy $\pi(\cdot|s_h)$. The objective of reinforcement learning is to learn an optimal policy $\pi^{*}$ that maximizes the expected cumulative reward, $\pi^{*}=\arg \max _\pi \mathbb{E}^{\pi} \left[\sum_{h=0}^{\infty} \gamma^hr_h\right]$.

\subsection{Soft Actor-Critic algorithm}

To encourage policies to maintain stochasticity and explore more effectively, the standard objective is augmented with an entropy term, $\hat{J}(\pi)=\mathbb{E}^\pi[\sum_{h=0}^{\infty} \gamma^h (r_h+\alpha\mathcal{H})]$, where $\mathcal{H}(\pi(\cdot \mid s_h)) = - \mathbb{E}_{a \sim \pi(\cdot \mid s_h)} \left[ \log \pi(a \mid s) \right]$ denotes the state-conditional policy entropy. In this setting, the Soft Actor-Critic algorithm \citep{SAC_arxiv} is introduced to optimize this objective. The target $\hat{J}(\pi)$ is typically approximated with the soft $Q$-function $Q_{\psi}(s_h, a_h)$, which is updated through the TD loss:
\begin{equation}
\label{eq:Critic_loss}
    L(\psi)=\left[Q_\psi\left(s_h, a_h\right)-(r_h+\gamma Q_{\bar{\psi}}\left(s_{h+1}, a_{h+1}\right)-\alpha \log \pi_\theta\left(a_{h+1} \mid s_{h+1}\right))\right]^2, 
\end{equation}
where $a_{h+1} \sim \pi_\theta\left(\cdot \mid s_{h+1}\right)$, $(s_h, a_h, s_{h+1}, r_h)$ are sampled from the replay buffer, and $\bar{\psi}$ is a delayed copy of $\psi$ through which gradients do not flow for stability. To maximize the soft $Q$-function $Q_{\psi}(s_h, a_h)$, the policy $\pi_\theta$ is updated through
\begin{equation}
\label{eq:SAC_Actor}
    L(\theta)=\alpha \log \pi_\theta\left(a^\theta_h \mid s_h\right)-Q_\psi\left(s_h, a^\theta_h\right), \quad a^\theta_{h} \sim \pi_\theta\left(\cdot \mid s_{h}\right).
\end{equation}
Here, $a^\theta_h$ highlights a reparameterized action sample that allows gradients to propagate from the policy to the action, in contrast to the TD update, where the action is detached.

\subsection{Flow-based Policy in Reinforcement Learning}

Gaussian policies are the standard choice in continuous-control RL \citep{Gaussian_AAAI,Gaussian_nips}, yet a single unimodal Gaussian cannot capture inherently multimodal action distributions, a limitation that is especially harmful in long-horizon robotic control. Diffusion policies alleviate this by modeling arbitrary normalizable distributions and have achieved state-of-the-art results on manipulation benchmarks \citep{Diffusion1,Diffusion2,DPPO}, but their iterative denoising makes both training and inference expensive. Recently, flow-based policies have emerged as a simpler alternative: trained with flow-matching objectives, they offer easier training and faster inference while often matching or exceeding diffusion quality \citep{Flow1,FlowQ,ReinFlow}.

A flow-based policy transports a simple, state-conditioned base $p_0(\cdot\mid s)$ over the action space $\mathcal{A}=\mathbb{R}^d$ to a target policy $p_1(\cdot\mid s)$ via a time-indexed map $\varrho:[0,1]\!\times\!\mathcal{A}\!\times\!\mathcal{S}\rightarrow\mathcal{A}$, with $A_t:=\varrho_t(A_0\mid s)$ for $t\in[0,1]$, where $A_0\!\sim\!p_0(\cdot\mid s)$ and $A_1\!\sim\!p_1(\cdot\mid s)$. The trajectory satisfies the ODE $\tfrac{\mathrm{d}}{\mathrm{d}t}\varrho_t(A_0\mid s)=v\!\left(t,\varrho_t(A_0\mid s),s\right)$, where $v$ is a learnable velocity field. We adopt Rectified Flow \citep{ReFlow}, which uses the straight path $A_t=(1-t)A_0+tA_1$ and the standard Gaussian base $p_0(\cdot\mid s)=\mathcal{N}(0,I_d)$. In this case $v(t,A_t,s)=\tfrac{\mathrm{d}}{\mathrm{d}t}A_t=A_1-A_0$, yielding the flow-matching objective
\begin{equation}
\label{flow_matching}
\hat{\theta}
=\arg\min_{\theta}\;
\mathbb{E}_{\substack{A_0\sim \mathcal{N}(0,I_d),\; (A_1=a,s)\sim \mathcal{D},\\ t\sim \operatorname{Unif}[0,1]}}
\!\Big[\big\|A_1-A_0-v_\theta\!\left(t,(1-t)A_0+tA_1,s\right)\big\|_2^2\Big],
\end{equation}
where $\mathcal{D}$ denotes the dataset of state–action pairs. In inference, the learned field is integrated numerically with flow rollout to obtain:
\begin{equation}
\label{FR}
    A_{t_{i+1}}=A_{t_i}+\Delta t_i\, v_\theta(t_i,A_{t_i},s), \quad 0=t_0<\cdots<t_K=1,
\end{equation}
where $\Delta t_i=t_{i+1}-t_i$. The resulting distribution over $A_1$ induced by $A_0\!\sim\!\mathcal{N}(0,I_d)$ is denoted $\mu_\theta(\cdot\mid s)$ and serves as the stochastic policy $a = A_1 \sim \pi_\theta(\cdot\mid s)$.

Flow-based policies can be trained offline from demonstrations using Equ. (\ref{flow_matching}), and they can also be optimized with RL. \textbf{On-policy} methods (e.g., PPO-style training tailored to flows \citep{ReinFlow,DPPO,QSM}) attain strong performance on challenging robotics tasks but remain sample-inefficient. \textbf{Off-policy} methods (e.g., SAC, TD3) are highly sample-efficient \citep{off_policy1}, yet directly backpropagating through the $K$-step action sampling is often unstable, especially for large $K$ \citep{FlowQ}. To mitigate this, prior work either distills a flow-based policy into a simpler actor trained with standard off-policy losses \citep{FlowQ} or proposes surrogate off-policy objectives that train the velocity field without differentiating through the full flow rollout \citep{FlowRL}.

We take a different route. We recast the flow rollout as a sequential model and redesign the velocity parameterization accordingly. We introduce Flow-G, which uses a GRU-style gated velocity, and Flow-T, which uses a Transformer-style decoded velocity. These parameterizations stabilize the $K$-step backpropagating and allow direct off-policy training of the flow-based policy. We instantiate the framework with SAC, and the same formulation applies to other off-policy algorithms.

\section{From Flow Rollout to sequential models}
\label{sac:Link_f_q}
\begin{figure}[htbp]
    \centering
    \includegraphics[width=0.75\linewidth]{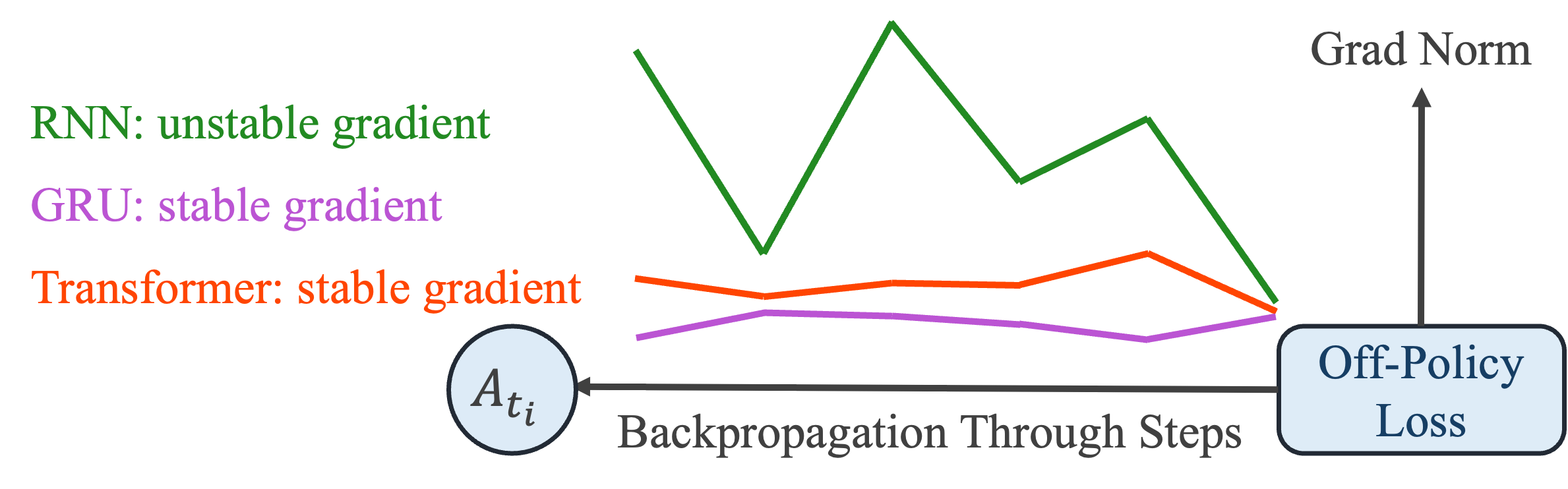}
    \caption{An illustration of gradient norms during training. By conceptualizing a flow-based model as an RNN, the most basic sequential models, we observe that it still suffers from the exploding gradients during training. This motivates our work to model the flow-based model as advanced sequential architectures, such as a GRU or a Transformer. These models can be updated with stable gradients during the backpropagation process.}
    \label{fig:grad}
\end{figure}

In this section, we reveal a key insight: flow-based policies are fundamentally sequential models. As conceptually illustrated in Fig. \ref{fig:grad}, standard flow rollouts exhibit gradient instabilities analogous to vanilla RNNs, while modern sequential architectures offer more stable gradient flow, motivating our velocity network designs.

\begin{figure}[tbp]
  \centering 

  \subfloat[\centering Flow-based Policy as RNN]{%
    \includegraphics[width=0.264\textwidth]{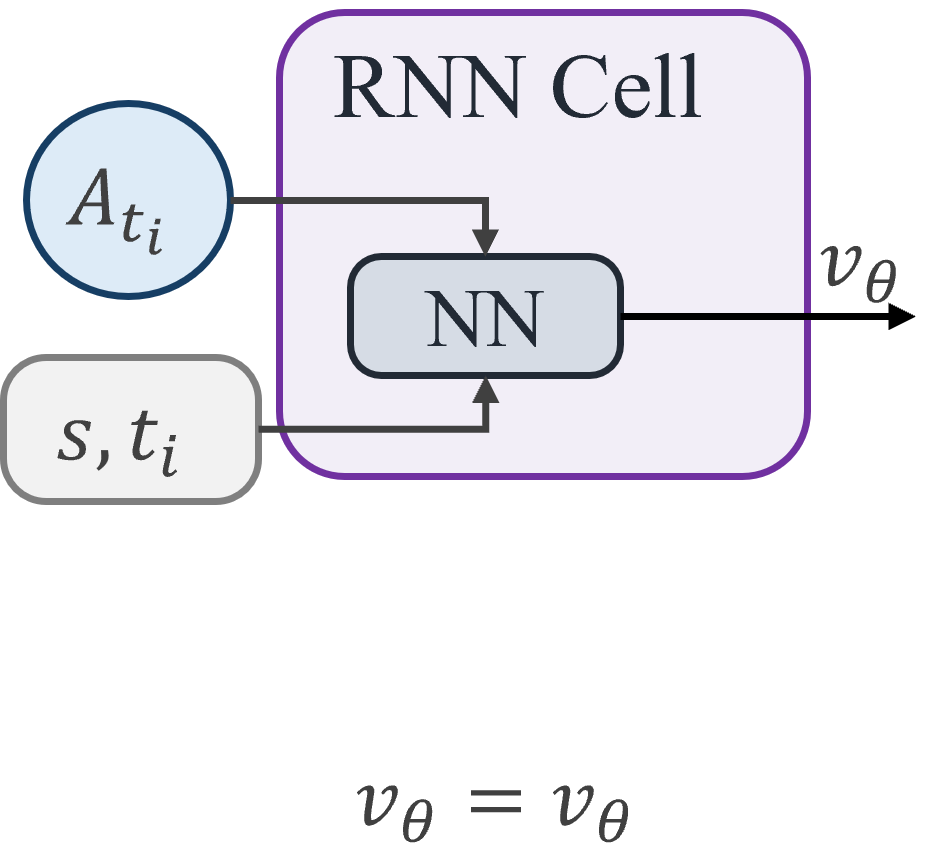}%
    \label{fig:RNN} 
  }
  \subfloat[\centering Flow-based Policy as GRU]{%
    \includegraphics[width=0.343\textwidth]{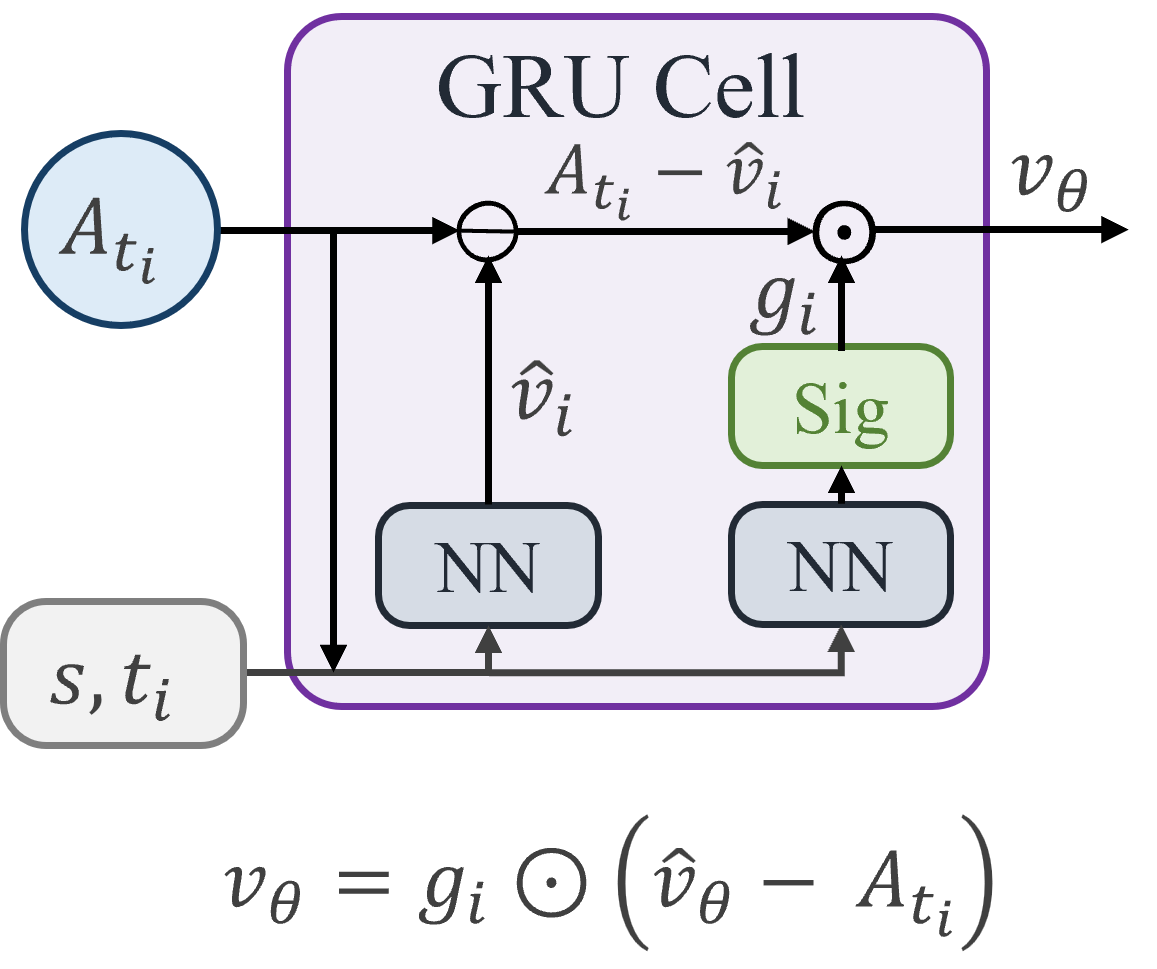}%
    \label{fig:GRU} 
  }
  \subfloat[\centering Flow-based Policy as Transformer]{%
    \includegraphics[width=0.313\textwidth]{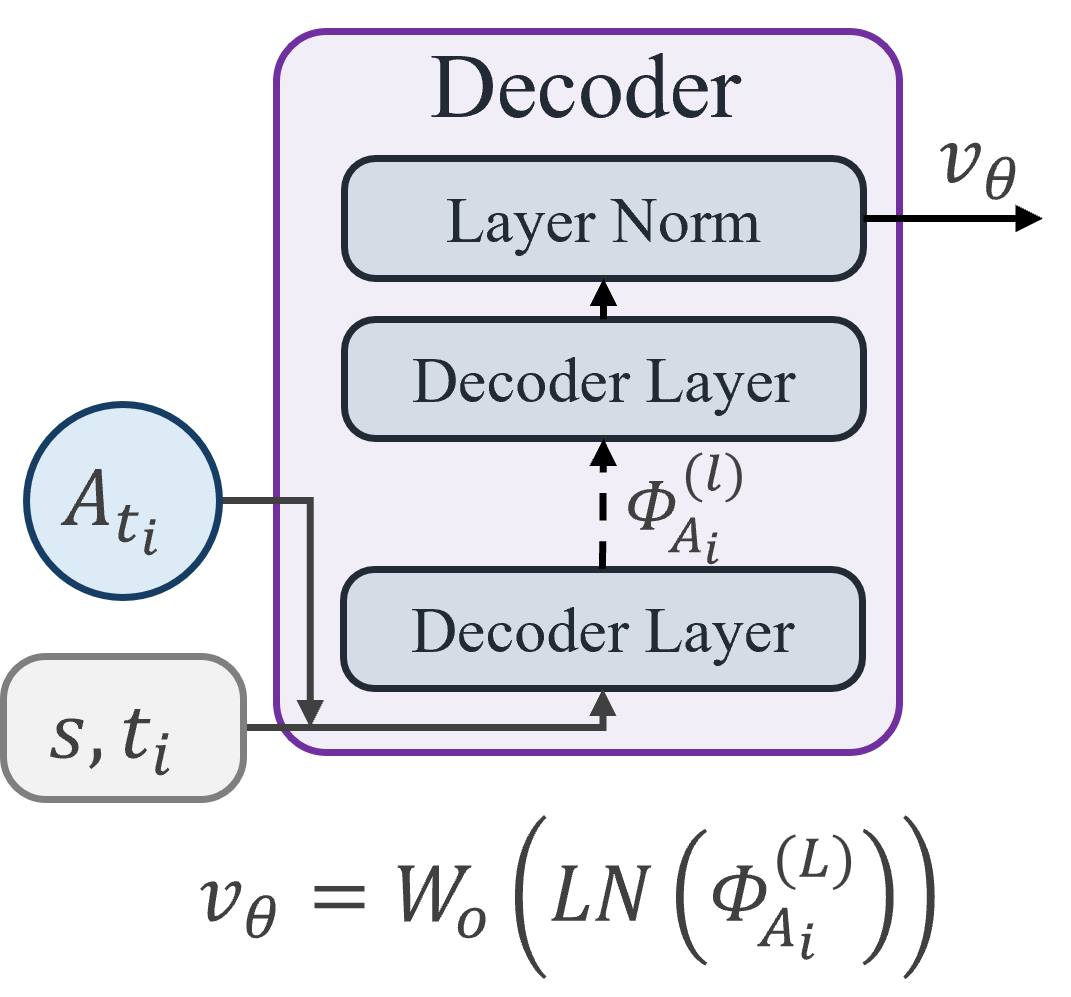}%
    \label{fig:TF} 
  }

\caption{
    Velocity network parameterizations for the flow-based policy, shown in the view of sequential models.
    (a) \textbf{RNN Cell}: It represents the standard flow-based policy where the velocity $v_{\theta}$ is the direct output of a neural network. This simple formulation is prone to gradient instability.
    (b) \textbf{GRU Cell}: The velocity is computed using a GRU-style gated mechanism. A gate $g_i$ adaptively controls the update strength from a candidate network $\hat{v}_i$, which stabilizes gradient flow.
    (c) \textbf{Decoder}: The velocity is modeled using a Transformer decoder, where the action-time token $A_{t_i}$ is refined through $L$ layers of state-conditioned cross-attention to produce a decoded velocity.
}
  \label{fig:detail_nn} 
\end{figure}

\paragraph{Flow-based policy as RNN (Fig. \ref{fig:RNN}).}
Treat the intermediate action $A_{t_i}$ as the hidden state and $(t_i,s)$ as the input. Then Equ. (\ref{FR}) is a residual RNN step \citep{RNN_Res}:
\begin{equation}
\label{eq:rnn_residual}
A_{t_{i+1}} \;=\; A_{t_i} \;+\; f_\theta(t_i, A_{t_i}, s),
\quad \text{with } f_\theta (\cdot) = \Delta t_i\, v_\theta (\cdot),
\end{equation}
where $f_\theta (\cdot)$ denotes the RNN cell. Consequently, training a flow-based policy with off-policy losses backpropagates through a deep recurrent stack of $K$ updates in RNN, which is prone to gradient explosion and vanishing. This explains the instability observed when naively applying off-policy reinforcement learning to standard flow-based policies.

\paragraph{Flow-based policy as GRU (Flow-G, Fig. \ref{fig:GRU}).}
To improve gradient stability, we endow the velocity with a GRU-style update gate. Let $g_i={Sig}\!\big(z_\theta(t_i,A_{t_i},s)\big)$ and let $\hat{v}_\theta$ be a candidate network. Define
\begin{equation}
\label{eq:gru_flow}
A_{t_{i+1}} \;=\; A_{t_i} \;+\; \Delta t_i \,\big( g_i \odot \left(\hat{v}_\theta(t_i, A_{t_i}, s) - A_{t_i} \right)\big),
\end{equation}
where $\odot$ denotes elementwise multiplication and $Sig(\cdot)$ is the logistic sigmoid. Equ. (\ref{eq:gru_flow}) is exactly a flow sampling step with gated velocity $v_\theta = g_i \odot \left(\hat{v}_\theta - A_{t_i}\right)$, which mirrors the structure of the update in a GRU cell but expressed in the velocity parameterization used by the flow rollout. The gate network $g_i$ adaptively interpolates between keeping the current intermediate action and forming a new one.

\paragraph{Flow-based policy as Transformer (Flow-T, Fig. \ref{fig:TF}).}
We parameterize the velocity function $v_\theta$ using a Transformer architecture conditioned on the environment state $s$. To maintain the Markov property of the flow, we depart from a traditional causal, autoregressive formulation. Instead, the model first computes independent embeddings for the current action-time token $A_{t_i}$ and a single, global embedding for the state $s$:
\begin{equation}
\label{eq:tf_embeddings}
\Phi_{A_i} = E_A\big(\phi_t(t_i), A_{t_i}\big), \qquad \Phi_S = E_S\big(\phi_s(s)\big),
\end{equation}
where $E_A$ and $E_S$ are linear projections. Within the Decoder layers, a diagonal mask is applied to the self-attention mechanism, effectively reducing it to a position-wise transformation that processes each token $\Phi_{A_i}$ independently, without mixing information across the time steps $i$. The crucial step for context integration is a dedicated cross-attention module, where each action token $\Phi_{A_i}$ queries the shared state embedding $\Phi_S$. A stack of $L$ pre-norm residual blocks refines the action tokens:
\begin{equation}
\label{eq:tf_cross_attn}
Y_i^{(l)} = \Phi_{A_i}^{(l-1)} + \text{Cross}_l\Big(\mathrm{LN}(\Phi_{A_i}^{(l-1)}),\text{context}=\mathrm{LN}(\Phi_S)\Big), \, \Phi_{A_i}^{(l)} = Y_i^{(l)} + \mathrm{FFN}_l\big(\mathrm{LN}(Y_i^{(l)})\big),
\end{equation}
for layers $l=1,\dots,L$, where $\Phi_{A_i}^{(0)} := \Phi_{A_i}$. Each block is completed by a feed-forward network, and the final representation is projected to the velocity space:
\begin{equation}
\label{eq:tf_ffn_velocity}
A_{t_{i+1}} \;=\; A_{t_i} \;+\; \Delta t_i \,W_o\Big(\mathrm{LN}(\Phi_{A_i}^{(L)})\Big),
\end{equation}

where $W_o$ is a linear projection and $v_\theta(t_i,A_{t_i},s) = W_o\Big(\mathrm{LN}(\Phi_{A_i}^{(L)})\Big)$ is the decoded velocity in Flow-T. Each velocity evaluation therefore executes $L$ layers that refine the current action token $\Phi_{A_i}$ based on the global state context from $\Phi_S$, not on a causal history of other tokens. This state-conditioned refinement of the entire trajectory maintains the fundamental Markov property of flow-based policy while enabling stable integration with off-policy learning algorithms.

\paragraph{Takeaway for off-policy reinforcement learning.}
 Equ. (\ref{eq:rnn_residual}) establishes that a standard flow rollout is a residual recurrent computation. Introducing a gate network leads to Flow-G in Equ. (\ref{eq:gru_flow}) and improves gradient stability. Replacing the velocity with the normalized residual block in Equ. (\ref{eq:tf_ffn_velocity}) yields Flow-T. This architecture provides well-conditioned depth and, crucially, aggregates context with the well-established Transformer architectures. These parameterizations serve as drop-in replacements for $v_\theta$ in Equ. (\ref{FR}) without altering the surrounding algorithm. As a result, they enable direct and stable off-policy training with methods such as SAC, remove the need for auxiliary distillation actors and surrogate objectives, and keep flow rollout efficient at test time.

\section{Training Flow-based Policy via SAC}
\label{sec:sac_flow_compact}

With gradient stability achieved through our sequential parameterizations (Flow-G and Flow-T), we can now train flow-based policies directly with off-policy reinforcement learning. The key technical challenge is computing policy likelihoods for the
K-step rollout in Equ. (\ref{FR})—a requirement for the entropy-regularized objective in SAC. We solve this through a principled noise-augmented rollout that preserves the final action distribution while enabling tractable per-step likelihood computation.

\paragraph{Likelihood via a noise-augmented rollout.}
SAC requires explicit policy likelihoods for entropy regularization, but the deterministic $K$-step rollout in Equ. (\ref{FR}) yields intractable densities. We address this by making the rollout stochastic while preserving the marginal of the final action, which induces a product of per-step Gaussian transitions and a tractable joint path density $p_c(\mathcal{A}\mid s)$ over intermediate actions $\mathcal{A}=(A_{t_0},\ldots, A_{t_K})$. The construction details are deferred to Appendix~\ref{SAC_loss_flow}; here we use the resulting $\log p_c(\mathcal{A}\mid s)$ as a drop-in entropy term.

\paragraph{From-scratch training.}
With tractable likelihoods established, the SAC losses become straightforward. Given a critic $Q_\psi$ and a flow-based policy $\pi_\theta$ (with Flow-G or Flow-T as $v_\theta$), we optimize:
\begin{equation}
\label{actor_main}
L_{\text{actor}}(\theta)
= \alpha\, \log p_c(\mathcal{A}^\theta \mid s_h) \;-\; Q_\psi\!\big(s_h, a_h^\theta\big), 
\quad \mathcal{A}^\theta \sim \pi_\theta(\cdot\mid s_h),\;\; a_h^\theta=\tanh(A_{t_K}^\theta),
\end{equation}
\begin{equation}
\label{critic_main}
L_{\text{critic}}(\psi)
= \Big[\,Q_\psi(s_h,a_h) - \big(r_h + \gamma\, Q_{\bar\psi}(s_{h+1},a_{h+1}) - \alpha\, \log p_c(\mathcal{A}_{h+1}\mid s_{h+1})\big)\Big]^2,
\end{equation}
where $(s_h,a_h,r_h,s_{h+1})$ comes from the replay buffer, $\mathcal{A}_{h+1}, a_{h+1}\sim\pi_\theta(\cdot\mid s_{h+1})$, and $\bar\psi$ is a delayed copy.

\paragraph{Offline-to-online training.}
For sparse-reward tasks where expert demonstrations are available, we modify the actor loss to include a proximity regularizer:
\begin{equation}
\label{actor_o2o}
L_{\text{actor}}^{o}(\theta)
= \alpha\, \log p_c(\mathcal{A}^\theta \mid s_h)
\;-\; Q_\psi(s_h,a_h^\theta)
\;+\; \beta\, \|a_h^\theta - a_h\|_2^2,
\quad (s_h,a_h)\sim \mathcal{B}.
\end{equation}
This approach begins with flow-matching pretraining on expert data via Equ. (\ref{flow_matching}), then transitions to online learning while maintaining proximity to the replay buffer. The complete procedures are summarized in Algos.~\ref{alg:flow_sac_fs} and~\ref{alg:flow_sac_o2o}.

\begin{algorithm}
\caption{SAC Flow (from scratch)}
\label{alg:flow_sac_fs}
\begin{algorithmic}[1]
\STATE Initialize critic $Q_\psi$, target $Q_{\bar\psi}$, flow-based policy $\pi_\theta$ with Flow-G or Flow-T; replay buffer $\mathcal{B}$.
\FOR{each update}
  \STATE Interact with the environment using $\pi_\theta$; push $(s_t,a_t,r_t,s_{t+1})$ to $\mathcal{B}$.
  \STATE Sample $\{(s_h,a_h,r_h,s_{h+1})\}_{h=1}^N \sim \mathcal{B}$.
  \STATE Actor: draw $a_h^\theta$ by a $K$-step noisy rollout; minimize Equ. (\ref{actor_main}).
  \STATE Critic: minimize Equ. (\ref{critic_main}); update target by an exponential moving average.
\ENDFOR
\end{algorithmic}
\end{algorithm}

\begin{algorithm}
\caption{SAC Flow (offline-to-online)}
\label{alg:flow_sac_o2o}
\begin{algorithmic}[1]
\STATE Initialize $Q_\psi$, $Q_{\bar\psi}$, $\pi_\theta$; set $\mathcal{B} \leftarrow \mathcal{D}_{\text{expert}}$.
\FOR{$\ell=1$ to $L_{\text{off}} + L_{\text{on}}$}
  \IF{$\ell > L_{\text{off}}$} \STATE Interact with the environment using $\pi_\theta$; append to $\mathcal{B}$. \ENDIF
  \STATE Sample $\{(s_h,a_h,r_h,s_{h+1})\}_{h=1}^N \sim \mathcal{B}$.
  \STATE Actor: minimize Equ. (\ref{actor_o2o}) with $a_h^\theta$ from the noisy rollout.
  \STATE Critic: minimize Equ. (\ref{critic_main}); update the target network.
  \IF{$\ell \le L_{\text{off}}$} \STATE Flow-matching pretraining via Equ. (\ref{flow_matching}). \ENDIF
\ENDFOR
\end{algorithmic}
\end{algorithm}

For clarity, we refer to our methods as SAC Flow-G and SAC Flow-T, corresponding to training our Flow-G and Flow-T architectures with SAC. These terms apply to both our from-scratch and offline-to-online training procedures. Additionally, we extend our framework to fine-tuning, where Flow-G and Flow-T are reformulated as lightweight adapters to stabilize the online fine-tuning of an arbitrary pre-trained flow-based policy. This adapter-based fine-tuning algorithm is detailed in Appendix \ref{fine-tuning}.

\section{Experiment}
\label{sec:exp}
We conduct extensive experiments on locomotion and manipulation benchmarks to validate our approach. The evaluation encompasses: (1) experimental setup and baseline comparisons for from-scratch and offline-to-online training, (2) performance benchmarking of SAC Flow-G and SAC Flow-T against recent methods, and (3) ablation studies analyzing the effectiveness of our design components. All results are averaged over 5 random seeds and use the 95\% confidence interval.

\subsection{Settings}
\subsubsection{Environments and Offline Datasets}
We evaluate our method on three benchmarks for locomotion and robotic manipulation: \textbf{MuJoCo}~\citep{Mujoco, brockman2016openai}, \textbf{OGBench}~\citep{Ogbench}, and \textbf{Robomimic}~\citep{Robomimic}. MuJoCo tasks, which feature dense rewards, are used to evaluate from-scratch learning performance. Then we conduct offline-to-online experiments on OGBench and Robomimic, using their respective official offline datasets\footnote{OGBench: \url{https://github.com/seohongpark/ogbench}, Robomimic: \url{https://robomimic.github.io/docs/datasets/robomimic_v0.1.html}}.

\begin{figure}[t]
  \centering 

  \subfloat[Hopper-v4]{%
    \includegraphics[width=0.245\textwidth]{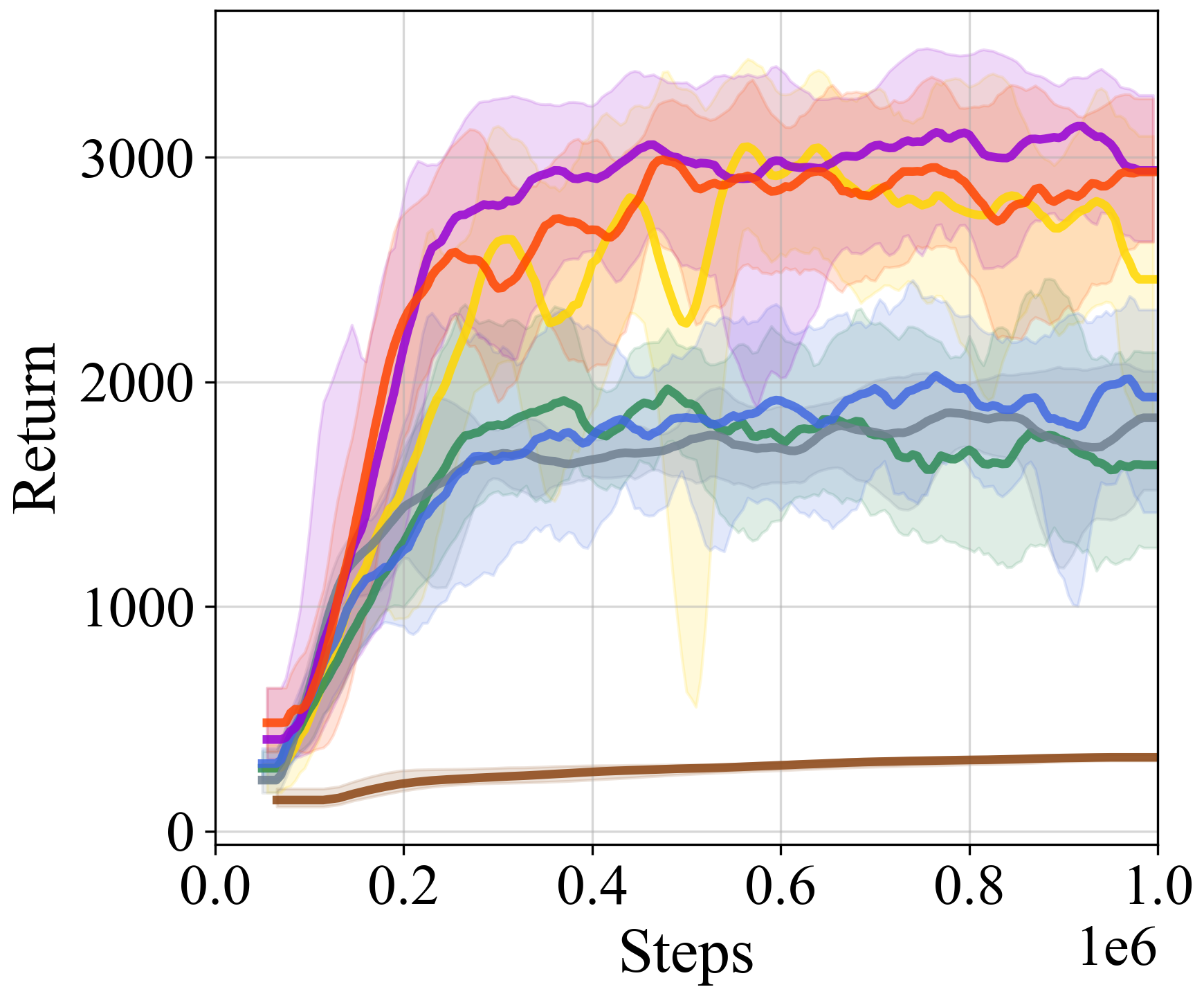}%
    \label{fig:hopper} 
  }
  \subfloat[Walker2D-v4]{%
    \includegraphics[width=0.245\textwidth]{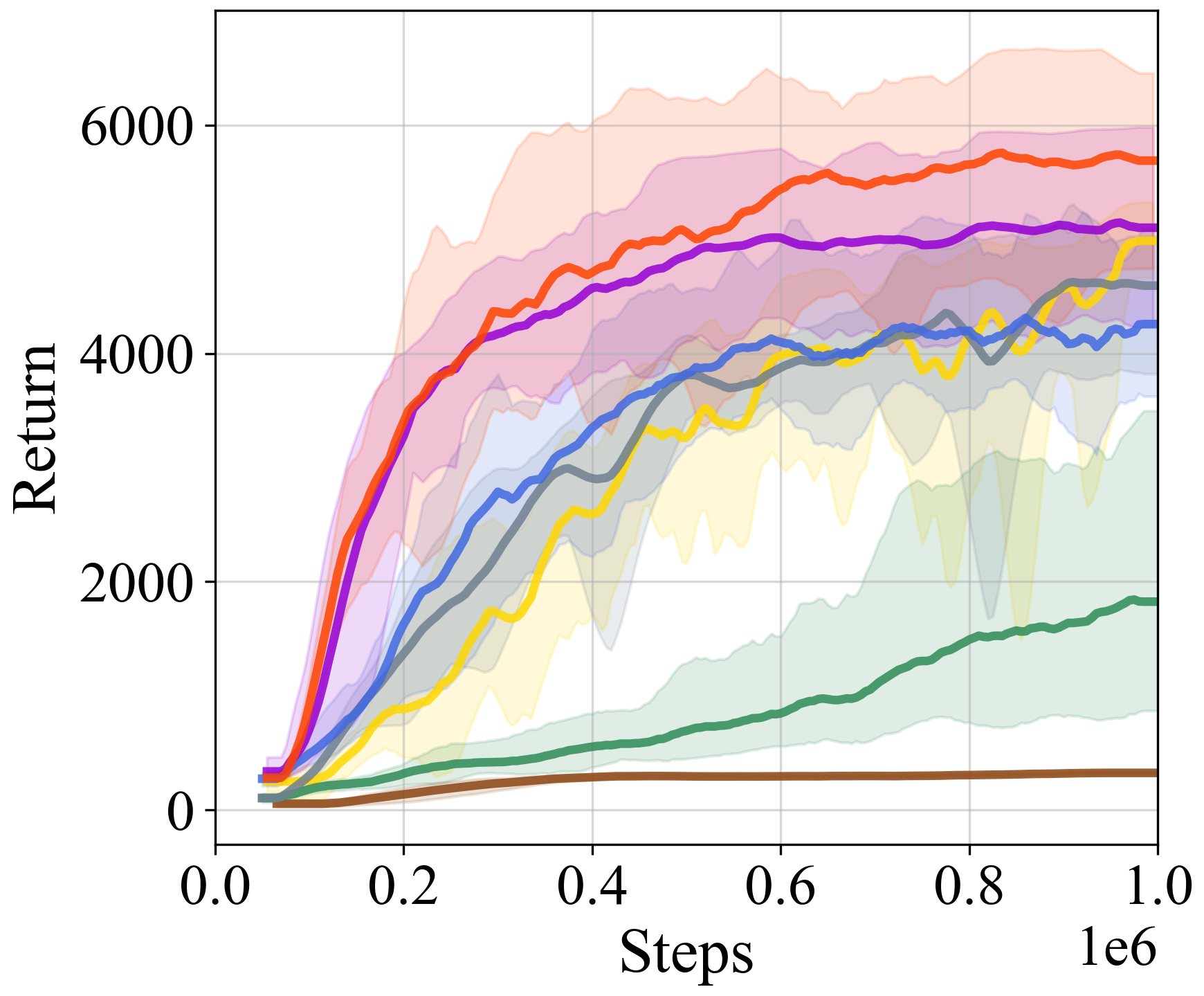}%
    \label{fig:walker} 
  }
  \subfloat[HalfCheetah-v4]{%
    \includegraphics[width=0.245\textwidth]{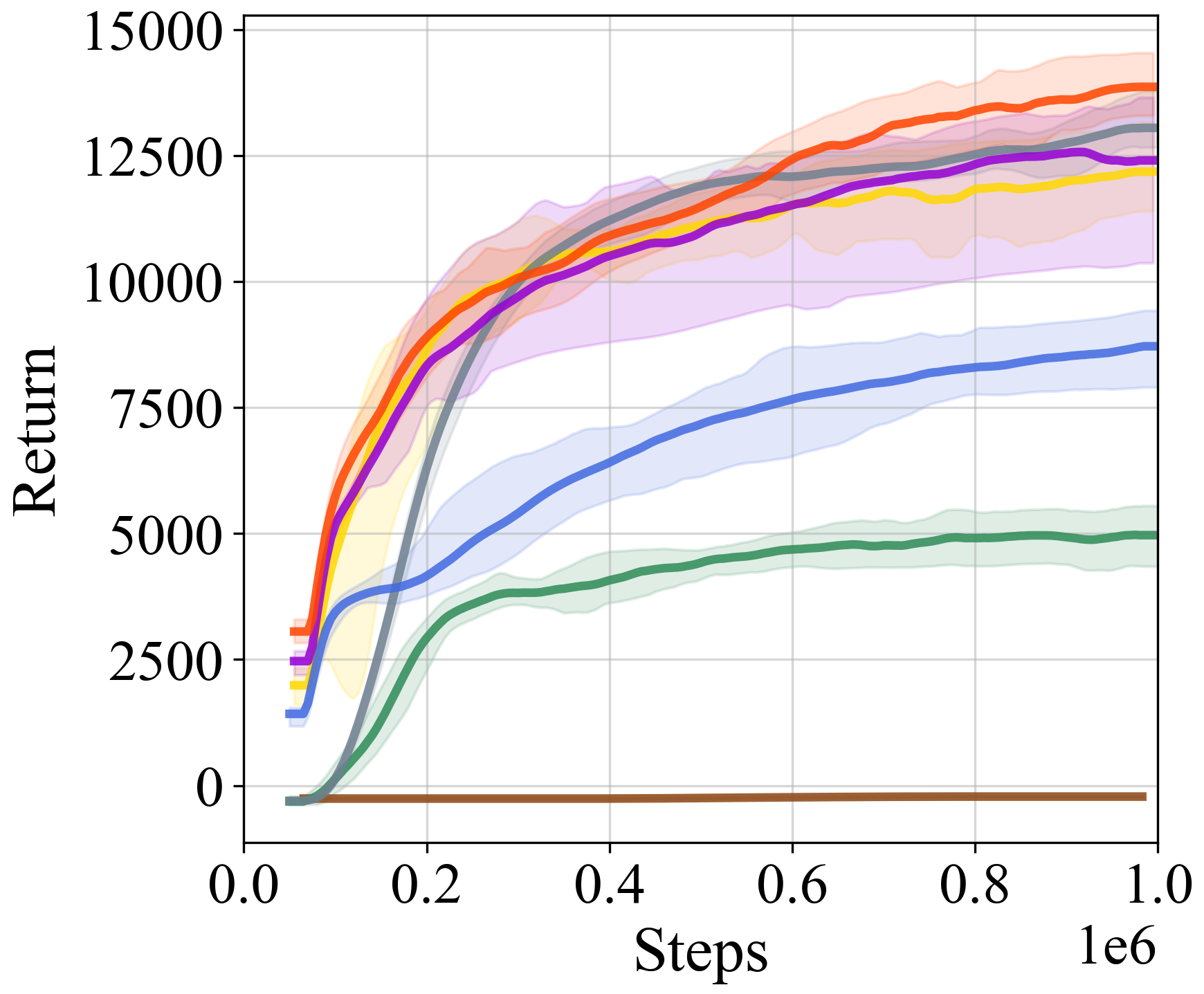}%
    \label{fig:halfcheetah} 
  }
  \subfloat[Ant-v4]{%
    \includegraphics[width=0.245\textwidth]{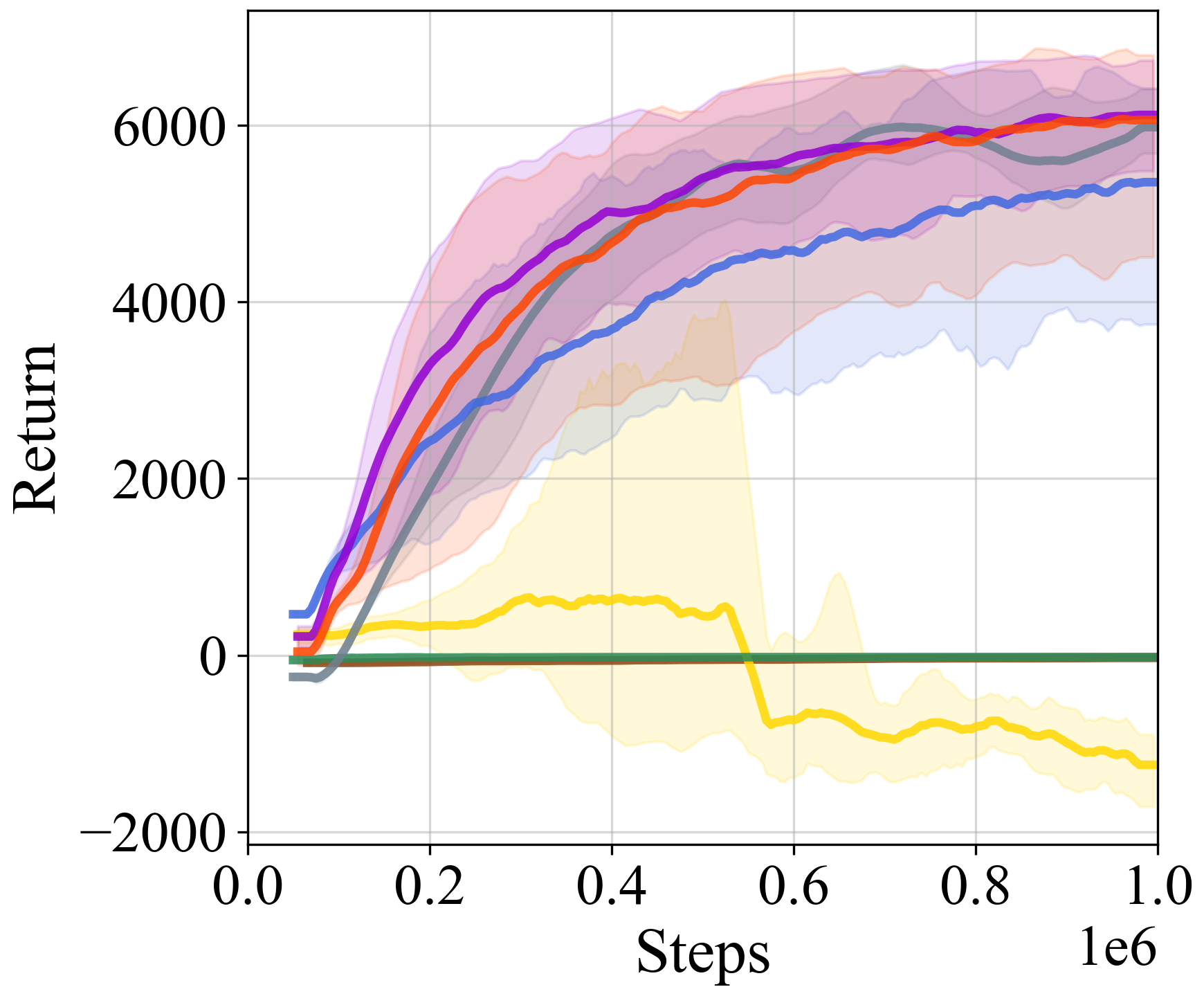}%
    \label{fig:ant} 
  }
  \\
  \subfloat[Humanoid-v4]{%
    \includegraphics[width=0.245\textwidth]{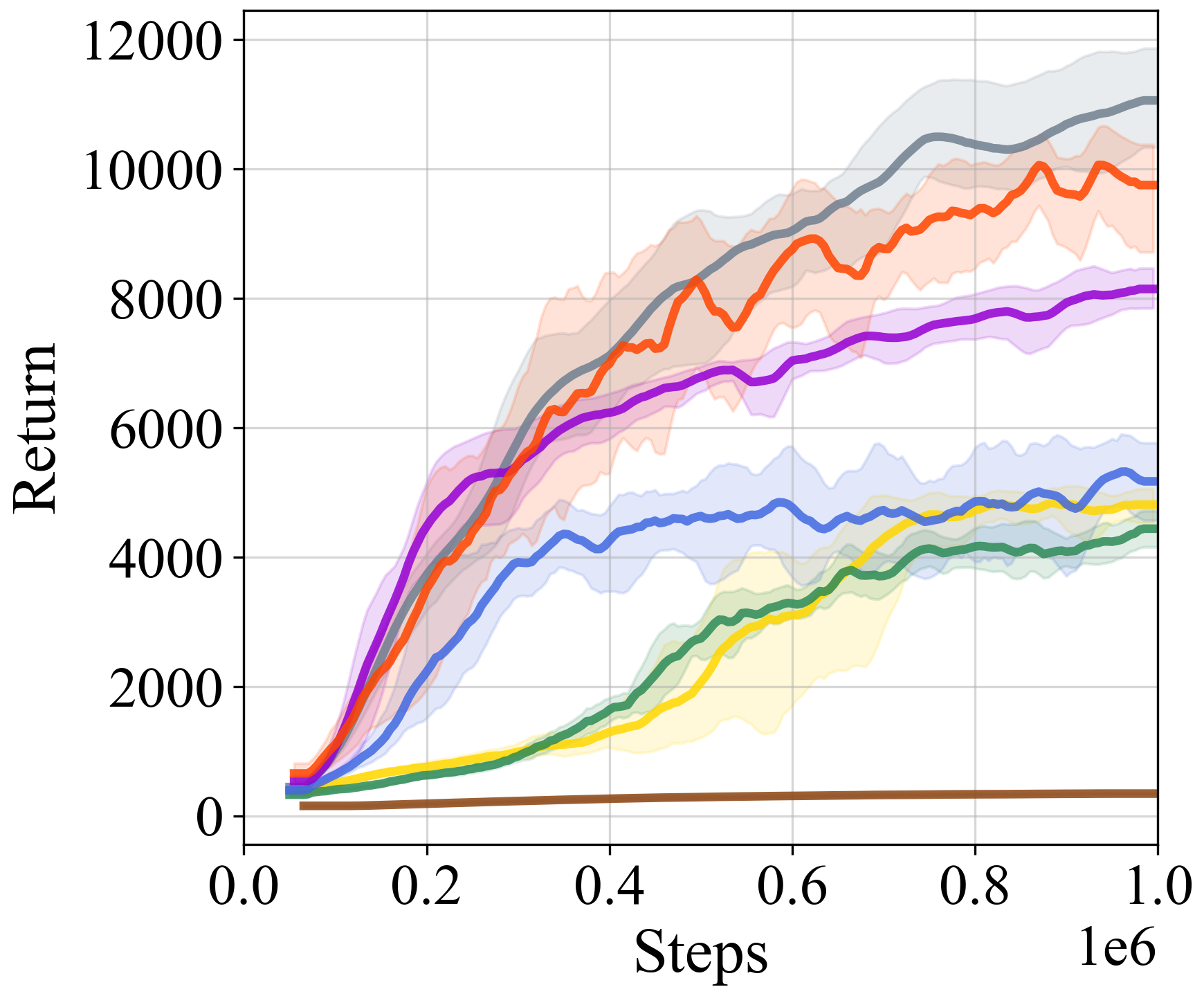}%
    \label{fig:humanoid} 
  }
  \subfloat[HumanoidStandup-v4]{%
    \includegraphics[width=0.245\textwidth]{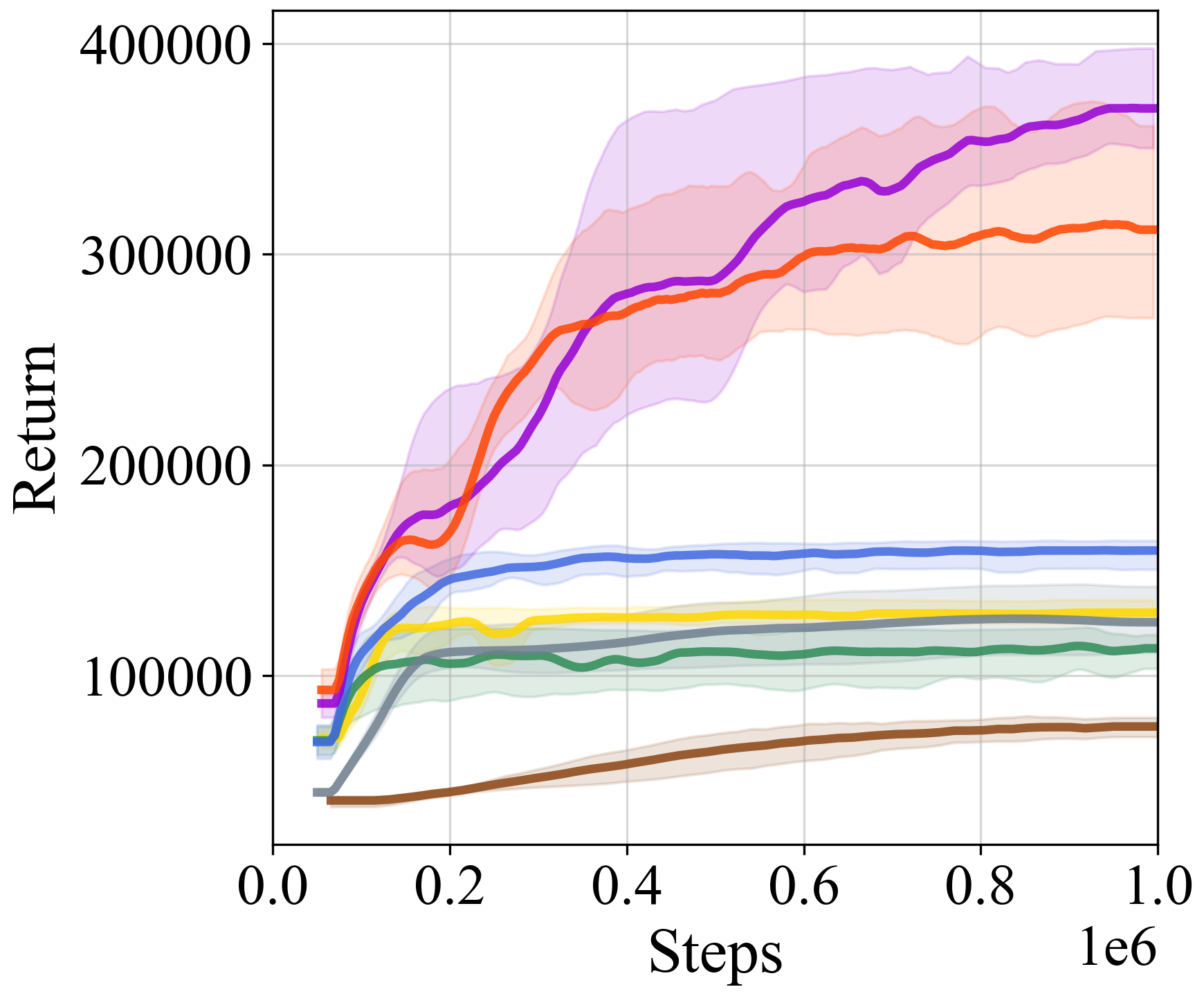}%
    \label{fig:humanoidstandup} 
  }
  \subfloat[Robomimic-Can]{%
    \includegraphics[width=0.245\textwidth]{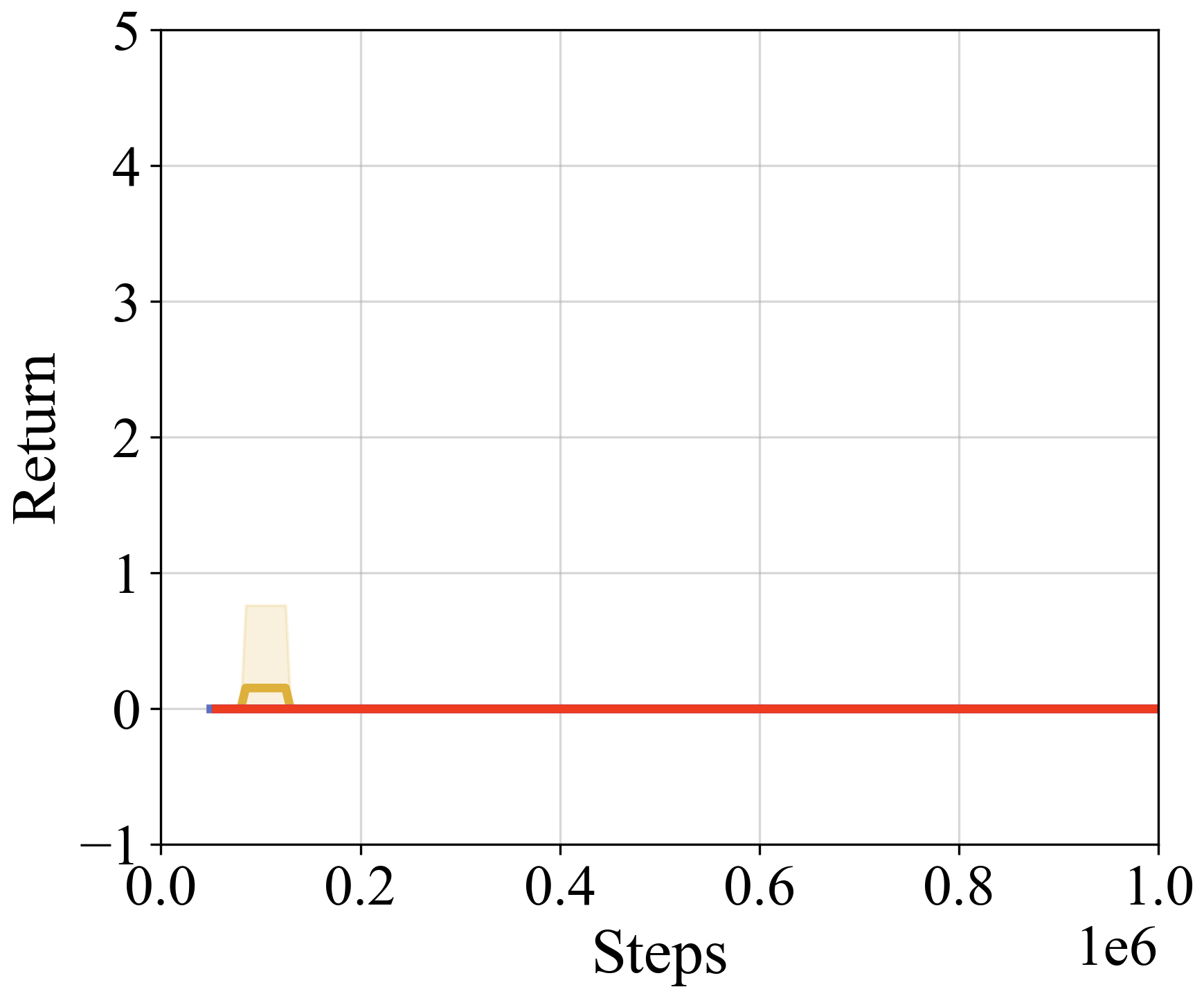}%
    \label{fig:robo-can} 
  }
  \subfloat[Cube-Double-Task2]{%
    \includegraphics[width=0.245\textwidth]{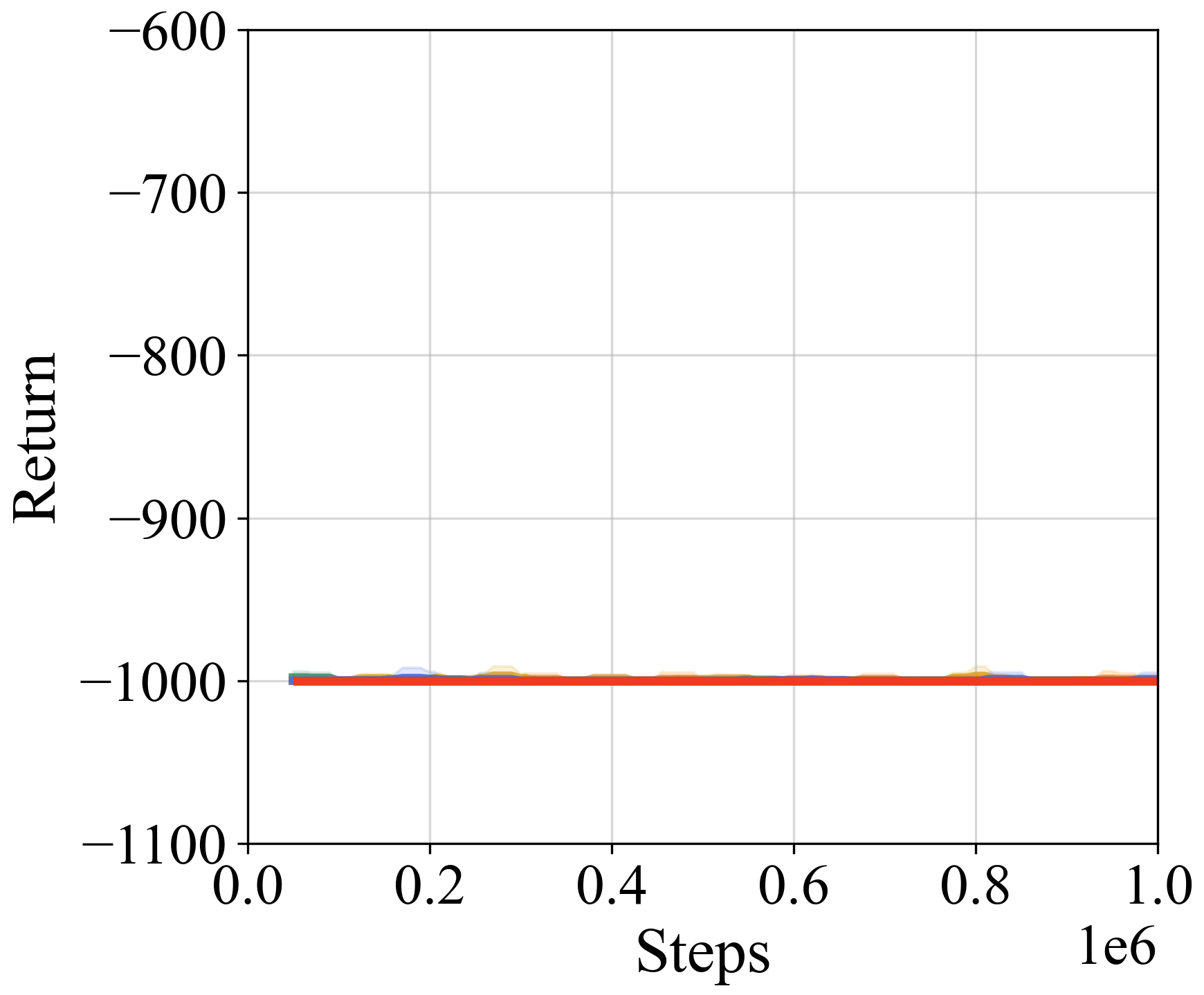}%
    \label{fig:og-cube} 
  }
  \\
  \subfloat{
  \includegraphics[width = 1.00\textwidth]{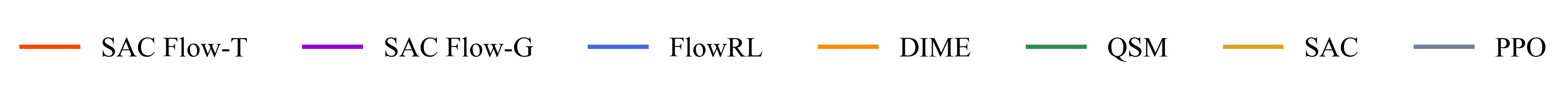}
  }
  \vspace{-3mm}
  \caption{From-scratch training performance. Our SAC Flow-T and SAC Flow-G achieve comparable or better performance accross all tasks except Humanoid (Fig. (a)-(f)), demonstrating significant sample efficiency and convergence stability. However, all methods struggle on the hard-exploration, sparse-reward tasks (\textit{Can} from \textbf{Robomimic}, and \textit{Cube-Double} from \textbf{OGBench}), highlighting the necessity of offline-to-online training.}
  \label{fig:main_res-fs}
  \vspace{-2mm}
\end{figure}

\subsubsection{Baselines}
For the from-scratch training, we compare SAC-Flow against five baselines.
\textbf{(1) Q-score matching (QSM)}~\citep{QSM} directly optimizes the diffusion policy's score function using the gradient of the Q-function. 
\textbf{(2) DIME}~\citep{DIME} is a representative max-entropy RL method for diffusion policy, addressing the challenge of entropy calculation.
\textbf{(3) FlowRL}~\citep{FlowRL} is the state-of-the-art (SOTA) method, which trains a flow-based policy by directly maximizing the Q-value, regularized by a Wasserstein-2 constraint.
Finally we apply two classical RL algorithms: \textbf{(4) SAC}~\citep{SAC_arxiv} and \textbf{(5) PPO}~\citep{PPO}, with Gaussian policies as fundamental from-scratch baselines.

To evaluate the offline-to-online capability, we select three baselines, including on-policy and off-policy methods. \textbf{(1) ReinFlow}~\citep{ReinFlow} solves the difficulty of calculating log probability through multi-step flow inference, enabling on-policy PPO update for flow-based policy. It should be noted that ReinFlow is only tested in Robomimic due to a lack of official implementation for its use in OGBench.
\textbf{(2) Flow Q-Learning (FQL)} \citep{FlowQ} uses SAC-style update to achieve high data-efficient RL tuning. FQL uses a one-step policy to estimate the flow model, avoiding the instability of backpropagation through time. And its successor, \textbf{(3) Q-chunking FQL (QC-FQL)} \citep{QC-FQL},  extends FQL to handle action chunking by operating in temporally extended action spaces.

Among all experiments, the sampling steps of flow-based policies are set to 4, and the denoising steps of diffusion policies are set to 16. More details of the experimental setting are described in Appendix \ref{sec:Env_settings} and Appendix \ref{sec:imp}.

\subsection{Main results}
\begin{figure}[t]
  \centering 

  \subfloat[Cube-Double-Task]{%
    \includegraphics[width=0.245\textwidth]{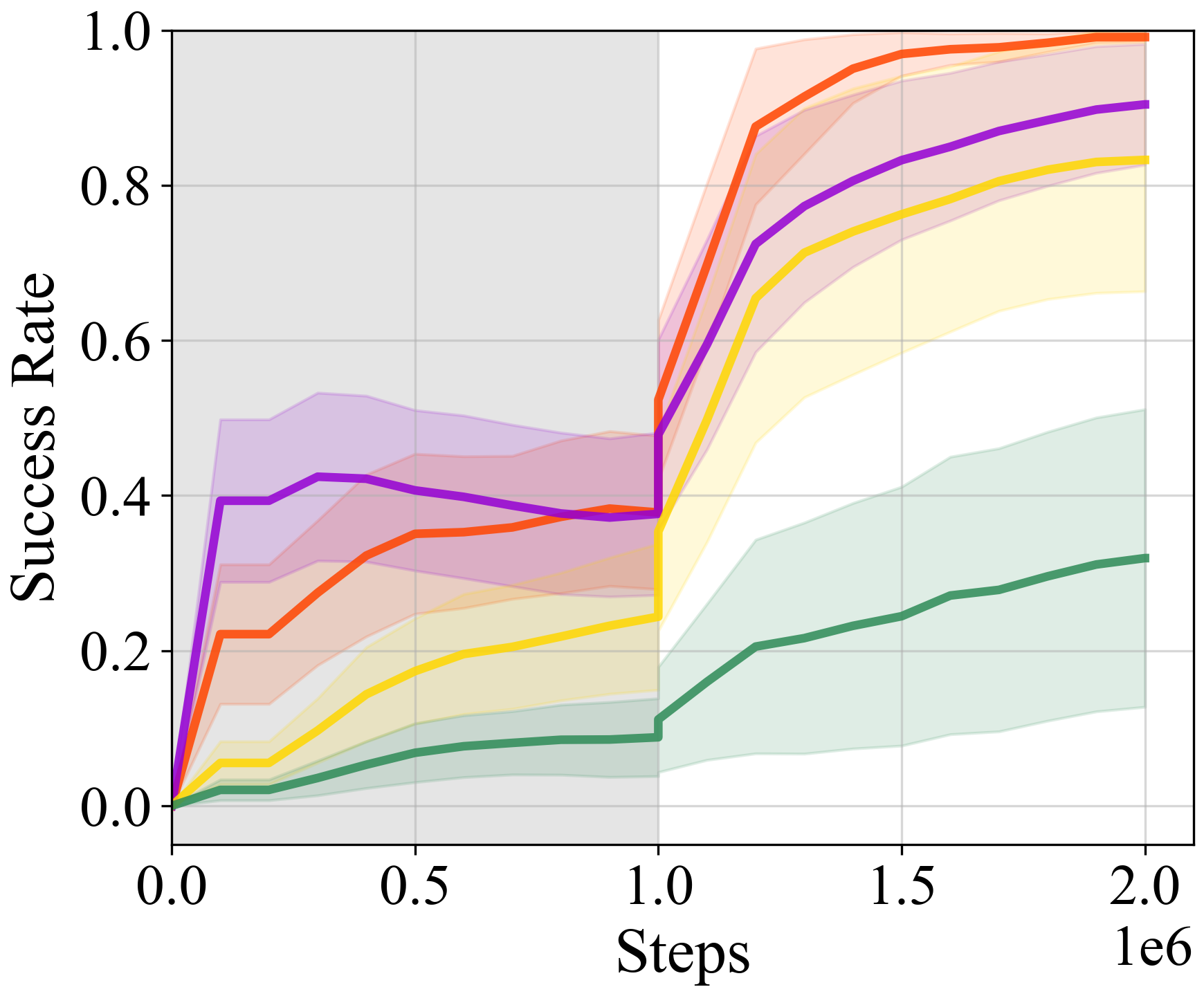}%
    \label{fig:quadruple-task} 
  }
  \subfloat[Cube-Triple-Task]{%
    \includegraphics[width=0.245\textwidth]{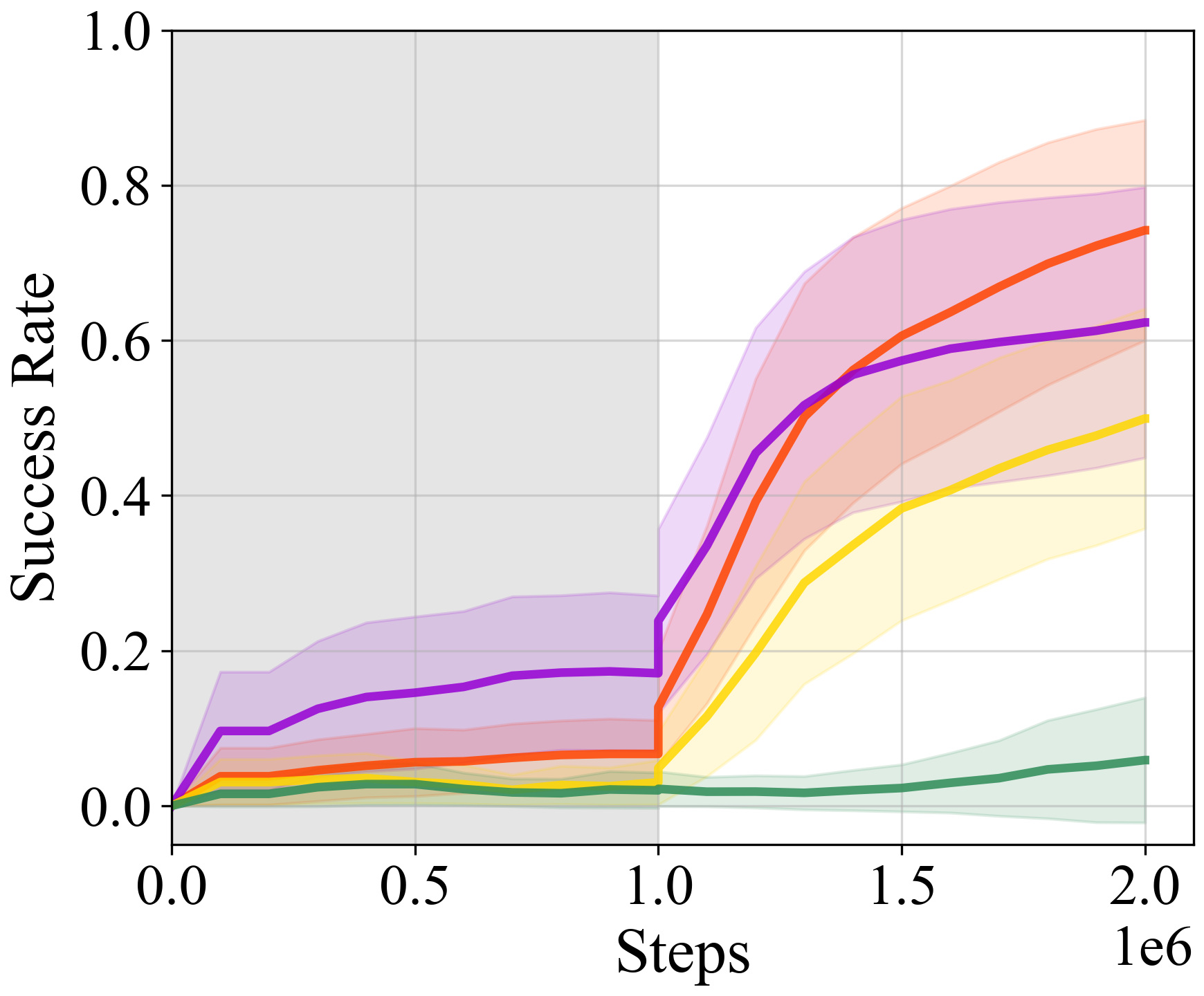}%
    \label{fig:triple-task} 
  }
  \subfloat[Cube-Quadruple-Task]{%
    \includegraphics[width=0.245\textwidth]{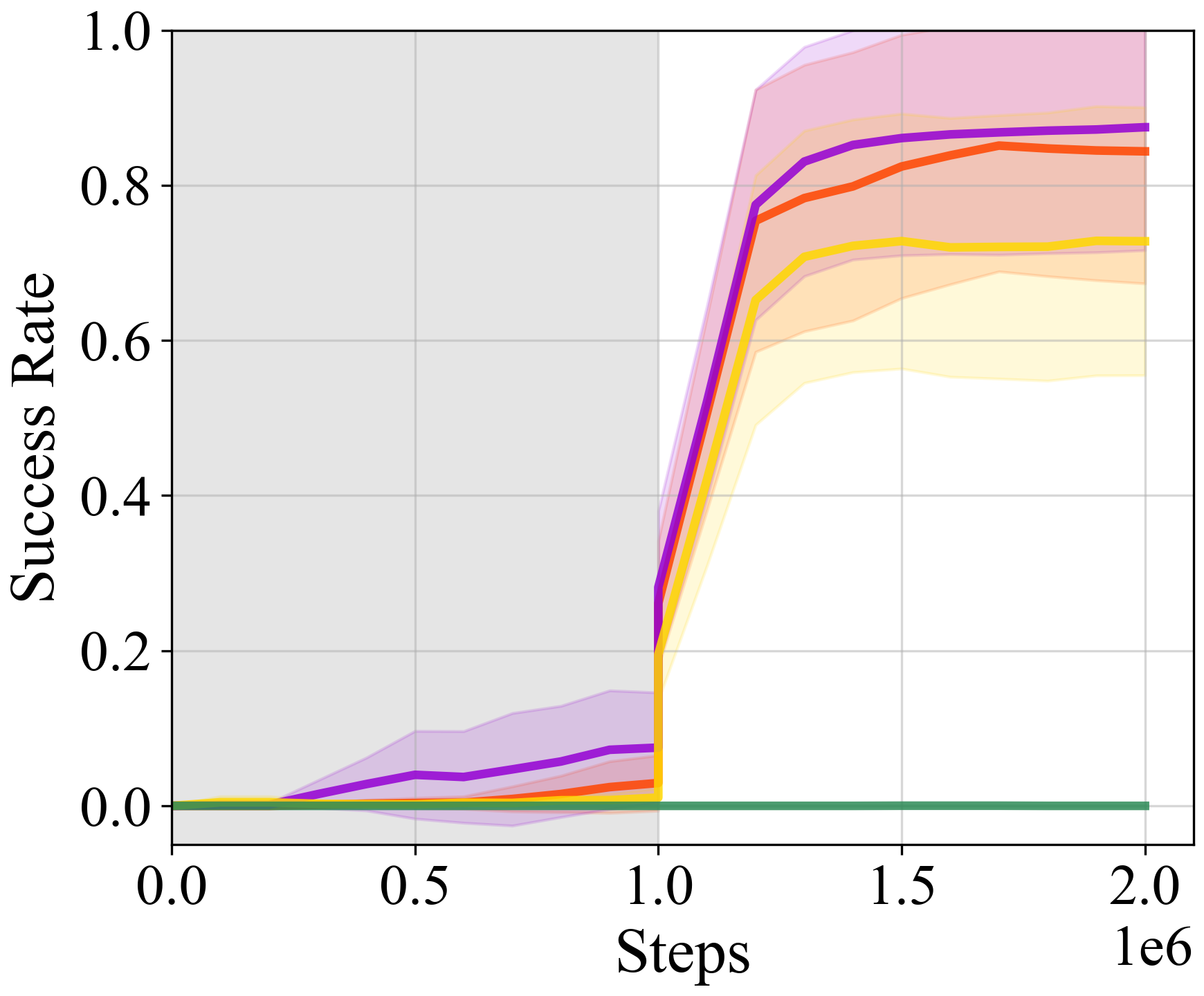}%
    \label{fig:double-task} 
  }
  \subfloat[Robomimic]{%
    \includegraphics[width=0.245\textwidth]{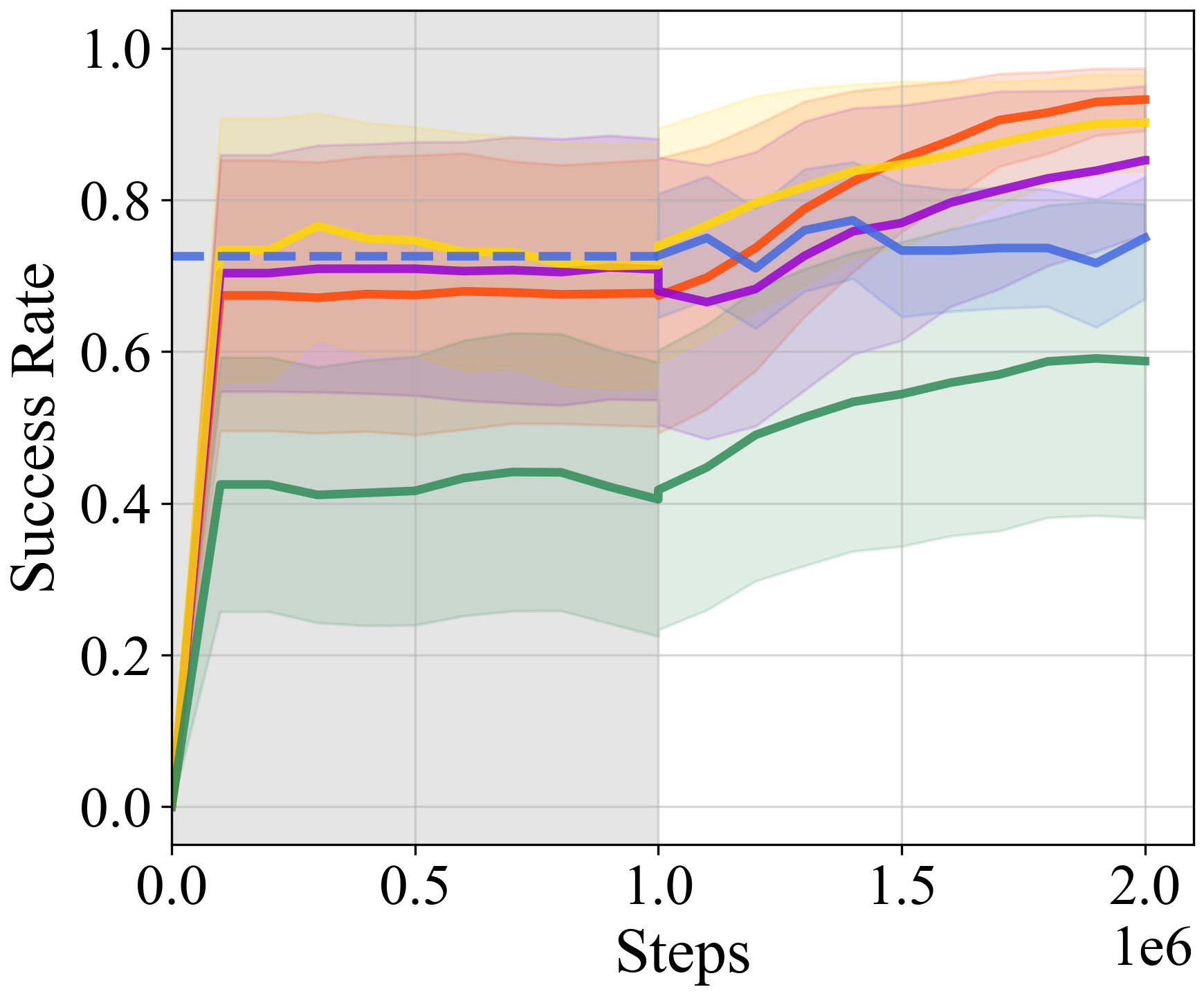}%
    \label{fig:o2o-robo} 
  }
  \\
  \subfloat{
  \includegraphics[width = 0.85\textwidth]{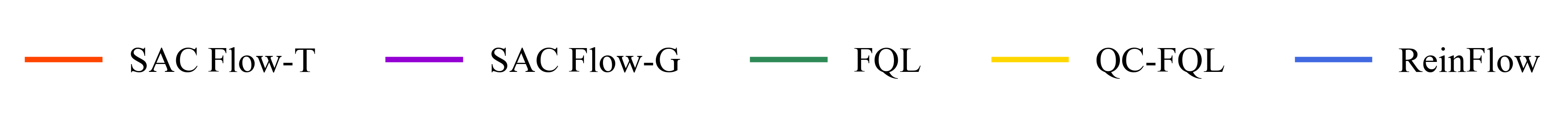}
  }
  \vspace{-3mm}
\caption{
    Aggregated offline-to-online performance on OGBench and Robomimic benchmarks. Each curve shows the mean success rate averaged across multiple task instances within a domain. Specifically, the \textbf{OGBench} results for \textit{Cube-Double}, \textit{Triple}, and \textit{Quadruple} (a-c) are each aggregated over five distinct single-task environments. The \textbf{Robomimic} result (d) is aggregated across the \textit{Lift}, \textit{Can}, and \textit{Square} tasks.
}
  \label{fig:Offline-to-online} 
  \vspace{-2mm}
\end{figure}
Fig. \ref{fig:main_res-fs} illustrates the results for from-scratch training. Our methods, SAC Flow-G and SAC Flow-T, achieve superior or comparable performance across most MuJoCo tasks, with the exception of Humanoid. Although DIME and FlowRL generally converge faster than other baselines, our methods consistently surpass FlowRL, benefiting from direct optimization of the SAC objective. Furthermore, SAC Flow outperforms DIME in Hopper (Fig.~\ref{fig:hopper}), Walker (Fig.~\ref{fig:walker}), and HumanoidStandup (Fig.~\ref{fig:humanoidstandup}), while achieving comparable results in HalfCheetah (Fig.~\ref{fig:halfcheetah}) and Ant (Fig.~\ref{fig:ant}). Moreover, with the expressive parameterization of flow-based policy, our method achieves much higher final performance in challenging tasks, demonstrating up to a $130\%$ improvement over the baseline ( \ref{fig:humanoidstandup}), and remains convergence stability in simple tasks (Fig. \ref{fig:hopper}, \ref{fig:walker}, and \ref{fig:halfcheetah}). For reference, we include the on-policy baseline, PPO, to highlight the superior sample efficiency of off-policy algorithms. However, all from-scratch methods struggle in tasks with large exploration spaces and sparse rewards (Figs. \ref{fig:robo-can}, \ref{fig:og-cube}), underscoring the necessity of an offline-to-online training setting.

Fig. \ref{fig:Offline-to-online} shows the offline-to-online training performance in sparse reward tasks. All methods are trained on 1M offline updates followed by 1M online steps. In the challenging OGBench environments, including cube-triple and cube-quadruple, our proposed methods, particularly SAC Flow-T, achieve rapid convergence and attain a state-of-the-art overall success rate, demonstrating up to a $60\%$ improvement over the baseline. In the Robomimic environment, however, SAC Flow-T and SAC Flow-G only yield results comparable to QC-FQL. This is primarily because the training is strictly regularized with a large $\beta$ value (Equ. ($\ref{actor_o2o}$)). As a result, the learning capacity of the flow model is severely limited, causing its performance to be similar to that of the one-step policy in QC-FQL. We further compare the on-policy baseline, Reinflow, in Robomimic. Leveraging the high data efficiency of off-policy learning, our SAC Flow-G and SAC Flow-T outperform Reinflow under 1M online steps. The additional results are available in Appendix \ref{more_exp}.
\begin{figure}[t]
  \centering 

  \subfloat[\centering Average Gradient Norm for Walker2d]{%
    \includegraphics[width=0.35\textwidth]{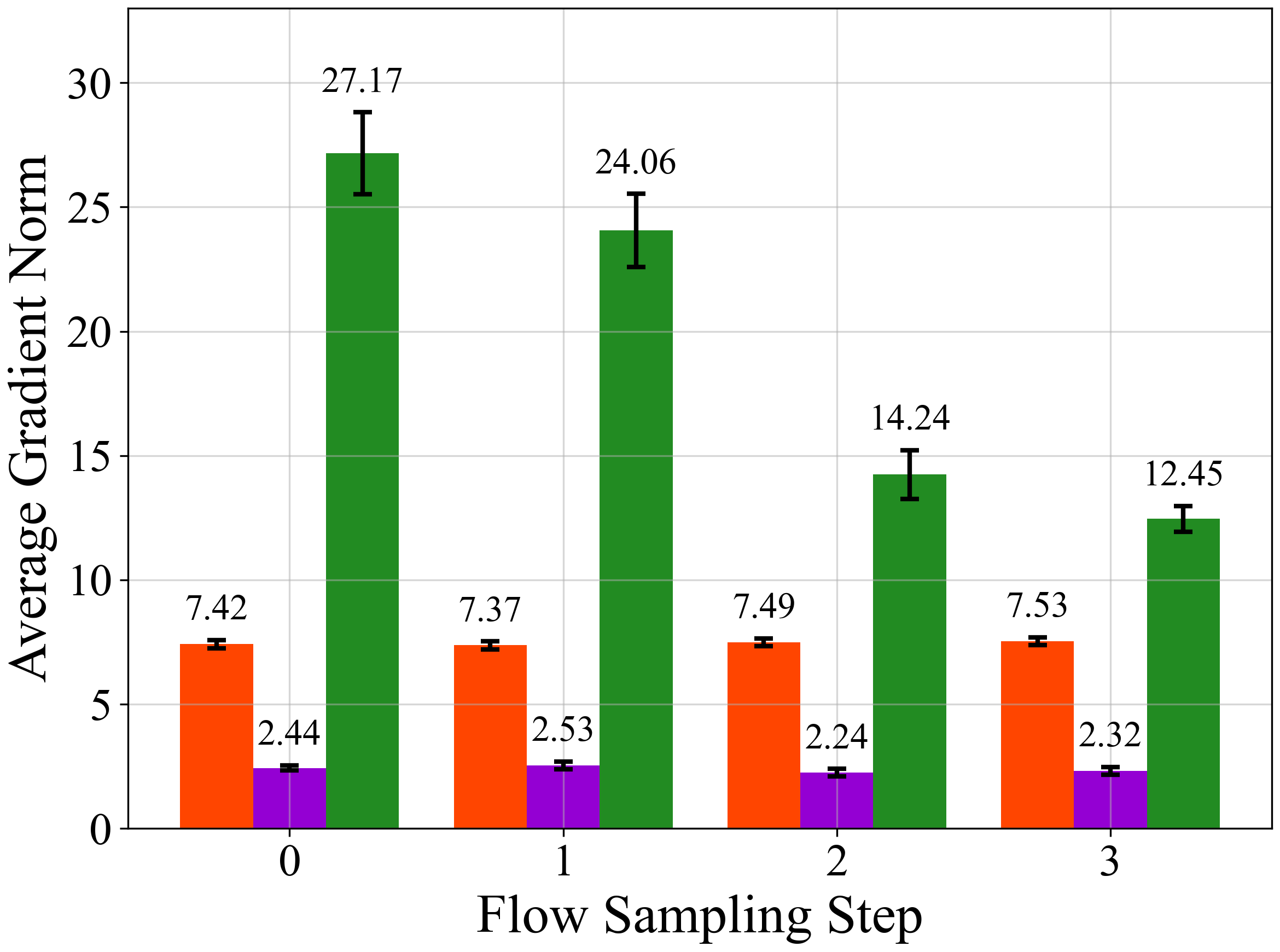}%
    \label{fig:bars} 
  }
  \subfloat[\centering Ant-v4]{%
    \includegraphics[width=0.315\textwidth]{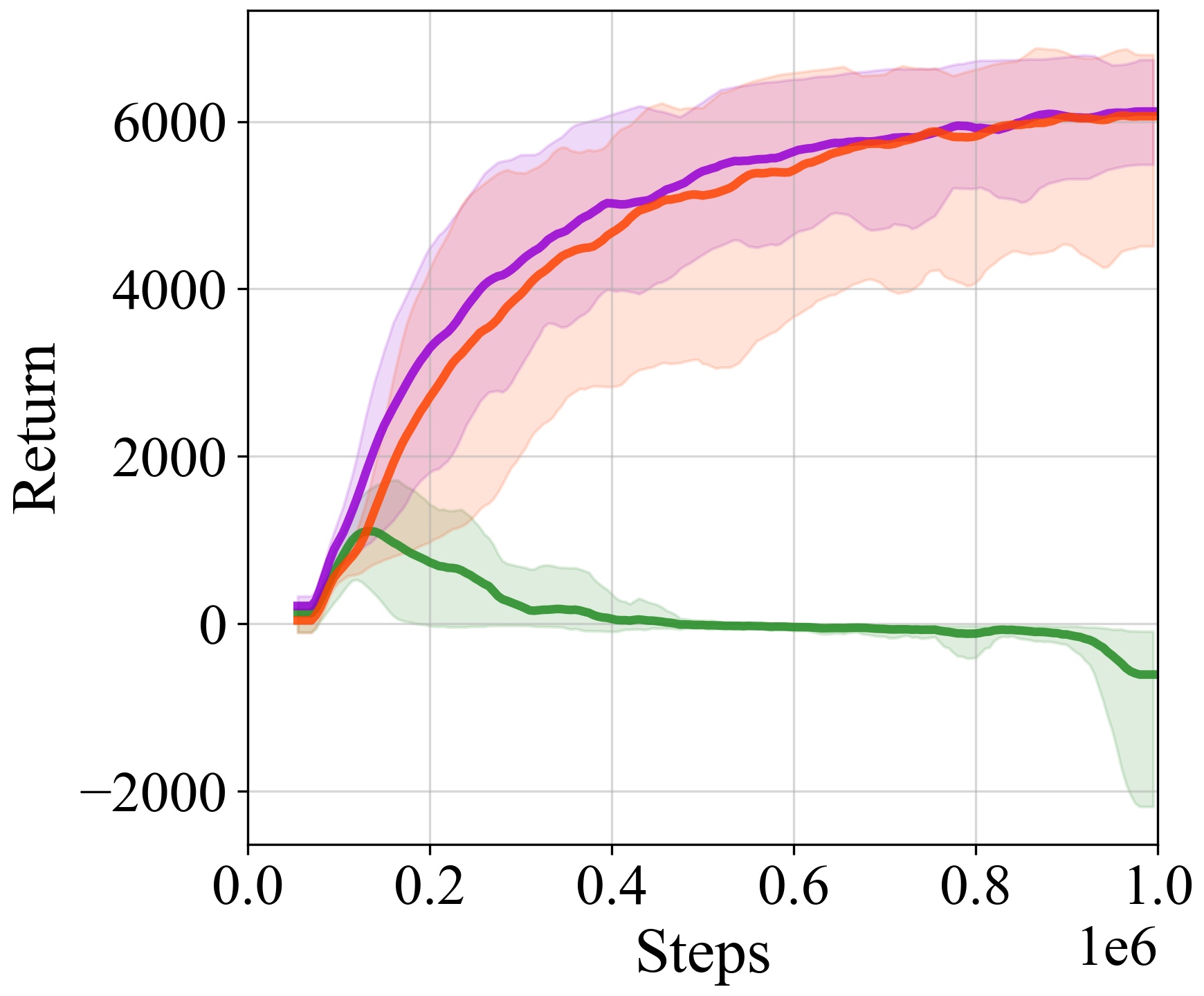}%
    \label{fig:gn_ant} 
  }
  \subfloat[\centering Cube-Double-Task4]{%
    \includegraphics[width=0.315\textwidth]{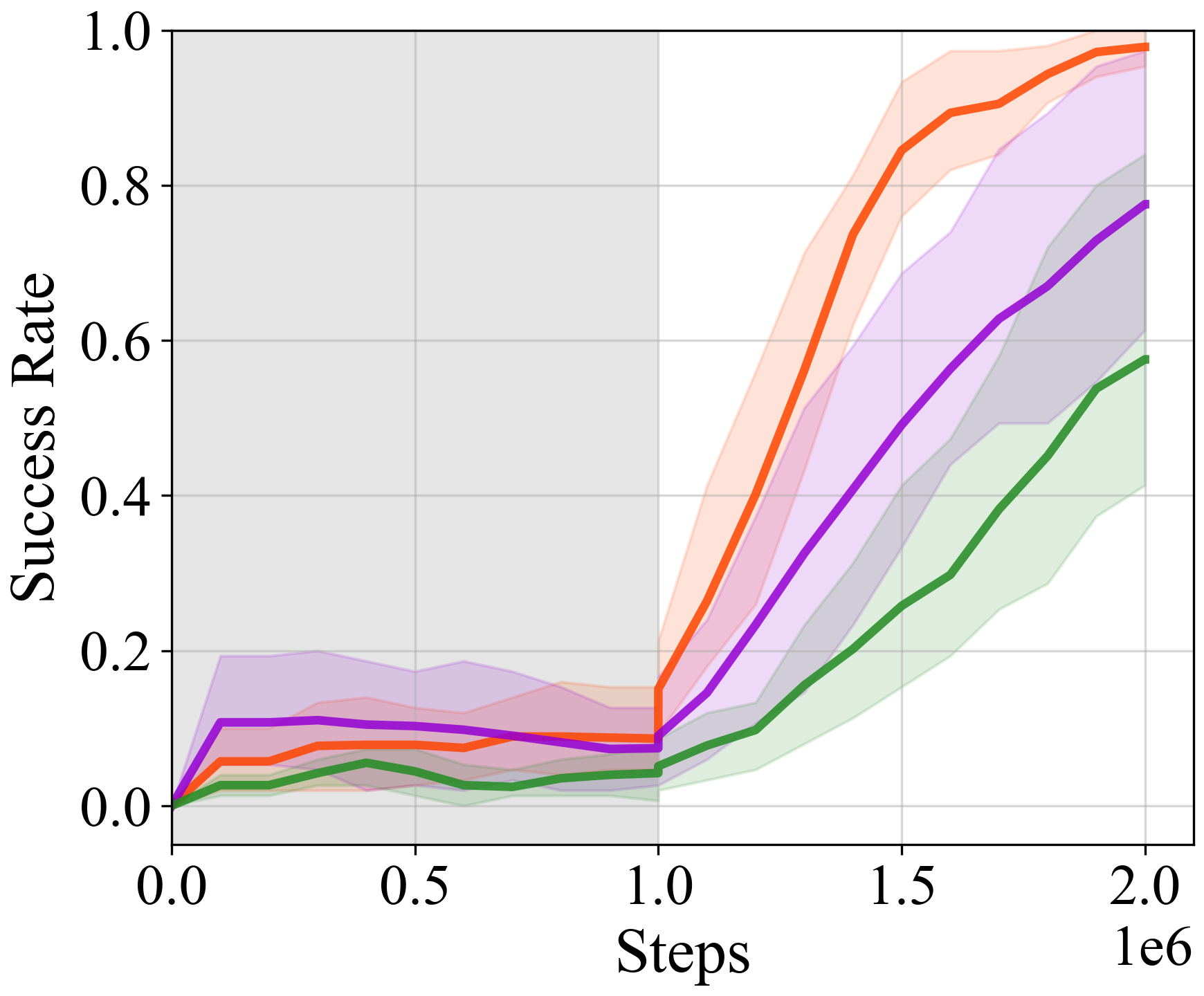}%
    \label{fig:gn_cube} 
  }
  \\
  \vspace{-1mm}
  \subfloat{
  \includegraphics[width = 0.7\textwidth]{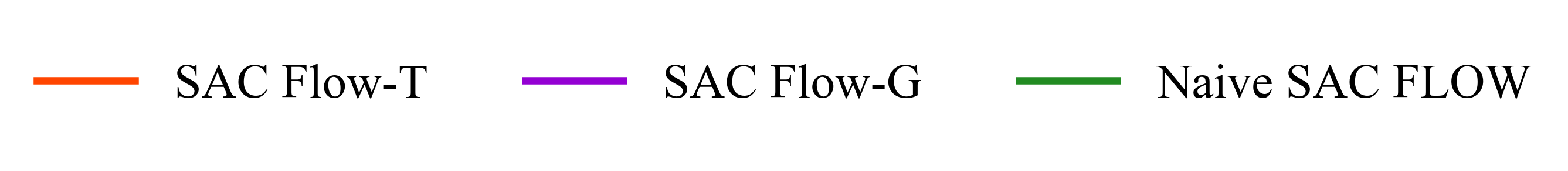}
  }
  \vspace{-4mm}
  \caption{Ablation study on velocity network parameterizations. Our SAC Flow-T and SAC Flow-G significantly reduce the gradient exploding and enable stable training.
  }
  \label{fig:gradient_norm} 
  \vspace{-2mm}
\end{figure}

\subsection{Ablation Study}
\begin{figure}[htbp]
  \centering 

  \subfloat[Ant-v4]{%
    \includegraphics[width=0.33\textwidth]{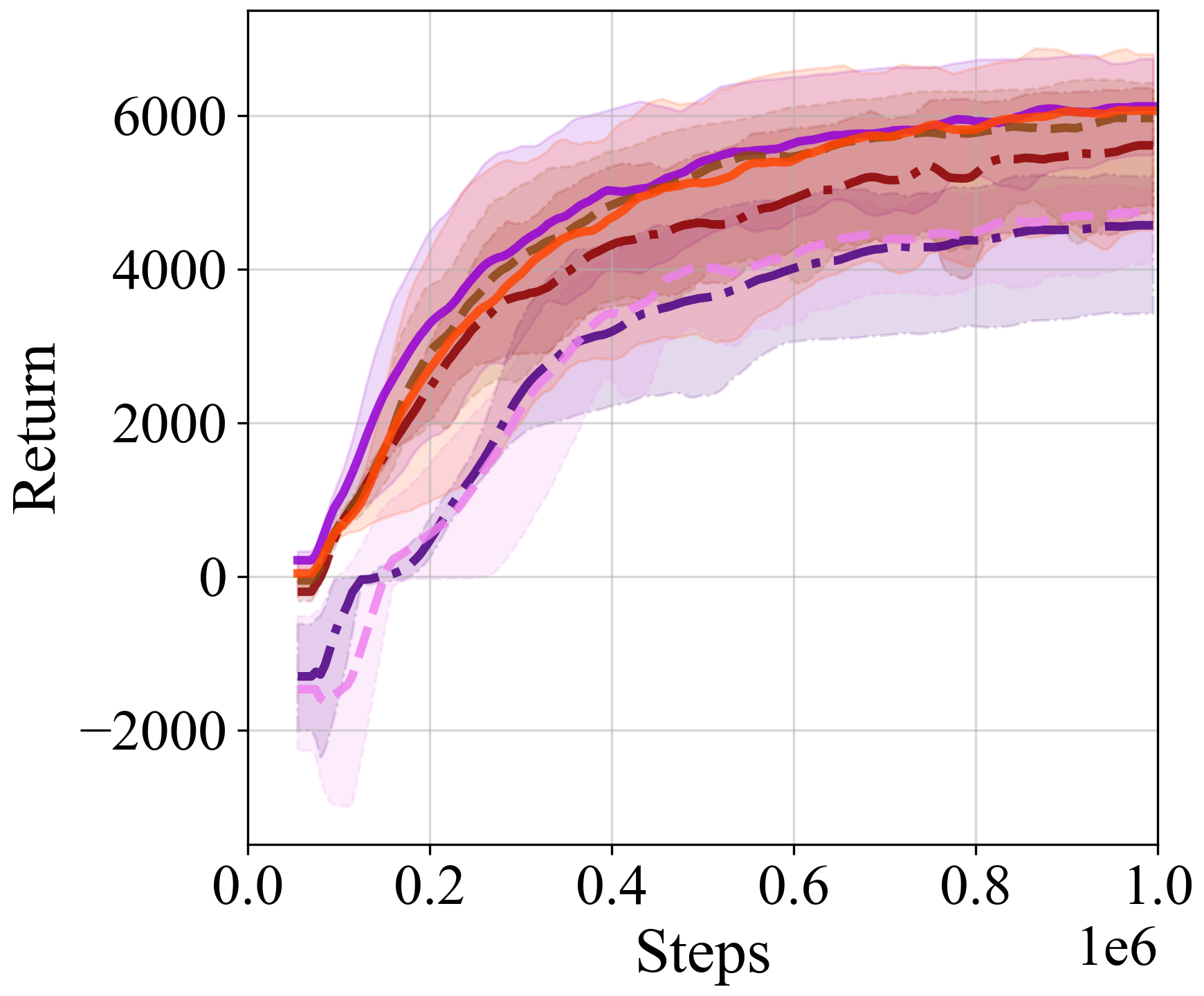}%
    \label{fig:abl-ant} 
  }
  \subfloat[Cube-Double-Task4]{%
    \includegraphics[width=0.33\textwidth]{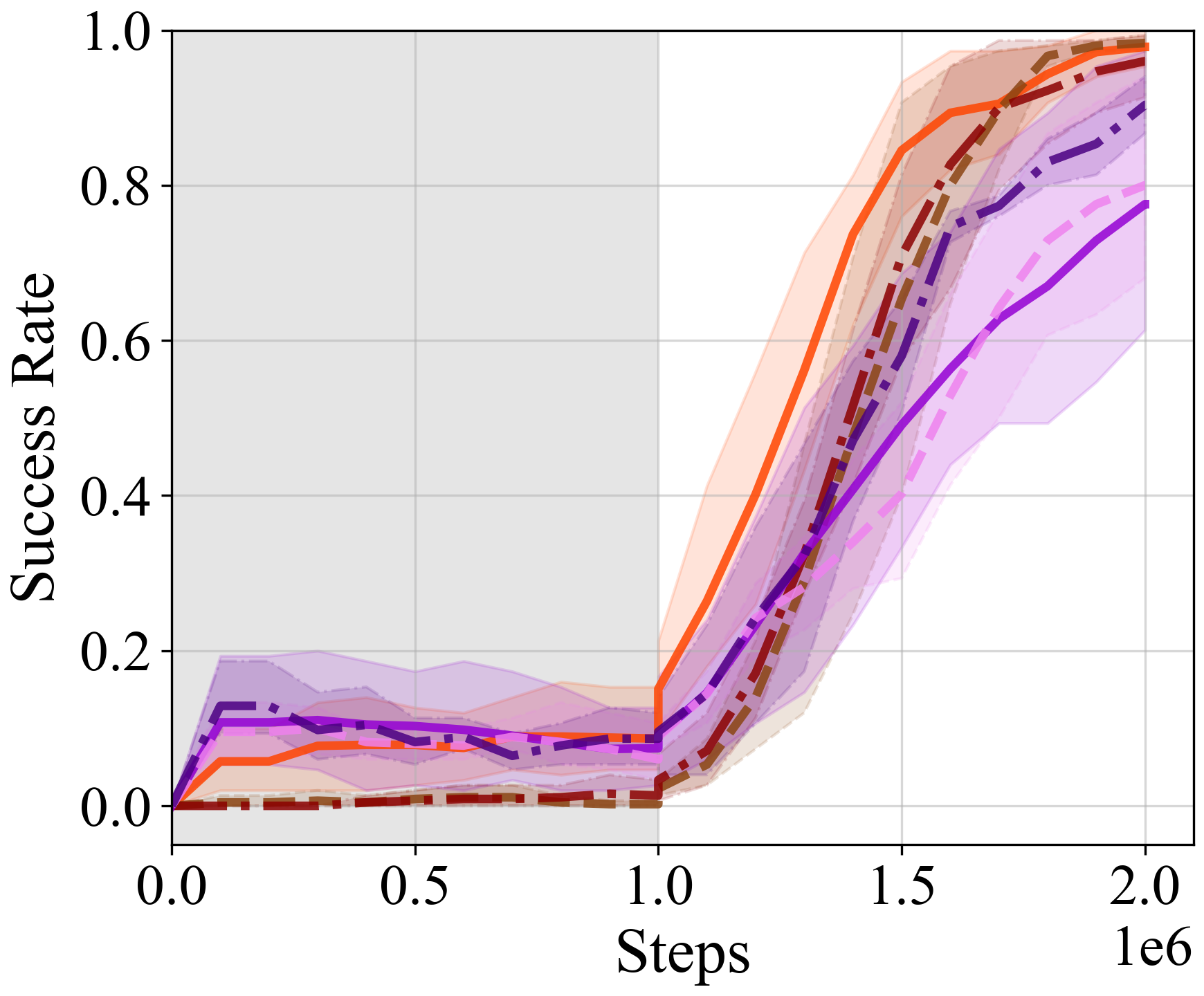}%
    \label{fig:abl-cube} 
  }
  \\
  \vspace{-1mm}
  \subfloat{
    \includegraphics[width = 0.65\textwidth]{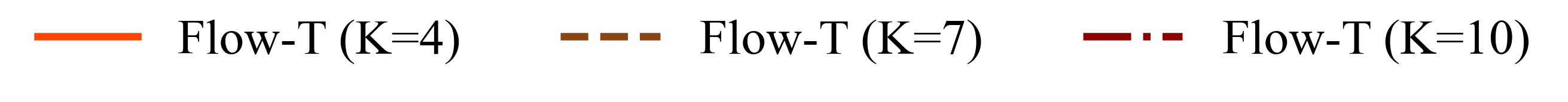}
  }
  \\
  \vspace{-0.5mm}
  \subfloat{
    \includegraphics[width = 0.65\textwidth]{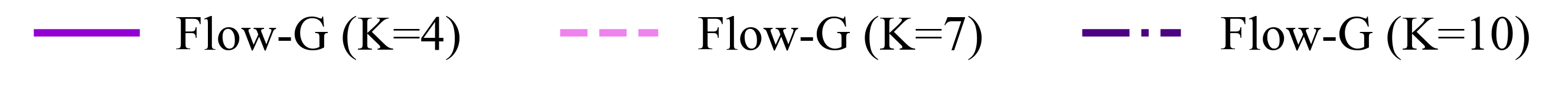}
  }
  \vspace{-4mm}
  \caption{Ablation study on flow sampling steps. Our SAC Flow-G and SAC Flow-T are robust to the number of sampling steps.}
  \label{fig:abl-steps} 
  \vspace{-2mm}
\end{figure}
\paragraph{Ablation study on velocity network parameterizations.}
We first visualize the gradient norms of our SAC Flow-G and SAC Flow-T. We compare against a naive SAC Flow baseline, which directly applies SAC to the flow-based model without reparameterizing the velocity network. As shown in Fig. \ref{fig:bars}, the naive baseline suffers from exploding gradients: the average norm escalates sharply along the backpropagation path (i.e., from sampling step $k=3$ to $k=0$). In contrast, our methods maintain stable gradient norms across the whole backpropagation, with a maximum variation of 0.29.
This gradient instability directly translates to poor performance, as the naive SAC Flow performs poorly in both from-scratch and offline-to-online settings (Figs. \ref{fig:gn_ant} and \ref{fig:gn_cube}). These results confirm that our approach effectively mitigates exploding gradients, enabling stable and high-performance training.

\paragraph{Ablation study on flow sampling steps.}
Fig. \ref{fig:abl-steps} shows the performance of SAC FLow-T and SAC Flow-G under sampling steps $K=4, 7, 10$. A larger number of sampling steps can further challenge the stability of gradient backpropagation. The experiments show that our approach, especially SAC Flow-T, is robust to the number of sampling steps.

\section{Conclusion}
In this paper, we introduce SAC Flow, a sample-efficient and high-performance off-policy RL algorithm for flow-based policies. SAC Flow addresses the issue of gradient instability in training flow-based policies by treating the flow-based model as a sequential model and reparameterizing its velocity network as a GRU or a Transformer. We evaluate the performance of SAC Flow in both from-scratch and offline-to-online training settings. SAC Flow demonstrates rapid convergence and achieves state-of-the-art performance across multiple locomotion and manipulation tasks.

Looking forward, we will validate SAC Flow on real robots, explore lighter sequential parameterizations, and study sim-to-real robustness for reliable deployment. 


\bibliography{iclr2026_conference}
\bibliographystyle{plainnat}

\newpage
\appendix
\section{The derivation of SAC loss in the flow-based policy}
\label{SAC_loss_flow}

This appendix consolidates and expands our derivations for training SAC on a $K$-step flow rollout, including the likelihood construction via a noise-augmented rollout, the joint path density, the pathwise score expansion, gradients for actor/critic, the temperature update, and practical notes for implementation.

\subsection{Noise-augmented rollout and drift correction}
\label{app:noise_rollout}

We start from the deterministic $K$-step Euler rollout in Equ. (\ref{FR}):
\[
A_{t_{i+1}} = A_{t_i} + \Delta t_i\, v_\theta(t_i,A_{t_i},s), \quad 0=t_0<\cdots<t_K=1.
\]
For likelihood-based training, we convert it into a stochastic rollout that leaves the final marginal invariant by adding isotropic Gaussian noise with a compensating drift:
\begin{equation}
\label{app:noisy_step}
A_{t_{i+1}}
=
A_{t_i}
+
b_\theta(t_i,A_{t_i},s)\,\Delta t_i
+
\sigma_\theta \sqrt{\Delta t_i}\,\varepsilon_i,\quad \varepsilon_i\sim\mathcal{N}(0,I_d).
\end{equation}
A convenient drift that matches rectified-flow families is
\begin{equation}
\label{app:drift_def}
b_\theta(t_i,A_{t_i},s)
=
\left(\frac{1-t_i+\tfrac{t_i\,\sigma_\theta^2}{2}}{1-t_i}\right) v_\theta(t_i,A_{t_i},s)
-
\left(\frac{t_i\,\sigma_\theta^2}{2(1-t_i)\,t_i}\right) A_{t_i},
\end{equation}
with $b_\theta(0,\cdot,\cdot)=v_\theta(0,\cdot,\cdot)$. Intuitively, the first factor inflates the learned velocity to counteract diffusion, and the second term contracts towards the straight path so that the terminal law remains unchanged.

\paragraph{Per-step transition.}
Under Equ. (\ref{app:noisy_step}), the conditional $A_{t_{i+1}}\mid A_{t_i},s$ is Gaussian:
\[
\eta_\theta\big(A_{t_{i+1}}\mid A_{t_i},s;\Delta t_i\big)
=
\mathcal{N}\Big(
A_{t_i}+b_\theta(t_i,A_{t_i},s)\,\Delta t_i,\;
\sigma_\theta^2\,\Delta t_i\, I_d
\Big).
\]
We denote $A_{t_0}\sim \mathcal{N}(0,I_d)$ as the base. The final action is $a=\tanh(A_{t_K})$.

\subsection{Joint path density and squashing Jacobian}
\label{app:path_density}

Let $\mathcal{A}=(A_{t_0},\ldots,A_{t_K})$. The joint density factorizes as
\begin{equation}
\label{eq:joint_eta_fact_app}
p_c(\mathcal{A} \mid s) = \zeta(A_{t_0}) \prod_{i=0}^{K-1} \eta_\theta\left(A_{t_{i+1}} \mid A_{t_i}, s; \Delta t_i \right) \cdot \|\det \mathcal{J}(a)\|^{-1},
\quad a = \tanh(A_{t_K}),
\end{equation}
where $\zeta$ is the standard Gaussian base density for $A_{t_0}$, $\eta_\theta(\cdot)$ is the per-step transition in Section~\ref{app:noise_rollout}, and $\mathcal{J}(a)$ is the Jacobian of the element-wise $\tanh$ squashing. The marginal policy density follows by integrating out the intermediate pre-activations:
\begin{equation}
\label{eq:marginal_pi_from_joint_app}
\pi_\theta(a \mid s) = \int \cdots \int p_c\left(A_{t_0}, \ldots, A_{t_{K-1}}, A_{t_K} = \tanh^{-1}(a) \mid s \right) \, dA_{t_0} \cdots dA_{t_{K-1}}.
\end{equation}
For element-wise $\tanh$, $\|\det \mathcal{J}(a)\|=\prod_{j=1}^{d} (1-a_j^2)^{-1}$.

\subsection{Pathwise expansion of the marginal score}
\label{sec:pathwise_score_expansion}

We derive the gradient of $\mathbb{E}_{a} [ \log \pi_\theta(a \mid s)]$. Using Equ. (\ref{eq:joint_eta_fact_app}):
\begin{align}
&\nabla_\theta \mathbb{E}_{a} [ \log \pi_\theta(a \mid s)] = \nabla_\theta \mathbb{E}_{\mathcal{A}} \left[ \log \pi_\theta(a \mid s) \right] \nonumber \\
&= \nabla_\theta \mathbb{E}_{\mathcal{A}} \left[ \log \left( \int \cdots \int p_c(A_{t_0}, \ldots, A_{t_{K-1}}, A_{t_K} = \tanh^{-1}(a) \mid s) \, dA_{t_0} \cdots dA_{t_{K-1}} \right) \right].
\end{align}
Expanding the inner gradient yields
\begin{align}
\nabla_\theta \log \pi_\theta(a \mid s) &= \frac{1}{\pi_\theta(a \mid s)} \nabla_\theta \int \cdots \int \zeta(A_{t_0}) \left[\prod_{i=0}^{K-1} \eta_\theta \left( A_{t_{i+1}} \mid A_{t_i}, s; \Delta t_i \right) \right] \|\det \mathcal{J}(a)\|^{-1} \, dA_{t_0:K-1} \nonumber \\
&= \frac{1}{\pi_\theta(a \mid s)} \int \cdots \int \zeta(A_{t_0}) \left[\prod_{i=0}^{K-1} \eta_\theta \left( A_{t_{i+1}} \mid A_{t_i}, s; \Delta t_i \right) \right] \nonumber \\
&\quad \cdot \sum_{i=0}^{K-1} \nabla_\theta \log \eta_\theta \left( A_{t_{i+1}} \mid A_{t_i}, s; \Delta t_i \right) \|\det \mathcal{J}(a)\|^{-1} \, dA_{t_0:K-1}.
\end{align}
Therefore,
\begin{equation}
\label{eq:score_pathwise_integral_app}
\nabla_\theta \mathbb{E}_{a} [ \log \pi_\theta(a \mid s)] = \mathbb{E}_{\mathcal{A}} \left[ \sum_{i=0}^{K-1} \nabla_\theta \log \eta_\theta \left( A_{t_{i+1}} \mid A_{t_i}, s; \Delta t_i \right) \right],
\end{equation}
where the Jacobian term does not contribute because it is independent of $\theta$. Since $\eta_\theta$ is Gaussian with mean $m_i = A_{t_i}+b_\theta\,\Delta t_i$ and covariance $\Sigma_i=\sigma_\theta^2\Delta t_i I$, each term is closed form:
\[
\nabla_\theta \log \eta_\theta
=
\frac{1}{\sigma_\theta^2\Delta t_i}
\big(A_{t_{i+1}}-m_i\big)^\top
\frac{\partial m_i}{\partial \theta}
-
\frac{d}{\sigma_\theta}\frac{\partial \sigma_\theta}{\partial \theta}
+\text{higher-order terms if }\sigma_\theta\text{ depends on }\theta.
\]

\subsection{Gradients of the SAC losses under the joint path factorization}
\label{app:sac_grads}

\paragraph{Critic update.}
The target-matching loss is
\begin{equation}
\label{eq:critic_loss_app}
L(\psi) = \left[ Q_\psi(s_h, a_h) - \left( r_h + \gamma Q_{\bar{\psi}}(s_{h+1}, a_{h+1}) - \alpha \log \pi_\theta(a_h \mid s_h) \right) \right]^2,
\end{equation}
where $a_{h+1}\sim\pi_\theta(\cdot\mid s_{h+1})$. Using the joint-path form,
\begin{equation}
\label{eq:critic_grad_app}
\nabla_\psi L(\psi) = 2 \left( Q_\psi(s_h, a_h) - \left( r_h + \gamma Q_{\bar{\psi}}(s_{h+1}, a_{h+1}) - \alpha \log p_c(\mathcal{A} \mid s)\right) \right) \nabla_\psi Q_\psi(s_h, a_h),
\end{equation}
where no gradients flow through $Q_{\bar\psi}$. Replacing the marginal $\log \pi_\theta$ by $\log p_c$ only changes a baseline and has a negligible effect on learning behavior.

\paragraph{Actor update.}
The actor loss is
\begin{equation}
\label{eq:actor_loss_app}
L(\theta) = \alpha \log \pi_\theta(a_h^\theta \mid s_h) - Q_\psi(s_h, a_h^\theta),
\end{equation}
with $a_h^\theta=\tanh(A_{t_K}^\theta)$. Its gradient uses the pathwise form:
\begin{equation}
\label{eq:actor_grad_mixed_app}
\nabla_\theta L(\theta) = \alpha \sum_{i=0}^{K-1} \nabla_\theta \log \eta_\theta\left( A_{t_{i+1}}^\theta \mid A_{t_i}^\theta, s_h; \Delta t_i \right) - \nabla_\theta Q_\psi(s_h, a_h^\theta),
\end{equation}
where the $Q$-term differentiates through $a_h^\theta$.

\subsection{Temperature update (learned $\alpha$)}
\label{app:alpha}

When learning $\alpha$ to match a target entropy $\bar{\mathcal{H}}$:
\begin{equation}
\label{eq:alpha_loss_app}
L(\alpha) = \mathbb{E}_{s_h, a_h^\theta \sim \pi_\theta(\cdot \mid s_h)} \left[ -\alpha \left( \log \pi_\theta(a_h^\theta \mid s_h) + \bar{\mathcal{H}} \right) \right].
\end{equation}
The gradient is
\begin{equation}
\label{eq:alpha_grad_app}
\nabla_\alpha L(\alpha) = -\mathbb{E}_{s_h, a_h^\theta} \left[ \log \pi_\theta(a_h^\theta \mid s_h) + \bar{\mathcal{H}} \right].
\end{equation}
Using the joint-path surrogate yields
\begin{equation}
\label{eq:alpha_grad_joint_app}
\nabla_\alpha L(\alpha) = -\mathbb{E} \left[ \sum_{i=0}^{K-1} \log \eta_\theta\left( A_{t_{i+1}}^\theta \mid A_{t_i}^\theta, s_h; \Delta t_i \right) - \log \|\det \mathcal{J}(a_h^\theta)\| + \bar{\mathcal{H}} \right],
\end{equation}
and we set $\bar{\mathcal{H}}=0$ unless otherwise noted.

\subsection{Practical notes and implementation details}
\label{app:practical}

\textbf{Rollout length and noise.} Use small $K$ (e.g., $4$) to control backprop depth and latency. Fix $\sigma_\theta$ (e.g., $0.10$) or learn a lightweight state head; fixed schedules simplify tuning.

\textbf{Squashing and Jacobian.} Always squash $A_{t_K}\!\mapsto\!a=\tanh(A_{t_K})$ and include the exact Jacobian in $\log p_c$ of Equ. (\ref{eq:joint_eta_fact_app}) to keep the entropy term correct.

\textbf{Targets and normalization.} Maintain a delayed target $Q_{\bar\psi}$ with EMA. Pre-normalization in Flow-T and a mild positive gate bias in Flow-G improve early stability.

\textbf{Gradient flow.} Flow-G gates the residual change to damp gradient amplification; Flow-T uses pre-norm residual blocks. Both act as drop-in $v_\theta$ inside Equ. (\ref{FR}).

\textbf{Offline-to-online.} In the regularized actor loss of the main text (Equ. (\ref{actor_o2o})), choose $\beta$ large early to stay on-replay, then anneal as online data grows. Flow-matching pretraining via Equ. (\ref{flow_matching}) is optional but helpful for sparse rewards.

\textbf{Efficiency.} The entropy term scales linearly in $K$ and action dimension $d$ because it decomposes into per-step Gaussian factors.

\textbf{Reproducibility.} We evaluate $\log p_c$ and its gradient with a single noisy rollout per update; additional variance reduction is possible but not required in our settings.

\section{Extended Related Work}
\label{Related_Work_extended}

 We evaluate our approach against several state-of-the-art methods, categorized into two groups based on their training paradigm. From-scratch algorithms initialize randomly and learn entirely through environment interaction, while offline-to-online methods first pre-train on expert demonstrations before transitioning to online reinforcement learning.

\subsection{From-scratch Training Methods}
The integration of generative models into reinforcement learning has emerged as a prominent research direction, with particular focus on training policies parameterized by diffusion and flow-based models. This line of work addresses the limitations of traditional unimodal policy representations by leveraging the expressive power of generative models to capture complex, multimodal action distributions.

Early efforts in this domain primarily concentrated on diffusion-based policies. Q-Score Matching (QSM) \citep{QSM} pioneered this direction by establishing a theoretical connection between score functions and Q-value gradients, enabling direct policy optimization through score matching objectives. Building upon this foundation, several advanced methods have been proposed: QVPO \citep{QVPO} introduces Q-weighted variational policy optimization for improved sample efficiency; DDiffPG \citep{D_PPO} extends policy gradient methods to diffusion models; MaxEntDP \citep{MaxentDP} incorporates maximum entropy principles; and DIME \citep{DIME} reformulates diffusion policy training through KL divergence minimization between denoising chains and exponentiated critic targets.

More recently, attention has shifted toward flow-based policies, which offer computational advantages over diffusion models through deterministic ODE integration. FlowRL \citep{FlowRL} represents the current state-of-the-art in this category, proposing Wasserstein-2 regularized policy optimization that constrains the learned policy to remain within proximity of optimal behaviors identified in the replay buffer.

For our experimental evaluation, we select DIME and FlowRL as primary benchmarks for diffusion and flow-based approaches, respectively, based on their reported performance improvements over earlier methods such as QVPO and QSM. We additionally include QSM in our comparison as it established many of the foundational concepts underlying subsequent developments in this field. Meanwhile, classical RL training methods for Gaussian policy, including PPO \citep{PPO} and SAC \citep{SAC_arxiv}.

\paragraph{FlowRL \citep{FlowRL}.} This approach directly optimizes flow-based policies using off-policy RL with Wasserstein regularization. The critic $Q_\psi(s,a)$ follows standard SAC updates, minimizing the temporal difference error:
\begin{equation}
L_Q(\psi) = \mathbb{E}_{(s,a,r,s') \sim \mathcal{D}}\left[\left(Q_\psi(s,a) - \left(r + \gamma \mathbb{E}_{a' \sim \pi_\theta}[Q_\psi(s',a')]\right)\right)^2\right].
\end{equation}
The key innovation lies in the actor update, which formulates policy optimization as a constrained problem that maximizes Q-values while regularizing the velocity field $v_\theta$ using a Wasserstein-2 distance constraint. In practice, this is solved using a Lagrangian relaxation:
\begin{equation}
L_\pi(\theta) = \mathbb{E}_{\substack{s,a \sim \mathcal{D}, a' \sim \pi_\theta \\ t \sim U(0,1)}} \left[f(Q_{\pi_{\beta^*}}(s,a) - Q_\psi(s,a'))\|v_\theta(s,A_t,t) - (a - a_0)\|^2\right],
\end{equation}
where $f(\cdot)$ is a non-negative weighting function, $A_t = (1-t)a_0 + ta$ represents the flow interpolation path, and $\pi_{\beta^*}$ denotes the optimal behavior policy derived from the replay buffer. The constraint adaptively regularizes the policy toward high-performing behaviors when $Q_{\pi_{\beta^*}} > Q_\psi$, effectively aligning the flow optimization with value-based policy improvement.

\paragraph{DIME \citep{DIME}.} This method treats diffusion policies as exponential family distributions and optimizes them via KL divergence minimization. The critic update remains standard:
\begin{equation}
L_Q(\psi) = \frac{1}{2}\mathbb{E}\left[\left(Q_\psi(s_t,a_t)-Q_{\text{target}}(s_t,a_t)\right)^2\right].
\end{equation}
The actor update is more sophisticated, defining a target marginal through the exponentiated critic $\bar{\pi}_0(a|s) = \exp(Q_\psi(s,a))/Z_\psi(s)$ and minimizing the KL divergence between the denoising chain and this target:
\begin{equation}
L(\theta) = \mathbb{E}_{\pi_\theta}\left[\log \pi_\theta(a_N|s) - Q_\psi(s,a_0) + \sum_{n=1}^{N}\log \frac{\pi_\theta(a_{n-1}|a_n,s)}{\bar{\pi}(a_n|a_{n-1},s)}\right].
\end{equation}

\paragraph{QSM (Q-Score Matching) \citep{QSM}} This approach leverages score matching to align the policy's score function with the action gradient of the Q-function, providing a principled connection between value-based and score-based learning. The critic follows a double Q-learning update with target networks for stability:
\begin{equation}
L_Q(\theta) = \mathbb{E}_{(s_t, a_t, r_{t+1}, s_{t+1}) \sim \mathcal{B}}\left[\left(Q_\theta(s_t, a_t) - \left(r_{t+1} + \gamma \min_{i=1,2} Q_{\theta_i'}(s_{t+1}, a_{t+1})\right)\right)^2\right],
\end{equation}
where $Q_{\theta_i'}$ denotes the target networks. The actor update represents the core innovation, training a score function $\Psi_\phi(s_t, a_t)$ to match the scaled action gradient of the Q-function:
\begin{equation}
L_\pi(\phi) = \mathbb{E}_{(s_t, a_t) \sim \mathcal{B}}\left[\left\|\Psi_\phi(s_t, a_t) - \alpha \nabla_a Q_\theta(s_t, a_t)\right\|^2\right],
\end{equation}
where $\alpha$ controls the alignment strength. This formulation enables the policy to naturally follow the Q-function's action gradients, providing implicit policy improvement without explicit action sampling.

\subsection{Offline-to-Online Training Methods}
While from-scratch training is viable for many reinforcement learning tasks, it often struggles with sample efficiency in complex environments, particularly those with dense rewards. To address this limitation, the offline-to-online paradigm has become a prominent approach. This strategy involves two stages: first, pre-training a policy on an offline dataset of expert behaviors, and second, fine-tuning this policy through online interaction with the environment.

This paradigm was initially explored with diffusion-based policies, leading to the development of methods such as DPPO \citep{DPPO}, D3P \citep{D3PO}, Resip \citep{Resip}, and PA-RL \citep{PA-RL}. More recently, research has extended this approach to flow-based policies, which are the focus of our work.

Within the flow-policy literature, methods can be categorized by their online fine-tuning algorithm. For on-policy fine-tuning, ReinFlow \citep{ReinFlow} stands out by successfully adapting a pre-trained flow-based policy using the PPO algorithm. For off-policy fine-tuning, FQL \citep{FlowQ} and its successor QC-FQL \citep{QC-FQL} are state-of-the-art. However, a crucial characteristic of these off-policy methods is their reliance on an auxiliary, distilled policy for online updates; they do not directly fine-tune the original flow model. Instead, they distill knowledge from the pre-trained flow-based policy into a simpler, unimodal policy that is more amenable to traditional off-policy RL.

For our experiments, we select ReinFlow, FQL, and QC-FQL as benchmarks. Our evaluation primarily concentrates on the off-policy methods to demonstrate the effectiveness of our proposed direct fine-tuning approach for flow-based policies.

\paragraph{QC-FQL \citep{QC-FQL}} This approach employs a three-network architecture: a critic $Q_\theta$, a one-step noise-conditioned policy $\mu_\psi$, and a behavior flow-based policy $f_\xi$. The method extends FQL to handle action chunking by operating in temporally extended action spaces. The critic processes action sequences and is updated via:
\begin{equation}
L_Q(\theta_k) = \left(Q_{\theta_k}(s_t, a_t,\ldots,a_{t+h-1}) - r_t^h - \frac{1}{K}\sum_{k'=1}^K Q_{\bar{\theta}_{k'}}(s_{t+h}, a_{t+h},\ldots,a_{t+2h-1})\right)^2,
\end{equation}
where $r_t^h$ represents the cumulative discounted reward over the action chunk horizon. The one-step policy is trained to maximize Q-values while maintaining proximity to the behavior policy outputs:
\begin{equation}
L_\pi(\psi) = -Q_\theta(s_t, \mu_\psi(s_t,z_t)) + \alpha\left\|\mu_\psi(s_t,z_t) - \left[a_t^\xi \cdots a_{t+h-1}^\xi\right]\right\|_2^2.
\end{equation}

\paragraph{FQL \citep{FlowQ}.} This method represents a simplified version of QC-FQL with unit action chunks ($h=1$). The critic follows standard Bellman updates while the actor combines value maximization with distillation regularization. The actor loss explicitly balances Q-value optimization against behavioral constraints:
\begin{equation}
L_\pi(\omega) = \mathbb{E}_{s \sim \mathcal{D}, a^\pi \sim \mu_\omega}[-Q_\phi(s,a^\pi)] + \alpha \mathbb{E}_{s \sim \mathcal{D}, z \sim \mathcal{N}(0,I)}\left[\|\mu_\omega(s,z) - \mu_\theta(s,z)\|_2^2\right],
\end{equation}
where $\mu_\theta$ represents a pre-trained behavioral clone used for regularization.

\paragraph{ReinFlow \citep{ReinFlow}} This approach augments flow-based policies with noise injection networks to enable efficient likelihood computation during policy gradient updates. Following a warm-up phase for critic training, the method jointly optimizes the flow-based policy $\pi_\theta$ and noise injection network $\sigma_{\theta'}$ through:
\begin{equation}
L_\pi(\theta,\theta') = \mathbb{E}\left[-A_{\theta_{\text{old}}}(s,a) \sum_{k=0}^{K-1} \log \pi_\theta(a_{k+1}|a_k,s) + \alpha \cdot R(a,s;\theta,\theta_{\text{old}})\right],
\end{equation}
where $A_{\theta_{\text{old}}}(s,a)$ denotes advantage estimates and $R(\cdot)$ provides regularization to prevent excessive deviation from the previous policy.

\paragraph{Key Distinctions.} Unlike these baseline approaches, our method enables direct training the flow-based policy via SAC (off-policy methods) without requiring auxiliary distillation actors, surrogate objectives, or complex multi-network architectures. The Flow-G and Flow-T parameterizations provide gradient stability while maintaining the expressive power of the original flow-based policy throughout training.

\section{Experimental Domain}
\label{sec:Env_settings}
\begin{figure}[htbp]
	\centering
	\begin{subfigure}{0.24\linewidth}
		\centering
		\includegraphics[width=1\linewidth]{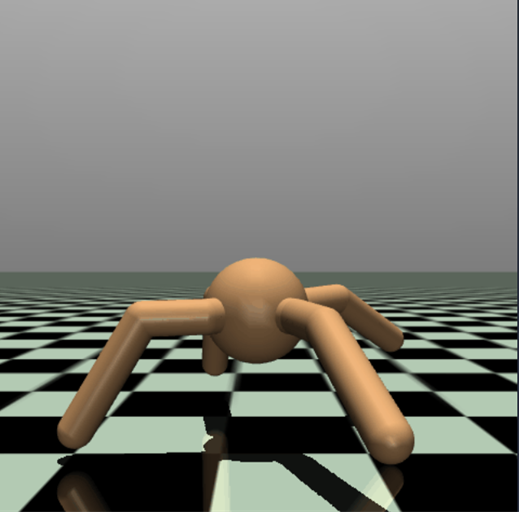}
		\caption{Ant}
		\label{fig:Ant}
	\end{subfigure}
	\hfill
	\begin{subfigure}{0.24\linewidth}
		\centering
		\includegraphics[width=1\linewidth]{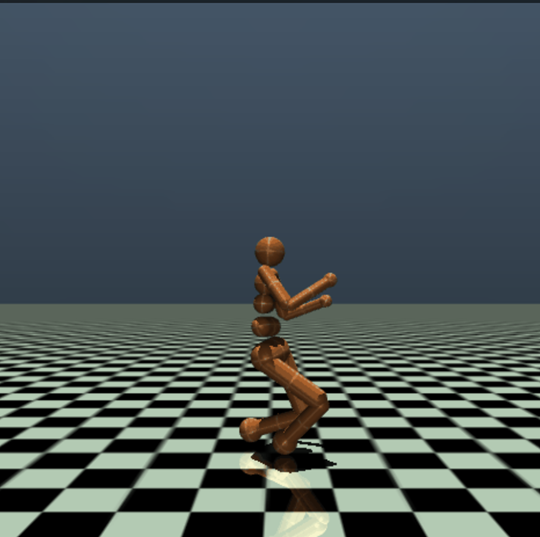}
		\caption{Humanoid}
		\label{fig:Humanoid}
	\end{subfigure}
	\hfill
	\begin{subfigure}{0.23\linewidth}
		\centering
		\includegraphics[width=0.98\linewidth]{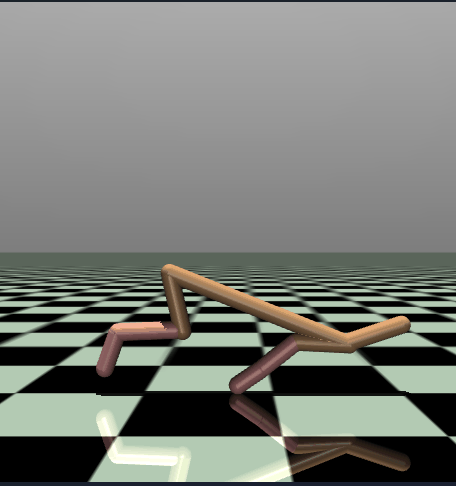}
		\caption{HalfCheetah}
		\label{fig:HalfCheetah}
	\end{subfigure}
	\hfill
	\begin{subfigure}{0.24\linewidth}
		\centering
		\includegraphics[width=1\linewidth]{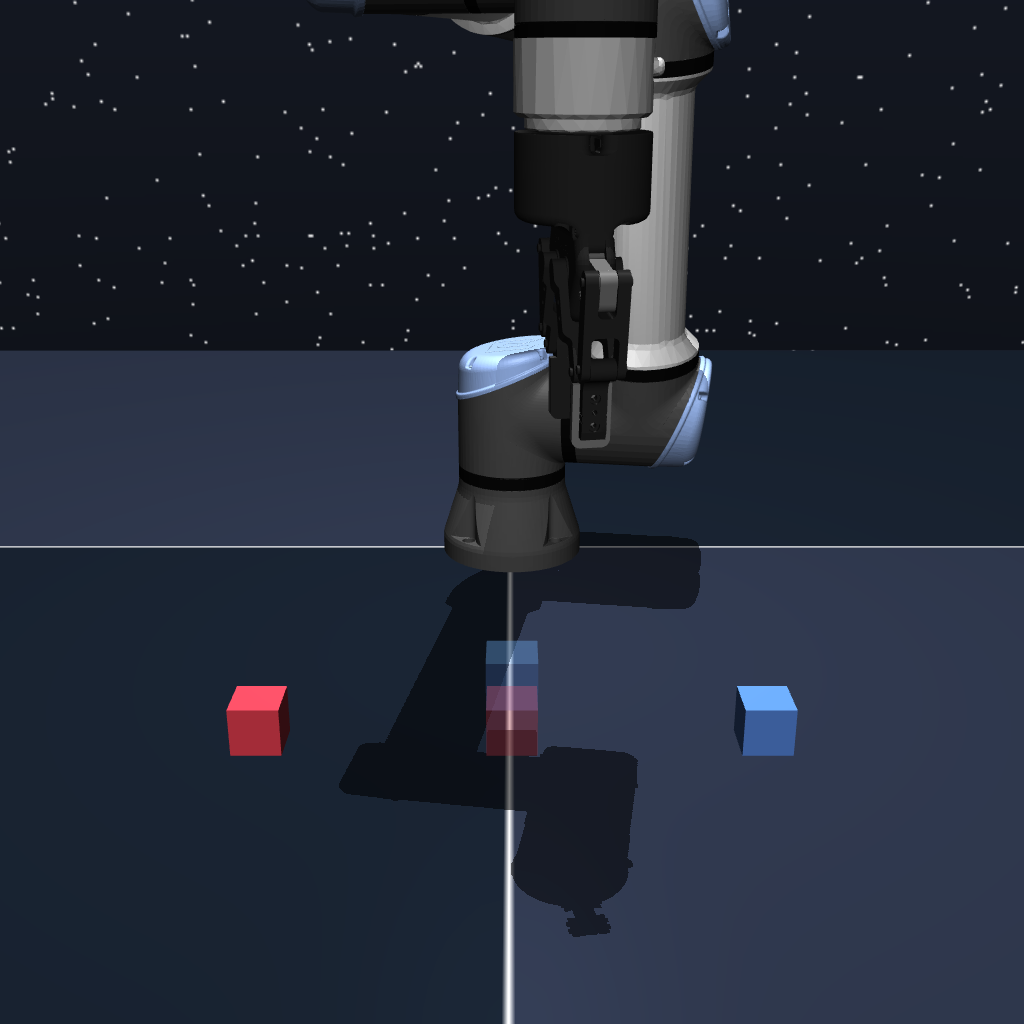}
		\caption{cube-double}
		\label{fig:cube-double}
	\end{subfigure}
    \hfill
    \vspace{1ex}
    \begin{subfigure}{0.23\linewidth}
		\centering
		\includegraphics[width=1\linewidth]{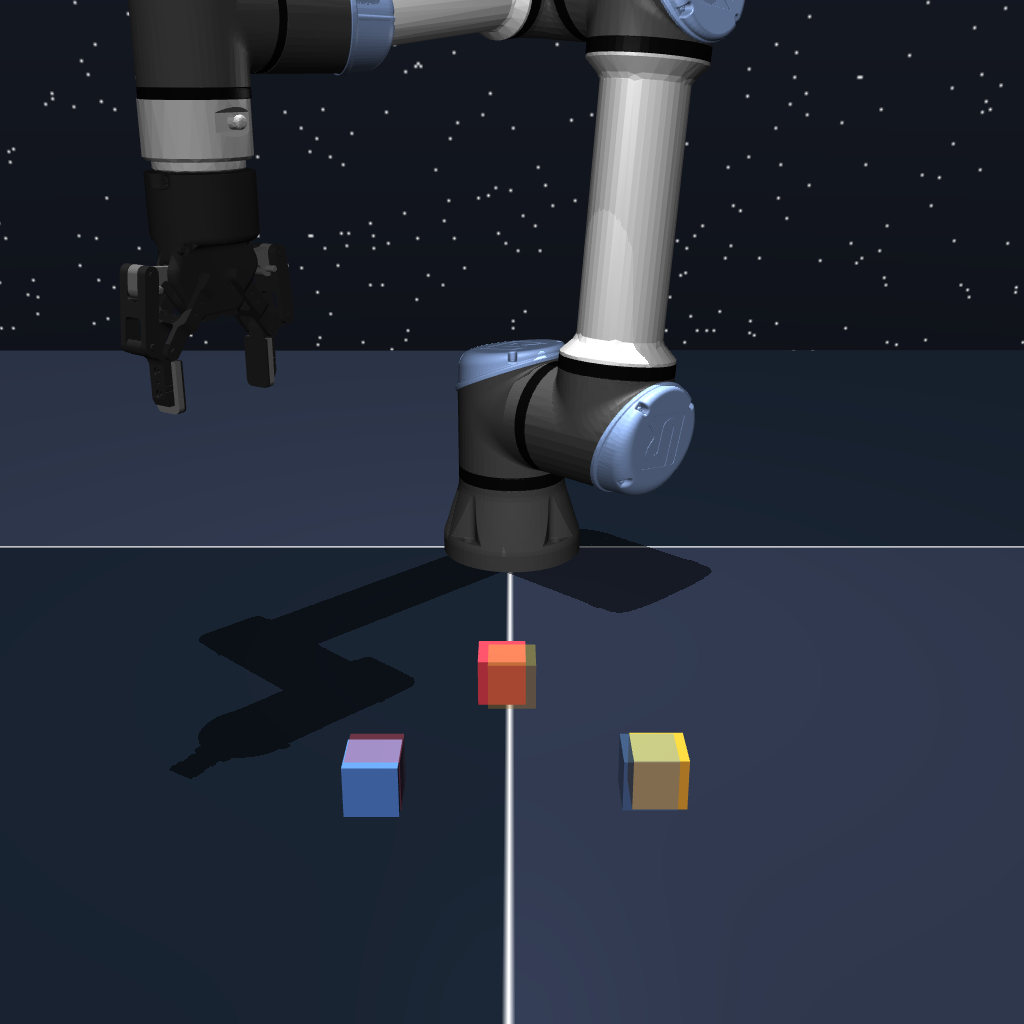}
		\caption{cube-triple}
		\label{fig:cube-triple}
	\end{subfigure}
	\hfill
	\begin{subfigure}{0.23\linewidth}
		\centering
		\includegraphics[width=1\linewidth]{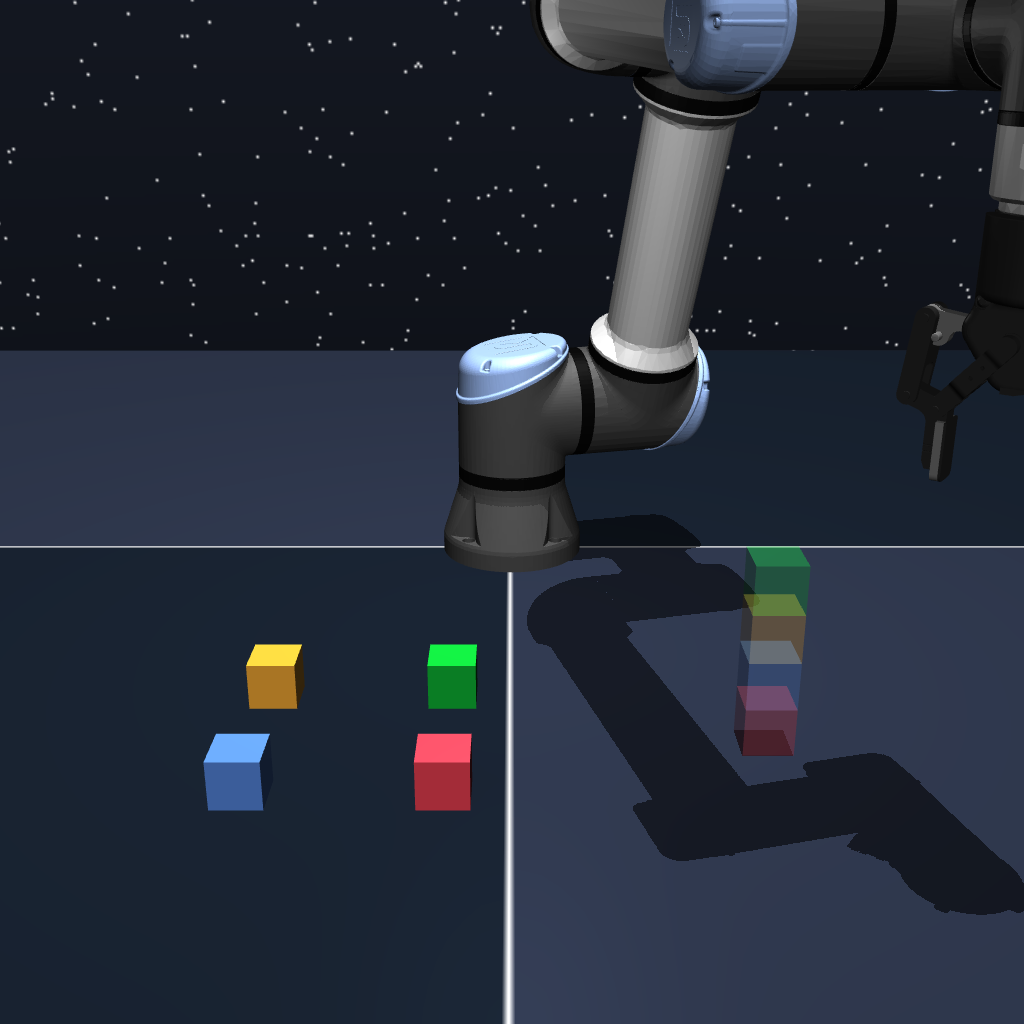}
		\caption{cube-quadruple}
		\label{fig:cube-quadruple}
	\end{subfigure}
	\hfill
	\begin{subfigure}{0.23\linewidth}
		\centering
		\includegraphics[width=1\linewidth]{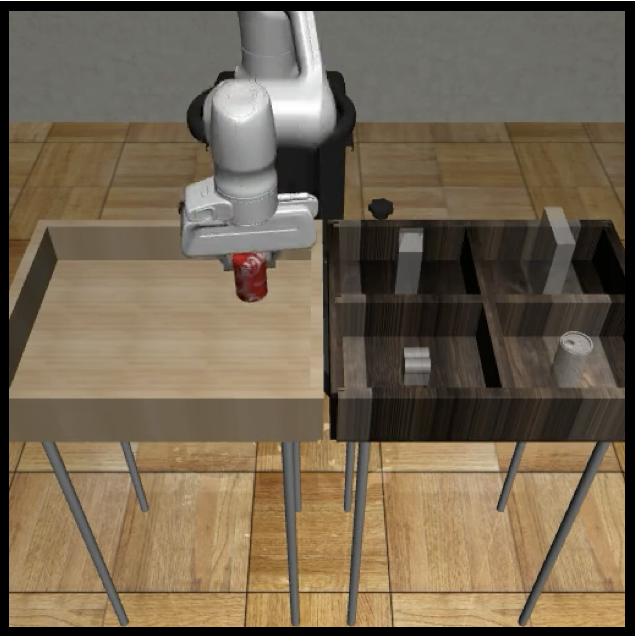}
		\caption{can}
		\label{fig:can}
	\end{subfigure}
    \hfill
    \begin{subfigure}{0.24\linewidth}
		\centering
		\includegraphics[width=1\linewidth]{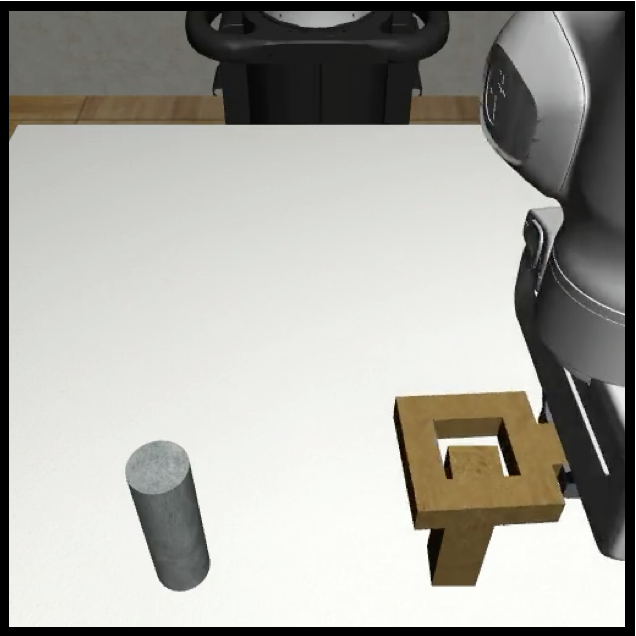}
		\caption{square}
		\label{fig:square}
	\end{subfigure}
	\caption{Visualizations of the diverse simulation environments used for evaluation. Subfigures (a-c) show the \textbf{MuJoCo} locomotion tasks. Subfigures (d-f) depict the complex, sparse-reward manipulation tasks from \textbf{OGBench}. Subfigures (g-h) illustrate the demonstration-based tasks from \textbf{Robomimic}. This selection provides a comprehensive testbed for evaluating both from-scratch learning and offline-to-online fine-tuning.}
	\label{fig:All_envs}
\end{figure}
To comprehensively evaluate our method, we conduct experiments across a diverse suite of simulated environments. We utilize the classic \textbf{MuJoCo} benchmark \citep{Mujoco} for standard from-scratch reinforcement learning. To assess performance in the more challenging offline-to-online setting, particularly with sparse rewards, we employ complex manipulation tasks from \textbf{OGBench} \citep{Ogbench} and human-demonstration-based tasks from \textbf{Robomimic} \citep{Robomimic}. Visualizations of these environments are presented in Fig. \ref{fig:All_envs}.

\subsection{MuJoCo Environments}
We evaluate our method on six standard continuous control tasks from the \textbf{MuJoCo} physics simulation benchmark~\citep{Mujoco}: \texttt{Hopper-v4}, \texttt{Walker2d-v4}, \texttt{HalfCheetah-v4}, \texttt{Ant-v4}, \texttt{Humanoid-v4}, and \texttt{HumanoidStandup-v4}. These environments feature simulated robots with varying degrees of complexity, where the primary objective is to learn a locomotion policy that maximizes forward velocity without falling. They serve as a standard measure of performance for from-scratch RL algorithms.

\subsection{OGBench Environments}
From \textbf{OGBench} \citep{Ogbench}, we select four challenging manipulation domains using their publicly available single-task versions. The selected domains include \texttt{cube-double/triple/quadruple} tasks. In the cube tasks, an agent must control a UR-5 arm to place multiple objects in target locations, receiving a reward of $-n_{\text{wrong}}$, where $n_{\text{wrong}}$ is the number of incorrectly placed cubes. The cube-triple and cube-quadruple tasks are particularly difficult to solve from offline data alone, providing a rigorous testbed for the sample efficiency of offline-to-online algorithms. In the offline phase, we use the official 100M-size dataset\footnote{\url{https://github.com/seohongpark/ogbench?tab=readme-ov-file}}.

\subsection{Robomimic Environments}
We use three robotic manipulation tasks from the \textbf{Robomimic} benchmark \citep{Robomimic}, utilizing the multi-human datasets which contain 300 successful demonstration trajectories per task. The tasks are selected to represent a range of difficulties: \texttt{Lift}, a simple pick-and-place task involving a cube; \texttt{Can}, an intermediate task requiring placing a can into a bin; and \texttt{Square}, the most challenging task, which requires the precise insertion of a square nut onto a peg. We use the official Multi-Human (MH) dataset, containing 300 mixed trajectories per task, for offline pre-training.

\begin{table}[htbp]
\centering
\setlength{\tabcolsep}{3pt}
\caption{Details of the experimental environments. The tasks span classic continuous control with dense rewards (MuJoCo), complex manipulation with sparse rewards (OGBench), and challenging imitation-based tasks also framed with sparse rewards (Robomimic). This selection provides a comprehensive benchmark with diverse state spaces, action dimensions, and reward structures. We use the same dataset configuration in \citep{QC-FQL}.}
\label{tab:env_details}
\begin{tabular}{lcccc}
\toprule
\textbf{Tasks} & \textbf{Reward Type} & \textbf{Dataset Size} & \textbf{Episode Length} & \textbf{Action Dimension} \\
\midrule
\multicolumn{5}{l}{\textit{MuJoCo}} \\
\midrule
Hopper-v4          & Dense   & /   & 1000 & 3  \\
Walker2d-v4        & Dense   & /   & 1000 & 6  \\
HalfCheetah-v4     & Dense   & /   & 1000 & 6  \\
Ant-v4             & Dense   & /   & 1000 & 8  \\
Humanoid-v4        & Dense   & /   & 1000 & 17 \\
HumanoidStandup-v4        & Dense   & /   & 1000 & 17 \\
\midrule
\multicolumn{5}{l}{\textit{OGBench}} \\
\midrule
cube-double & Sparse  & 1M   & 500  & 5  \\
cube-triple & Sparse  & 3M   & 1000 & 5  \\
cube-quadruple-100M & Sparse & 100M & 1000 & 5  \\
\midrule
\multicolumn{5}{l}{\textit{Robomimic}} \\
\midrule
lift               & Sparse  & 31,127  & 500  & 7  \\
can                & Sparse  & 62,756  & 500  & 7  \\
square             & Sparse  & 80,731  & 500  & 7  \\
\bottomrule
\end{tabular}
\end{table}

\section{Implementation Details for Experiments}
\label{sec:imp}
In this section, we introduce the implementation details of the hyperparameter setting and network structures. We first begin with the from-scratch training:
\subsection{From-Scratch training setting}
In from-scratch training, we develop our algorithm based on CleanRL \citep{CleanRL}, which is a widely used benchmark codebase, where we also use the same implementation of PPO, SAC in it. For FlowRL and DIME, we use the 
official implementation except for unifying the parameter quantity. We run 5 seeds for all experiments and all plots use a 95\% confidenceinterval.

\begin{table}[h!]
\centering
\caption{Common Hyperparameters for Algorithms}
\label{tab:hyperparameters}
\begin{tabular}{ll}
\toprule
\textbf{Parameter} & \textbf{Value} \\
\midrule
Optimizer & Adam \\
& \quad ($b_1=0.5$ for Flow-based approaches) \\
Batch size ($M$) & 512 \\
Replay buffer size & $1 \times 10^6$ \\
Discount factor ($\gamma$) & 0.99 \\
Policy learning rate & $3 \times 10^{-4}$ \\
Critic learning rate & $1 \times 10^{-3}$ \\
Target network update rate ($\tau$) & 0.005 for \textbf{SAC} \\
& 1.0 for \textbf{Flow}, \textbf{Flow-G}, \textbf{Flow-T} \\
Learning starts & 50,000 \\
Entropy coefficient ($\alpha$) & 0.2 (initial value) \\
Target entropy  & $-dim(\mathcal{A})$ for \textbf{SAC}, 0 for \textbf{Flow}, \textbf{Flow-G}, \textbf{Flow-T} \\
Automatic entropy tuning & True \\
Number of online environment steps & $1 \times 10^6$ \\
\bottomrule
\end{tabular}
\end{table}

\begin{figure}[htbp]
    \centering
    \includegraphics[width=\linewidth]{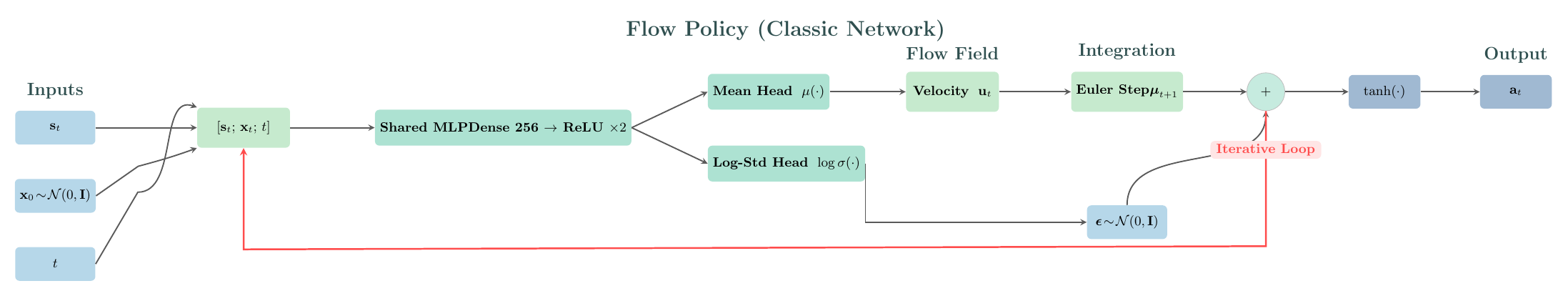}
    \caption{The flow-based policy designed with classic networks. The velocity is modeled with an arbitrary network; here, we use an MLP as the representative. The whole flow rollout corresponds to the recurrent computation of the residual RNN.}
    \label{Flow_policy_s}
\end{figure}

\begin{figure}[htbp]
    \centering
    \includegraphics[width=\linewidth]{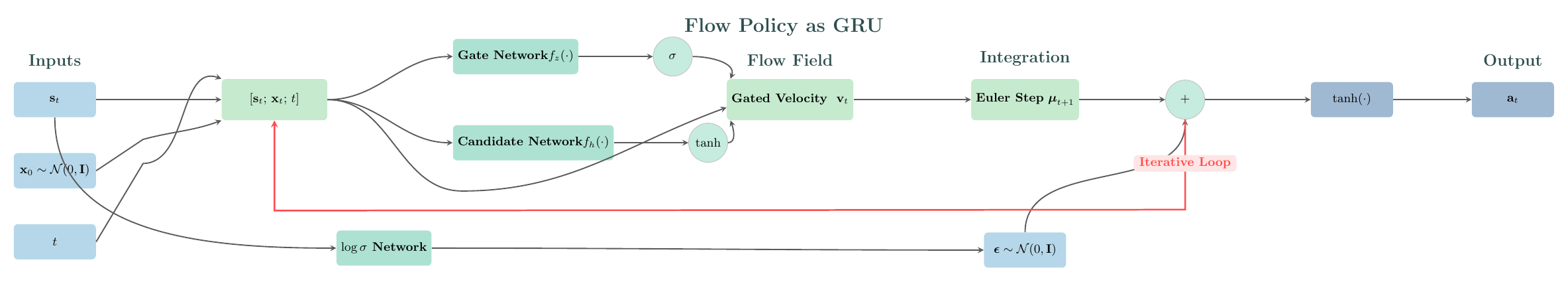}
    \caption{The flow-based policy designed with Gated velocity. The velocity is modeled with both the gate network and the candidate network. The whole flow rollout corresponds to the recurrent computation of GRU.}
    \label{Flow_policy_G}
\end{figure}

\begin{figure}[htbp]
    \centering
    \includegraphics[width=\linewidth]{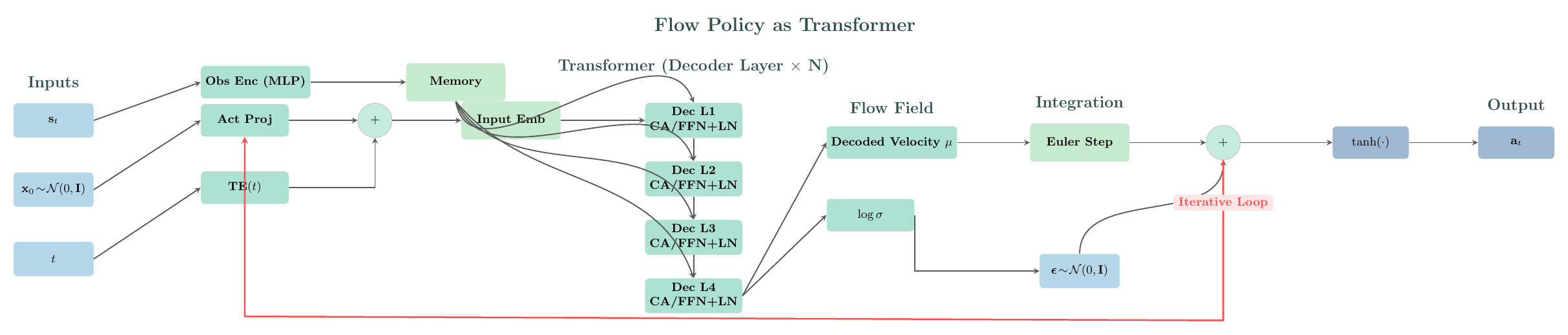}
    \caption{A schematic .}
    \label{Flow_policy_T}
\end{figure}

\paragraph{Architectures of the velocity network in flow-based policies (Figs.~\ref{Flow_policy_s}--\ref{Flow_policy_T}).}
We detail the network parameterizations for the velocity field $v_\theta$ used inside the flow rollout in Equ. (\ref{FR}). 
Across all variants, the policy starts from a state-conditioned base $A_{t_0}\!\sim\!\mathcal{N}(0,I_d)$, performs $K$ Euler updates
$A_{t_{i+1}}=A_{t_i}+\Delta t_i\,v_\theta(t_i,A_{t_i},s)$, and then applies $\tanh$ squashing to obtain the final action $a=\tanh(A_{t_K})$.
During training, we optionally pair $v_\theta$ with a log-standard-deviation head to define the per-step Gaussian transition factors used by our noisy/likelihood-friendly rollout.
\paragraph{Fig. ~\ref{Flow_policy_s}: Classic (MLP) velocity network.}
The baseline flow-based policy instantiates $v_\theta$ with a feed-forward network that is conditioned on the current intermediate action $A_{t_i}$, the environment state $s$, and the normalized time index $t_i$.
Concretely, the input token is the concatenation $[\;s;\,A_{t_i};\,t_i\;]$, followed by a shared MLP trunk and two small heads:
(i) a mean head $\mu_\theta(\cdot)$ that produces the deterministic velocity
\[
v_\theta(t_i,A_{t_i},s)=\mu_\theta\big([s;A_{t_i};t_i]\big),
\]
and (ii) a log-standard-deviation head $\log\sigma_\theta(\cdot)$ that parameterizes the per-step transition variance when we use the noisy rollout for likelihood-based training. 
Plugging this $v_\theta$ into Equ. (\ref{FR}) yields the standard residual update
$A_{t_{i+1}} = A_{t_i} + \Delta t_i\,\mu_\theta([s;A_{t_i};t_i])$.
Algebraically, this is a residual RNN step with residual function $f_\theta(\cdot)=\Delta t_i\,\mu_\theta(\cdot)$, matching our sequence-model view in Equ. (\ref{eq:rnn_residual}).

\paragraph{Fig. ~\ref{Flow_policy_G}: Gated velocity (\textbf{Flow-G}).}
To stabilize gradients across the $K$ sampling steps, we replace the plain MLP velocity with a GRU-style gated update. 
Let $f_z$ (gate network) and $f_h$ (candidate network) be two MLPs taking $[s;A_{t_i};t_i]$ as input.
Define the update gate and the candidate as
\[
g_i \;=\; \sigma\!\left(f_z([s;A_{t_i};t_i])\right), 
\qquad 
\hat{v}_\theta \;=\; \phi\!\left(f_h([s;A_{t_i};t_i])\right),
\]
where $\sigma(\cdot)$ is the logistic sigmoid and $\phi(\cdot)$ is a bounded activation (e.g., $\tanh$).
The gated velocity is then
\[
v_\theta(t_i,A_{t_i},s) \;=\; g_i \odot \big(\hat{v}_\theta - A_{t_i}\big),
\]
which, when inserted into Equ. (\ref{FR}), yields the GRU-like residual step
\[
A_{t_{i+1}} \;=\; A_{t_i} \;+\; \Delta t_i \,\big(g_i \odot (\hat{v}_\theta - A_{t_i})\big),
\]
exactly as in Equ. (\ref{eq:gru_flow}). 
Intuitively, $g_i$ interpolates between ``keeping'' the current intermediate action ($g_i\!\approx\!0$) and ``rewriting'' it by the candidate proposal ($g_i\!\approx\!1$).
As in Fig.~\ref{Flow_policy_s}, we also include a $\log\sigma_\theta(\cdot)$ head for the per-step Gaussian factors used by the noisy rollout.

\paragraph{Fig. ~\ref{Flow_policy_T}: Transformer-decoder velocity (\textbf{Flow-T}).}
Here we implement $v_\theta$ with a Transformer-style, pre-norm residual block that conditions on the state through cross-attention. 
We first form separate embeddings for the action-time token and the state:
\[
\Phi_{A_i} \;=\; E_A\big(\phi_t(t_i),\,A_{t_i}\big), 
\qquad 
\Phi_S \;=\; E_S\big(\phi_s(s)\big),
\]
as in Equ. (\ref{eq:tf_embeddings}), where $E_A,E_S$ are linear projections and $\phi_t,\phi_s$ are positional/feature encoders.
We stack $L$ ($L=4$ in this figure) pre-norm decoder blocks.
In each layer $l\!=\!1,\dots,L$, the action token is refined by a \emph{state-only} cross-attention and a feed-forward network (no token-to-token mixing across time positions):
\[
Y_i^{(l)} \;=\; \Phi_{A_i}^{(l-1)} 
\;+\;
\mathrm{Cross}_l\!\left(\mathrm{LN}\big(\Phi_{A_i}^{(l-1)}\big),\,\text{context}=\mathrm{LN}(\Phi_S)\right),
\qquad
\Phi_{A_i}^{(l)} \;=\; Y_i^{(l)} \;+\; \mathrm{FFN}_l\!\big(\mathrm{LN}(Y_i^{(l)})\big),
\]
as Equ. (\ref{eq:tf_cross_attn}).
Finally, the decoded token is projected to the velocity space
\[
v_\theta(t_i,A_{t_i},s) \;=\; W_o\!\big(\mathrm{LN}(\Phi_{A_i}^{(L)})\big),
\]
and the rollout step follows Equ. (\ref{FR}):
\[
A_{t_{i+1}} \;=\; A_{t_i} \;+\; \Delta t_i\, W_o\!\big(\mathrm{LN}(\Phi_{A_i}^{(L)})\big),
\]
which matches Equ. (\ref{eq:tf_ffn_velocity}). 
As in the other variants, a parallel $\log\sigma_\theta(\cdot)$ head provides per-step variances for the Gaussian transition factors.

\paragraph{Takeaway: mapping to sequential models.}
Under our sequence-model perspective, the classic MLP velocity in Fig.~\ref{Flow_policy_s} realizes a \emph{residual RNN} step in Equ. (\ref{eq:rnn_residual}), 
the gated velocity in Fig.~\ref{Flow_policy_G} realizes a \emph{GRU}-style residual update in Equ. (\ref{eq:gru_flow}), 
and the decoded velocity in Fig.~\ref{Flow_policy_T} realizes a \emph{Transformer Decoder} refinement in Equ. (\ref{eq:tf_embeddings})–(\ref{eq:tf_ffn_velocity}).
All three are drop-in parameterizations of $v_\theta$ inside the same flow rollout in Equ. (\ref{FR}), differing only in how they regulate and condition information flow across rollout steps.

\begin{table}[t]
\centering
\scriptsize
\setlength{\tabcolsep}{3pt}
\renewcommand{\arraystretch}{1.18}
\caption{Actor (velocity) architectures inside the $K$-step flow rollout $A_{t_{i+1}}=A_{t_i}+\Delta t\,v_\theta(t_i,A_{t_i},s)$. All variants apply $\tanh$ squashing with Jacobian correction. Notation: $d_a{:=}|\mathcal A|$, Transformer $d{=}64$, heads $n_H{=}4$, layers $n_L{=}2$.}
\label{tab:actor_only_compact}
\begin{tabular}{@{}cccc@{}}
\toprule
\textbf{Aspect} & \textbf{Classic (MLP)} & \textbf{Flow-G (GRU-gated)} & \textbf{Flow-T (Transformer-decoder)} \\
\midrule
Conditioning input &
\shortstack{$[s;\,A_{t_i};\,t_i]$} &
\shortstack{$[s;\,A_{t_i};\,\text{time\_emb}(t_i)]$} &
\shortstack{$A_{t_i}$ token $+$ time emb\\[1pt] state $s$ as memory} \\
\cmidrule(lr){1-4}
Backbone / blocks &
\shortstack{MLP $256\!\to\!256$\\[1pt]\texttt{ReLU}} &
\shortstack{Gate: $128\!\to\!d_a$ (\texttt{swish})\\[1pt]Cand: $128\!\to\!d_a$ (\texttt{swish})} &
\shortstack{Decoder $\times n_L{=}2$\\[1pt]self-only SA, cross-attn$(s)$, FFN $4d$, LN} \\
\cmidrule(lr){1-4}
Velocity form &
\shortstack{$v_\theta{=}\mu_\theta([s;A_{t_i};t_i])$\\[1pt]($\mu_\theta\!\in\!\mathbb R^{d_a}$)} &
\shortstack{$g_i{=}\sigma(f_z),\;\hat v{=}50\,\tanh(f_h)$\\[1pt]$v_\theta{=}g_i\odot(\hat v{-}A_{t_i})$} &
\shortstack{$z_i{=}W_o(\mathrm{LN}(\Phi_{A_i}^{(L)}))$\\[1pt]$v_\theta{=}z_i$} \\
\cmidrule(lr){1-4}
Log-std clamp &
\shortstack{$\tanh$ to\\[1pt]$[\!-5,2]$} &
\shortstack{$\tanh$ to\\[1pt]$[\!-5,2]$} &
\shortstack{$\tanh$ to\\[1pt]$[\!-5,2]$} \\
\cmidrule(lr){1-4}
Action sampling steps $K$ &
\shortstack{$4$} &
\shortstack{$4$} &
\shortstack{$4$} \\
\cmidrule(lr){1-4}
Notable inits / dims &
\shortstack{--} &
\shortstack{Gate head init: $W{=}0$, $b{=}5.0$\\[1pt]hidden $128$} &
\shortstack{$d{=}64$, $n_H{=}4$, $n_L{=}2$\\[1pt]obs-enc $32\!\xrightarrow{\texttt{silu}}\!64$} \\
\cmidrule(lr){1-4}
Per-step update &
\shortstack{$A\!\leftarrow\!A+v\,\Delta t$\\[1pt]$A\!\leftarrow\!\mathcal N(A,\sigma^2)$} &
\shortstack{Same as Classic} &
\shortstack{Same as Classic} \\
\bottomrule
\end{tabular}
\end{table}

\subsection{Offline-to-online training setting}
The network design in offline-to-online training is similar to the from-scratch training. Recall the actor loss:
\begin{equation}
\nonumber
L(\theta) =\alpha \log p_c(\mathcal{A}^\theta\mid s_h)-Q_\psi\!\left(s_h, a_h^\theta\right)+
\beta\,\|a_h^\theta-a_h\|^2,\quad (a_h,s_h)\sim\mathcal{B}.
\end{equation}
It is observed that the setting of hyper-parameter $\beta$ highly influences the training, where the regularization decides whether the optimized policy stays close or not to the policy in the buffer. We basically adopt the same setting of $\beta$ as \citep{QC-FQL}, where we detail in the Tabll \ref{tab:beta_alg}:

\begin{table}[htbp]
\centering
\caption{A comparison of the regularization parameter $\beta$ across different environments and algorithms. The notation $a/b$ specifies the value of the regularization parameter $\beta$ for the offline learning phase ($a$) and the subsequent online learning phase ($b$). For instance, $10000/1000$ indicates that $\beta=10000$ is used for offline training and $\beta=1000$ for online training.}
\label{tab:beta_alg}
\begin{tabular}{lrrrr}
\toprule
\textbf{Environments} & \textbf{FQL} & \textbf{QC-FQL} & \textbf{Flow-G} & \textbf{Flow-T} \\
\midrule
cube-double-* & 300   & 300   & 300   & 300   \\
cube-triple-* & 300   & 100   & 100   & 100   \\
cube-quadruple-100M-* & 300   & 100   & 100   & 100   \\
lift                  & 10000 & 10000 & 10000/1000 & 10000/1000 \\
can                   & 10000 & 10000 & 10000/1000 & 10000/1000 \\
square                & 10000 & 10000 & 10000/1000 & 10000/1000 \\
\bottomrule
\end{tabular}
\end{table}

Table \ref{tab:offline_online_actor} summarizes the actor-side architectures and hyperparameters for our offline-to-online variants. We adopt action chunking (horizon $H$), which has been shown to be effective on complex tasks \citep{QC-FQL}. The parameter counts of Flow-G and Flow-T are less than or comparable to that of QC-FQL. We also use fewer denoising/sampling steps $K$ than QC-FQL to improve efficiency without degrading training quality. For stability, we set the SAC target entropy to $0$ and employ a fixed sampling noise level—contrary to our from-scratch setting, where a separate network adaptively tunes the noise schedule.

\begin{table}[t]
\centering
\scriptsize
\setlength{\tabcolsep}{3pt}
\renewcommand{\arraystretch}{1.18}
\caption{Offline-to-online settings and actor-specific hyperparameters.}
\label{tab:offline_online_actor}
\begin{tabular}{@{}C{2.2cm}C{2.55cm}C{2.55cm}C{2.55cm}@{}}
\toprule
\textbf{Aspect} & \textbf{QC-FQL} & \textbf{Flow-G } & \textbf{Flow-T} \\
\midrule
Actor backbone &
\shortstack{MLP $(512{\times}4)$\\GELU, no LN} &
\shortstack{MLP $(512{\times}4)$ + gate\\($h{=}256$, \texttt{swish})} &
\shortstack{Decoder $n_L{=}2$, $d{=}128$\\$n_H{=}4$, FFN $4d$} \\
\cmidrule(lr){1-4}
Velocity form &
\shortstack{$v_\theta(s,a,t)$ by MLP} &
\shortstack{$v{=}z\odot(50\tanh(\hat v)-a)$\\$z{=}\sigma(f_z)$} &
\shortstack{$v$ from decoder head\\(self+cross attn)} \\
\cmidrule(lr){1-4}

Flow / steps &
\shortstack{Action sampling steps\\$K{=}10$} &
\shortstack{Action sampling steps\\$K{=}4$ } &
\shortstack{Action sampling steps\\$K{=}4$ } \\
\cmidrule(lr){1-4}
Sampling noise std &
\shortstack{deterministic} &
\shortstack{0.10} &
\shortstack{0.10} \\
\cmidrule(lr){1-4}
SAC entropy ($\alpha$) &
\shortstack{N/A (no SAC)} &
\shortstack{autotune (init $0.2$),\\$\alpha_{\text{lr}}{=}3\!\times\!10^{-4}$,\; $\bar{\mathcal H}\!=\!0$} &
\shortstack{autotune (init $0.2$),\\$\alpha_{\text{lr}}{=}3\!\times\!10^{-4}$,\; $\bar{\mathcal H}\!=\!0$} \\
\cmidrule(lr){1-4}
Action range &
\shortstack{$\tanh$ squash\\(deterministic)} &
\shortstack{$\tanh$ + Jacobian\\(for log-prob)} &
\shortstack{$\tanh$ + Jacobian\\(for log-prob)} \\
\cmidrule(lr){1-4}
Gate init / dims &
\shortstack{—} &
\shortstack{gate head: $W{=}0$, $b{=}5.0$\\hidden $256$} &
\shortstack{—} \\
\cmidrule(lr){1-4}
Transformer dims &
\shortstack{—} &
\shortstack{—} &
\shortstack{$d{=}128$, $n_H{=}4$, $n_L{=}2$} \\
\cmidrule(lr){1-4}
Actor hidden dims &
\shortstack{$(512,512,512,512)$} &
\shortstack{$(512,512,512,512)$} &
\shortstack{(used only in enc./FFN;\\decoder per row above)} \\
\cmidrule(lr){1-4}
Action chunking &
\shortstack{\texttt{True} } &
\shortstack{\texttt{True} } &
\shortstack{\texttt{True} } \\
\cmidrule(lr){1-4}
Opt / LR / WD &
\shortstack{Adam, $3\!\times\!10^{-4}$} &
\shortstack{Adam, $3\!\times\!10^{-4}$} &
\shortstack{Adam, $3\!\times\!10^{-4}$} \\
\cmidrule(lr){1-4}
Batch / $\gamma$ / $\tau$ &
\shortstack{$256$ / $0.99$ / $0.005$} &
\shortstack{$256$ / $0.99$ / $0.005$} &
\shortstack{$256$ / $0.99$ / $0.005$} \\
\bottomrule
\end{tabular}
\end{table}

\section{More Experiment Results}
\label{more_exp}
In this section, we show more tested experiments.
\begin{figure}[htbp]
  \centering 
  \subfloat[\centering Sampling Step $k=0$]{%
    \includegraphics[width=0.21\textwidth]{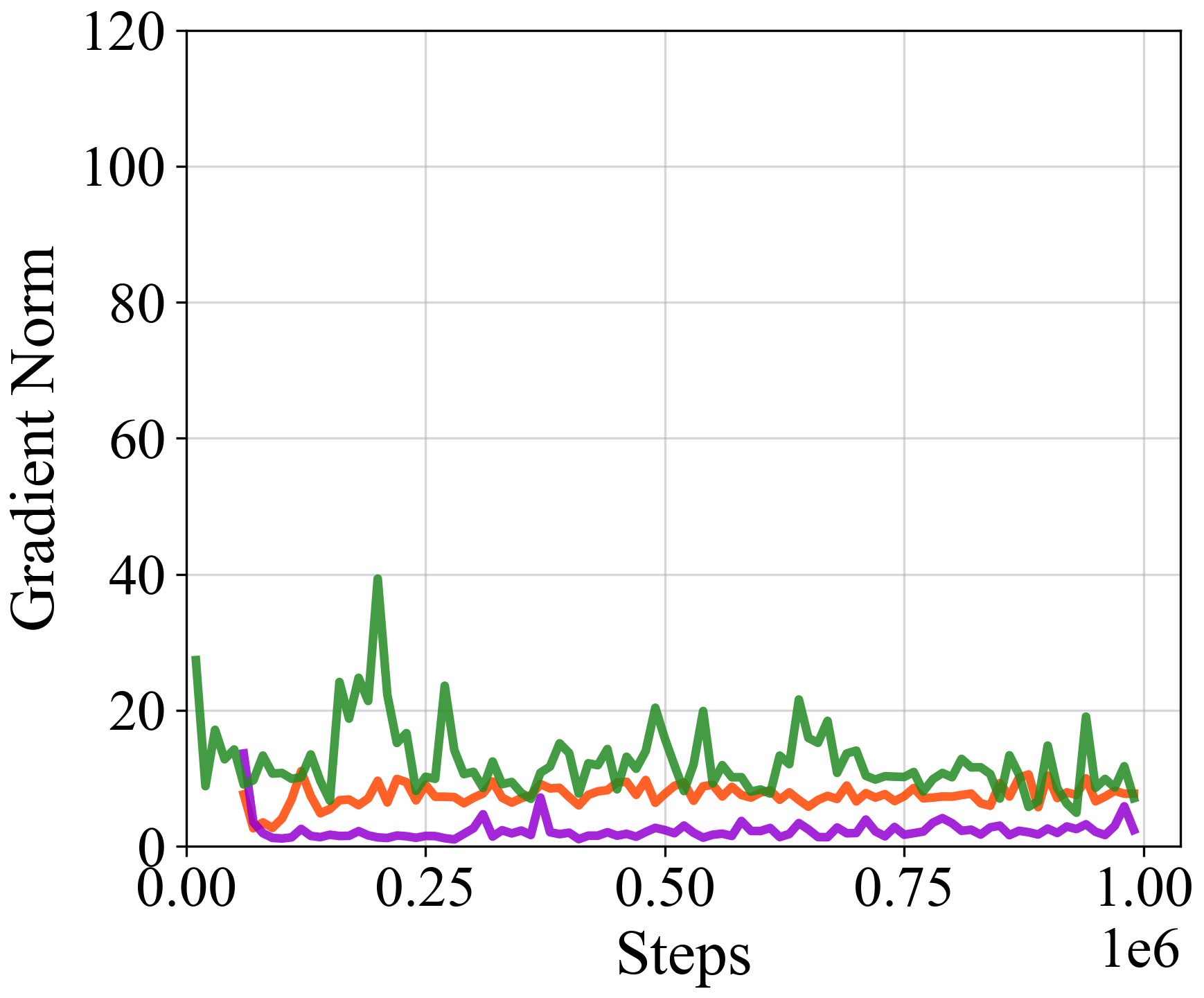}%
    \label{fig:bars} 
  }
  \subfloat[\centering Sampling Step $k=1$]{%
    \includegraphics[width=0.21\textwidth]{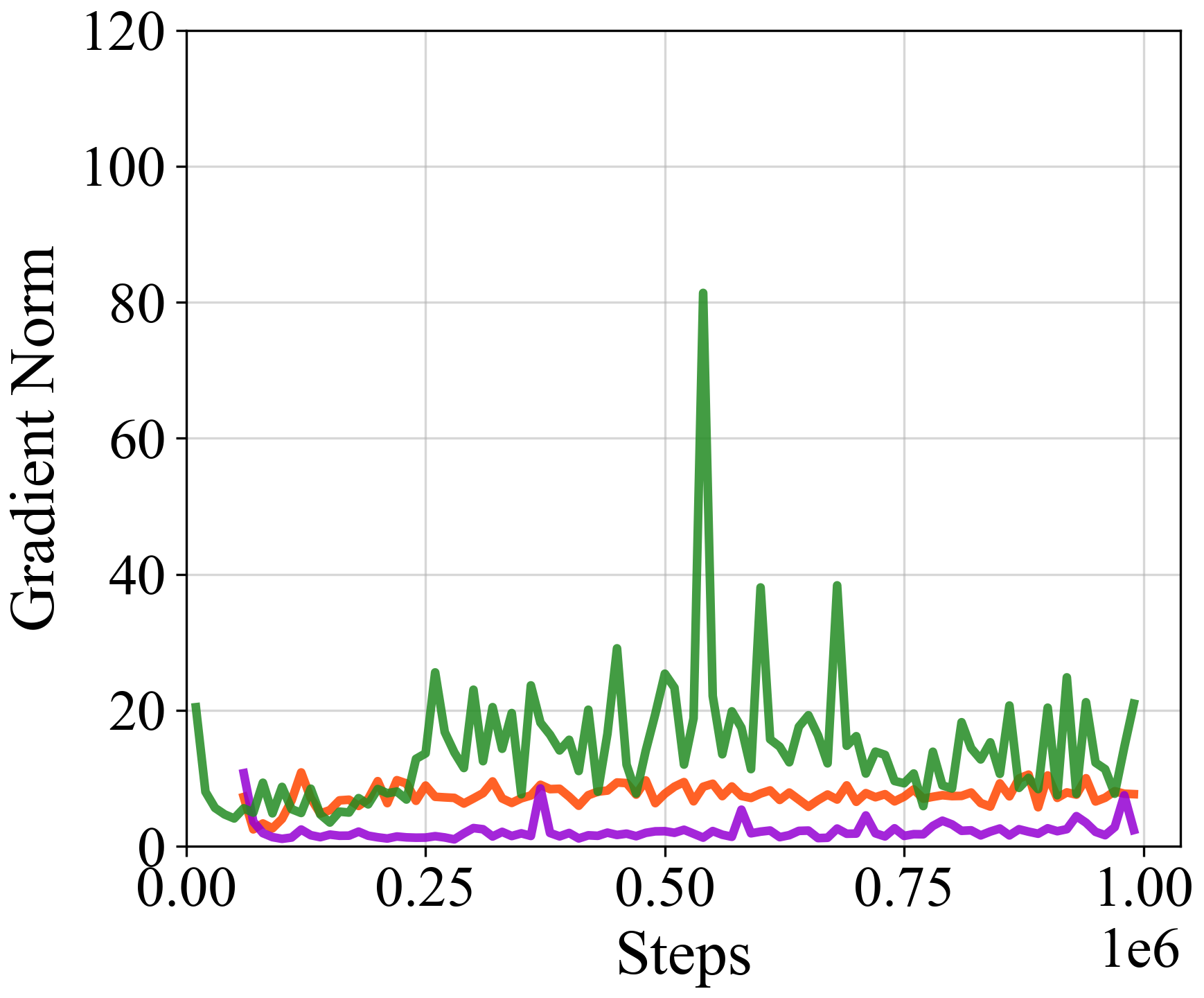}%
    \label{fig:gn_ant} 
  }
  \subfloat[\centering Sampling Step $k=2$]{%
    \includegraphics[width=0.21\textwidth]{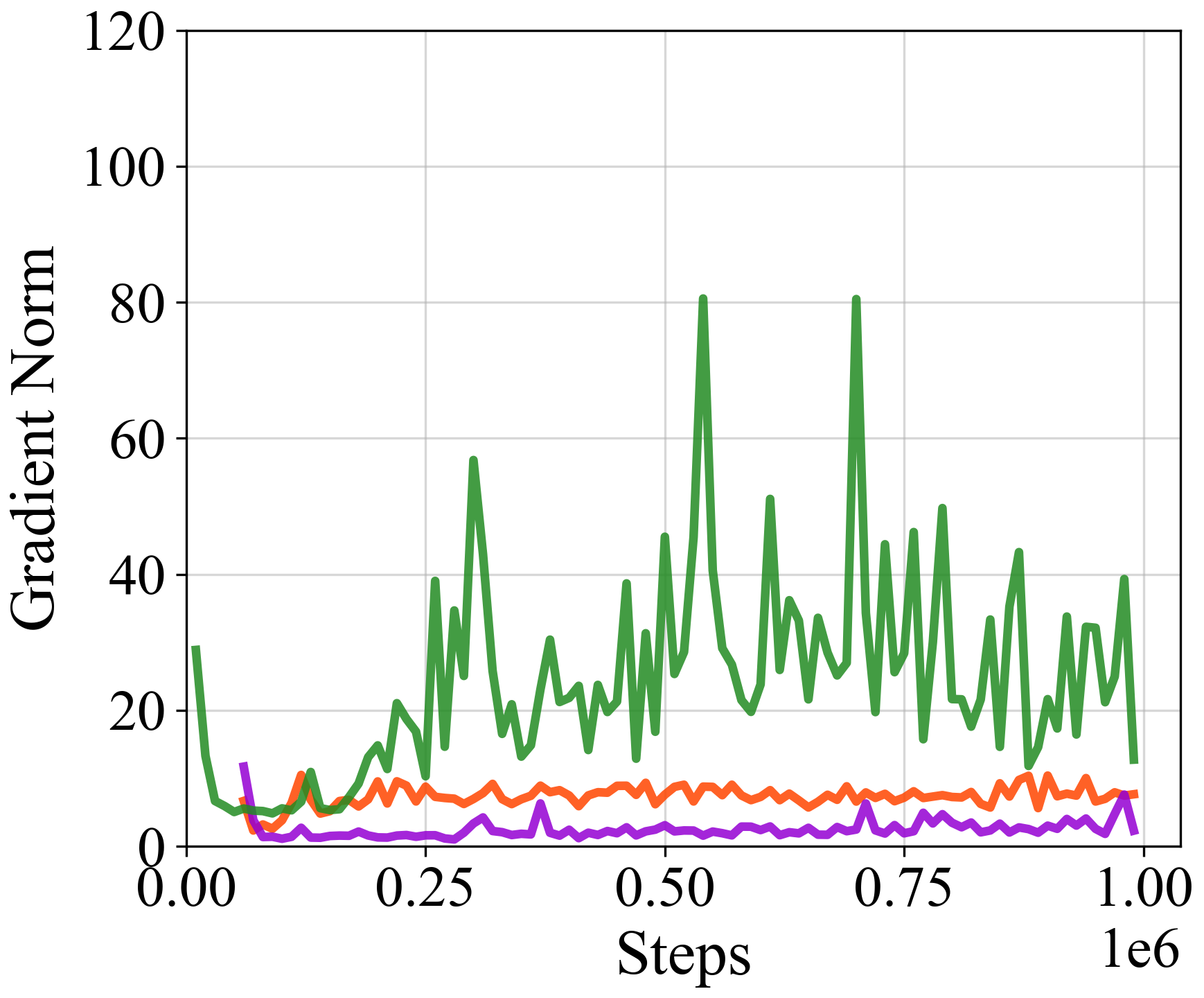}%
    \label{fig:gn_cube} 
  }
  \subfloat[\centering Sampling Step $k=3$]{%
    \includegraphics[width=0.21\textwidth]{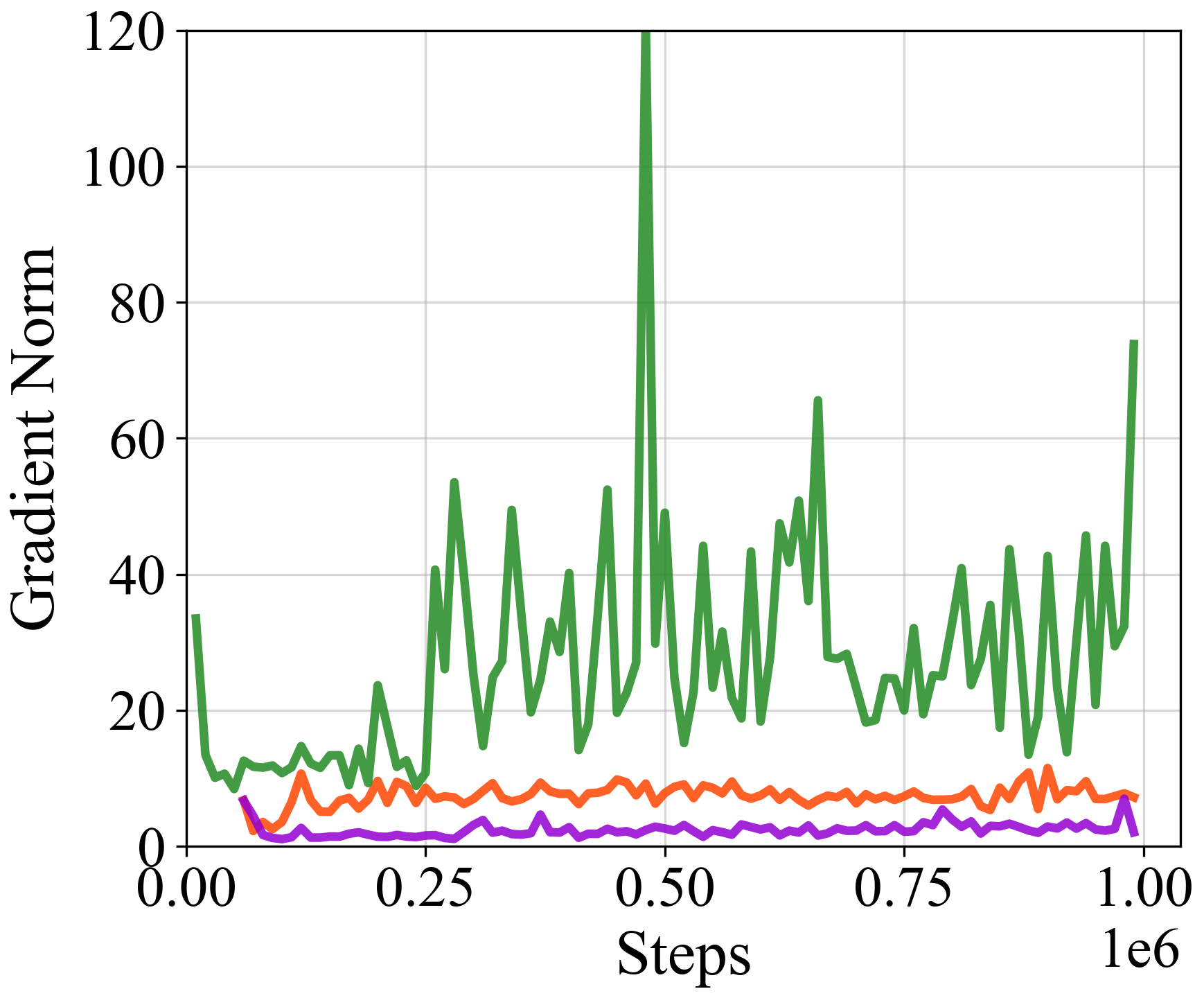}%
    \label{fig:gn_cube} 
  }
  \\
  \vspace{-1mm}
  \subfloat{
  \includegraphics[width = 0.7\textwidth]{figs/ablations/abl-cmp-legend.png}
  }
  \vspace{-4mm}
  \caption{Ablation study on velocity network parameterizations across different sampling steps in Walker2d. The gradient norm explodes as the sampling step $k$ increases.
  }
  \label{fig:gradient_norm} 
  \vspace{-2mm}
\end{figure}

\begin{figure}[htbp]
  \centering 
  \subfloat[\centering Sampling Step $k=0$]{%
    \includegraphics[width=0.21\textwidth]{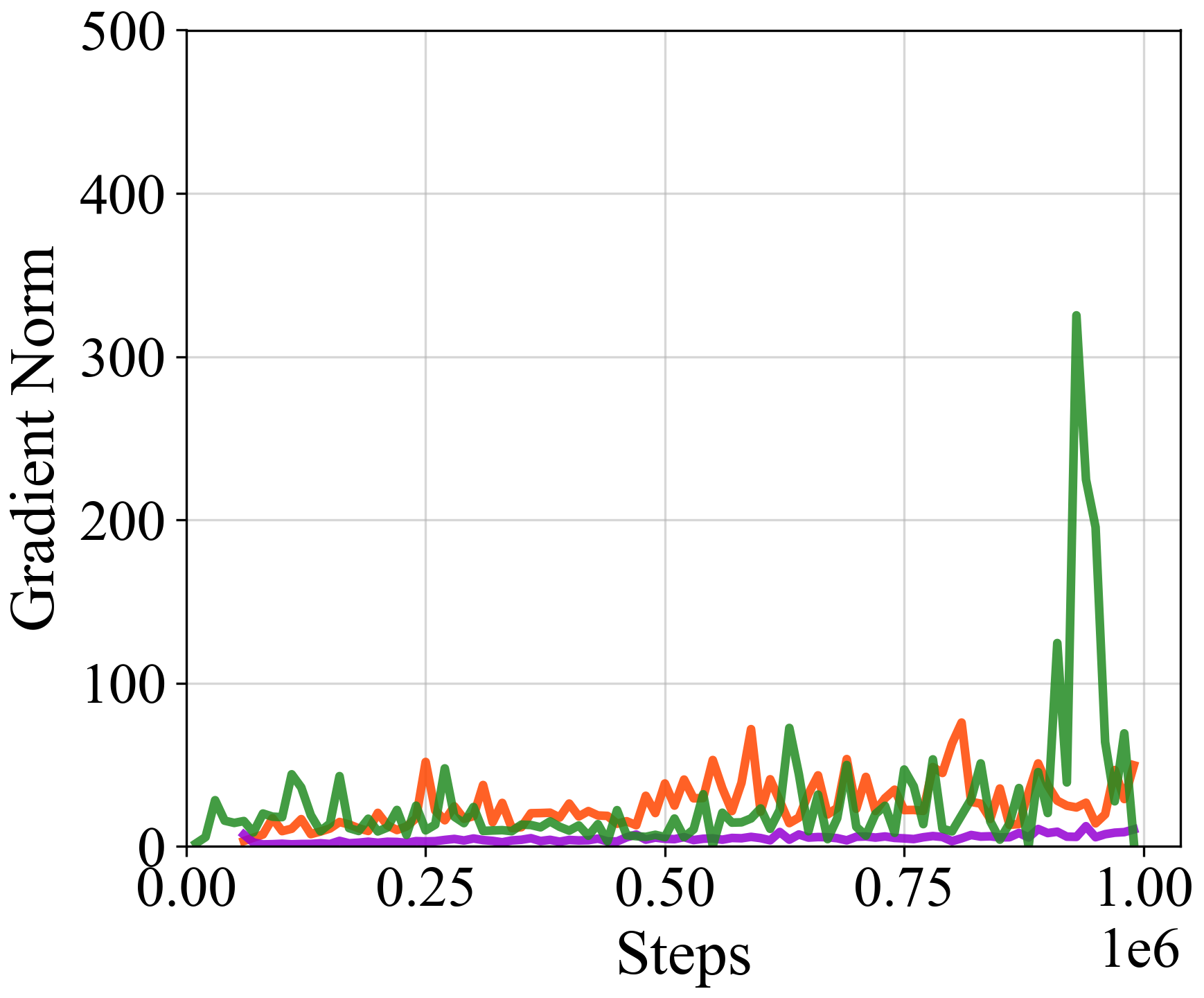}%
    \label{fig:bars} 
  }
  \subfloat[\centering Sampling Step $k=1$]{%
    \includegraphics[width=0.21\textwidth]{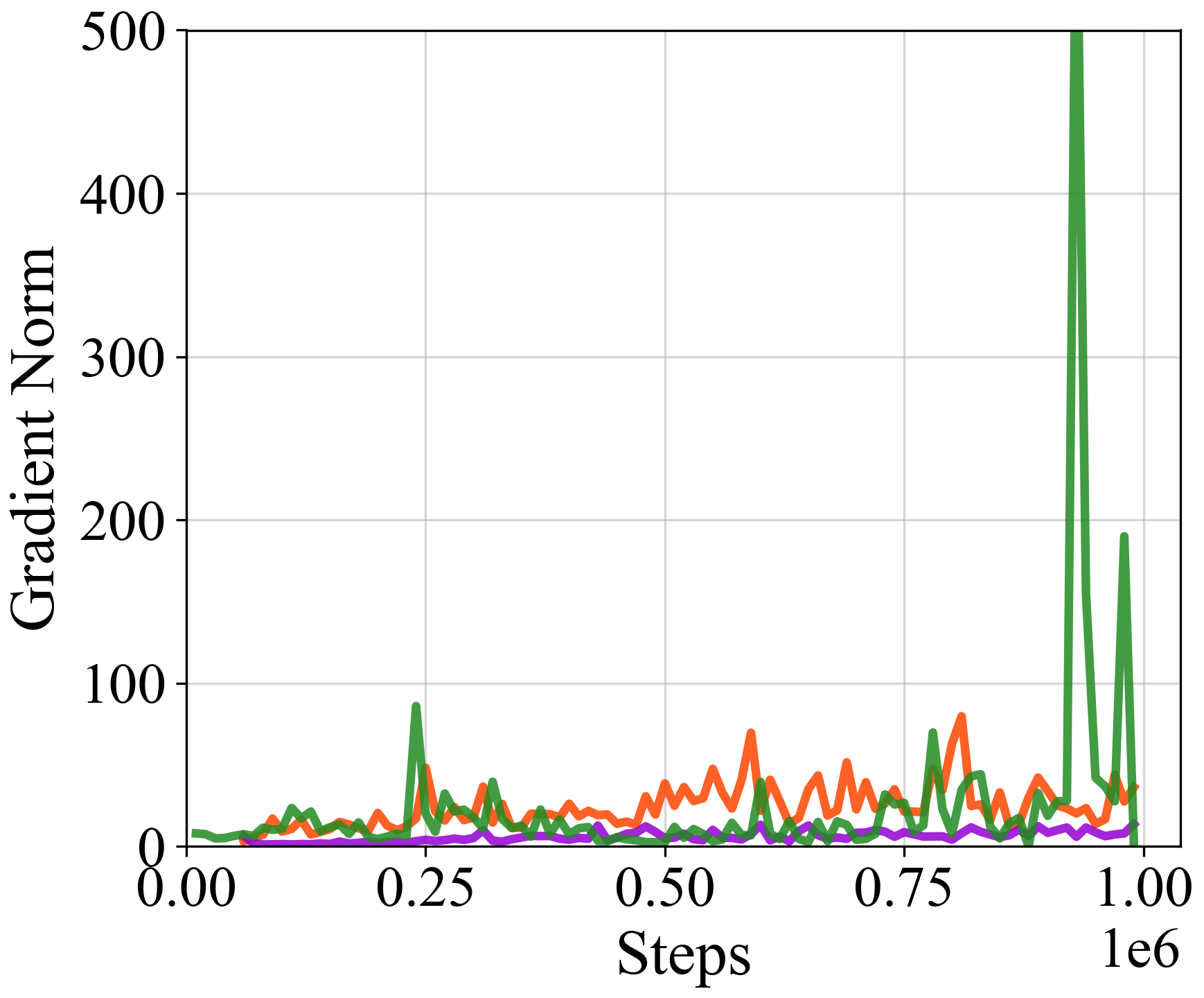}%
    \label{fig:gn_ant} 
  }
  \subfloat[\centering Sampling Step $k=2$]{%
    \includegraphics[width=0.21\textwidth]{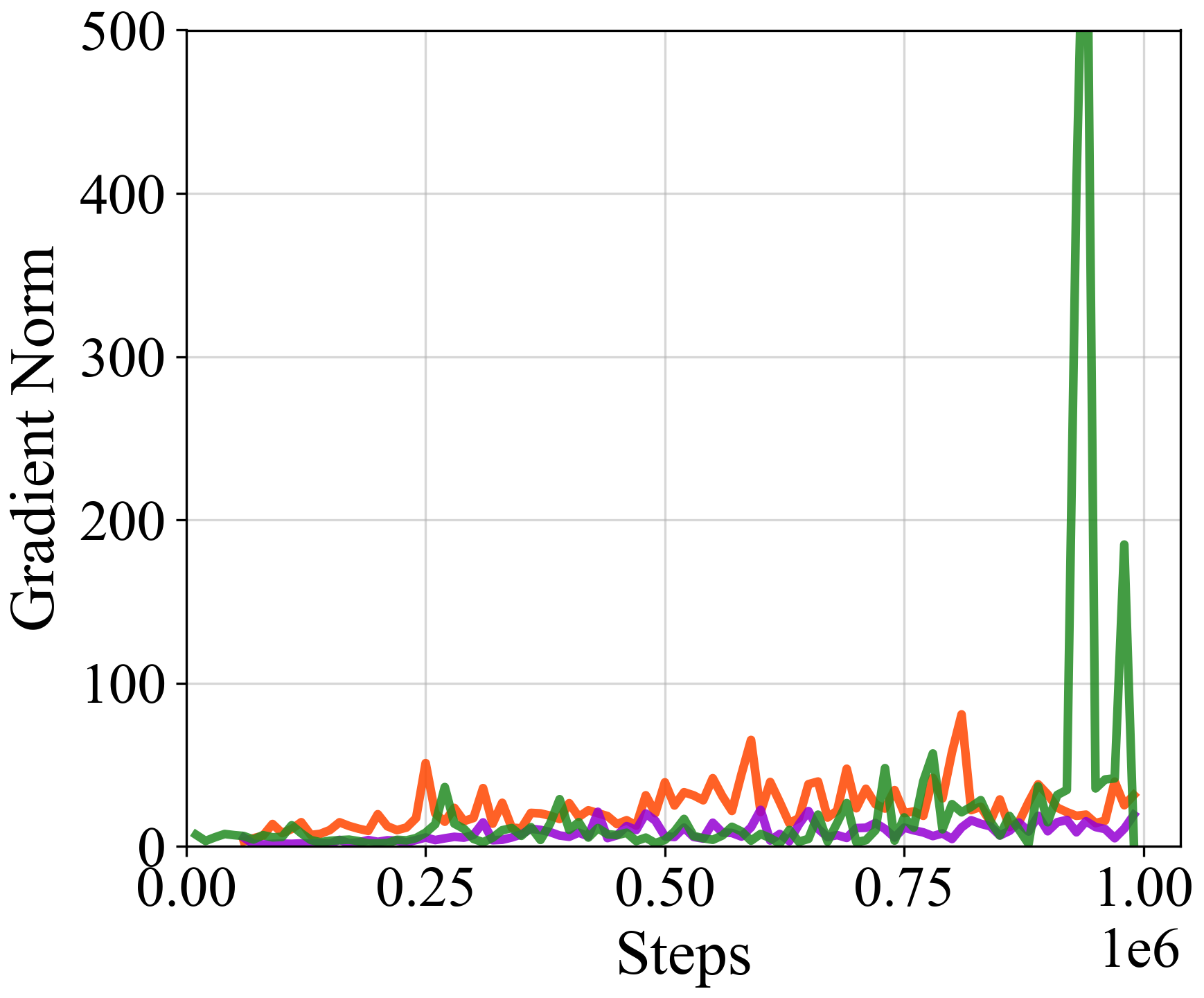}%
    \label{fig:gn_cube} 
  }
  \subfloat[\centering Sampling Step $k=3$]{%
    \includegraphics[width=0.21\textwidth]{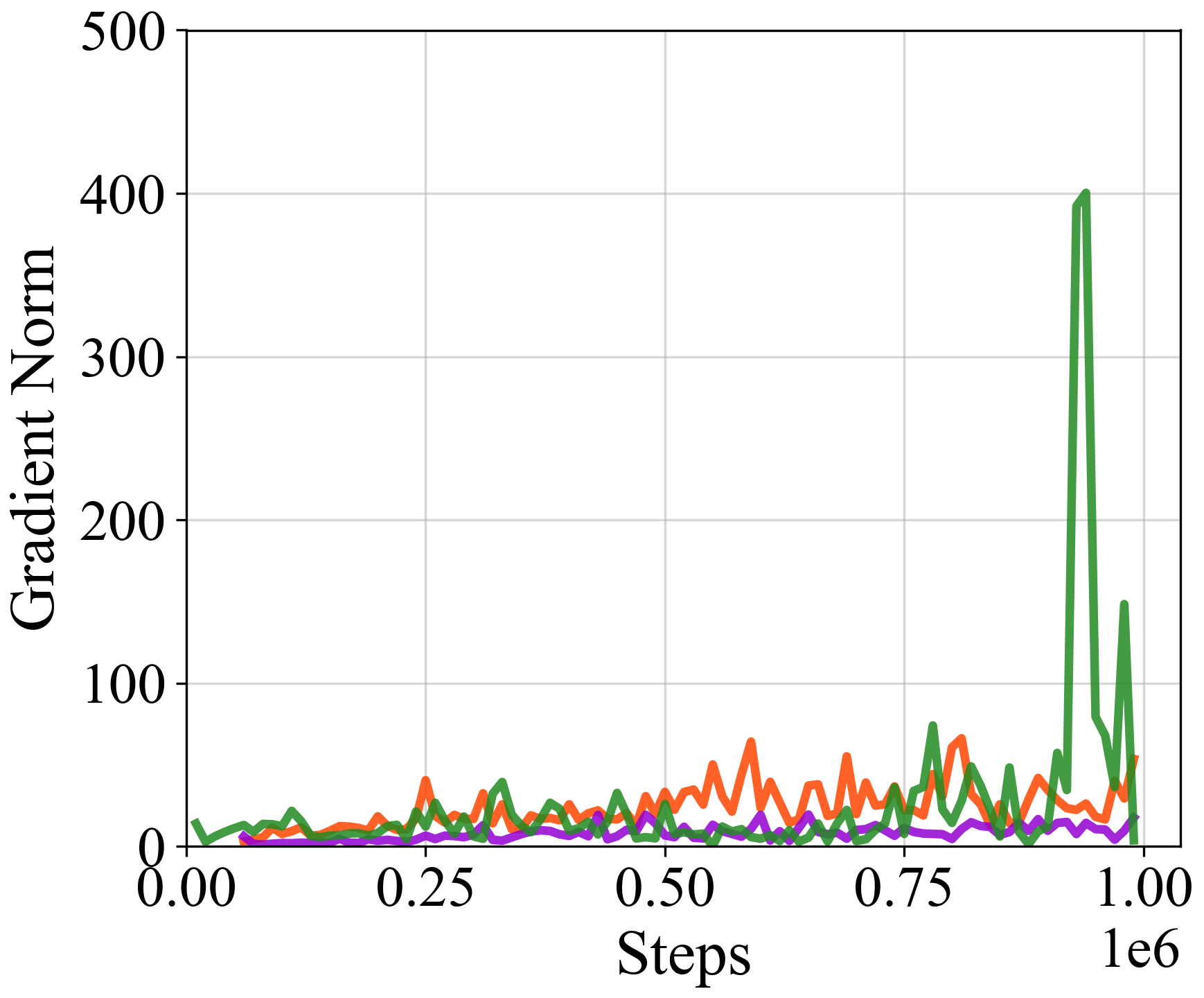}%
    \label{fig:gn_cube} 
  }
  \\
  \vspace{-1mm}
  \subfloat{
  \includegraphics[width = 0.7\textwidth]{figs/ablations/abl-cmp-legend.png}
  }
  \vspace{-4mm}
  \caption{Ablation study on velocity network parameterizations across different sampling steps in HalfCheetah.
  }
  \label{fig:gradient_norm} 
  \vspace{-2mm}
\end{figure}

\begin{figure}[t]
    \centering 

    \subfloat[\captionsetup{justification=centering}Cube-Quadruple-Task1]{%
        \includegraphics[width=0.19\textwidth]{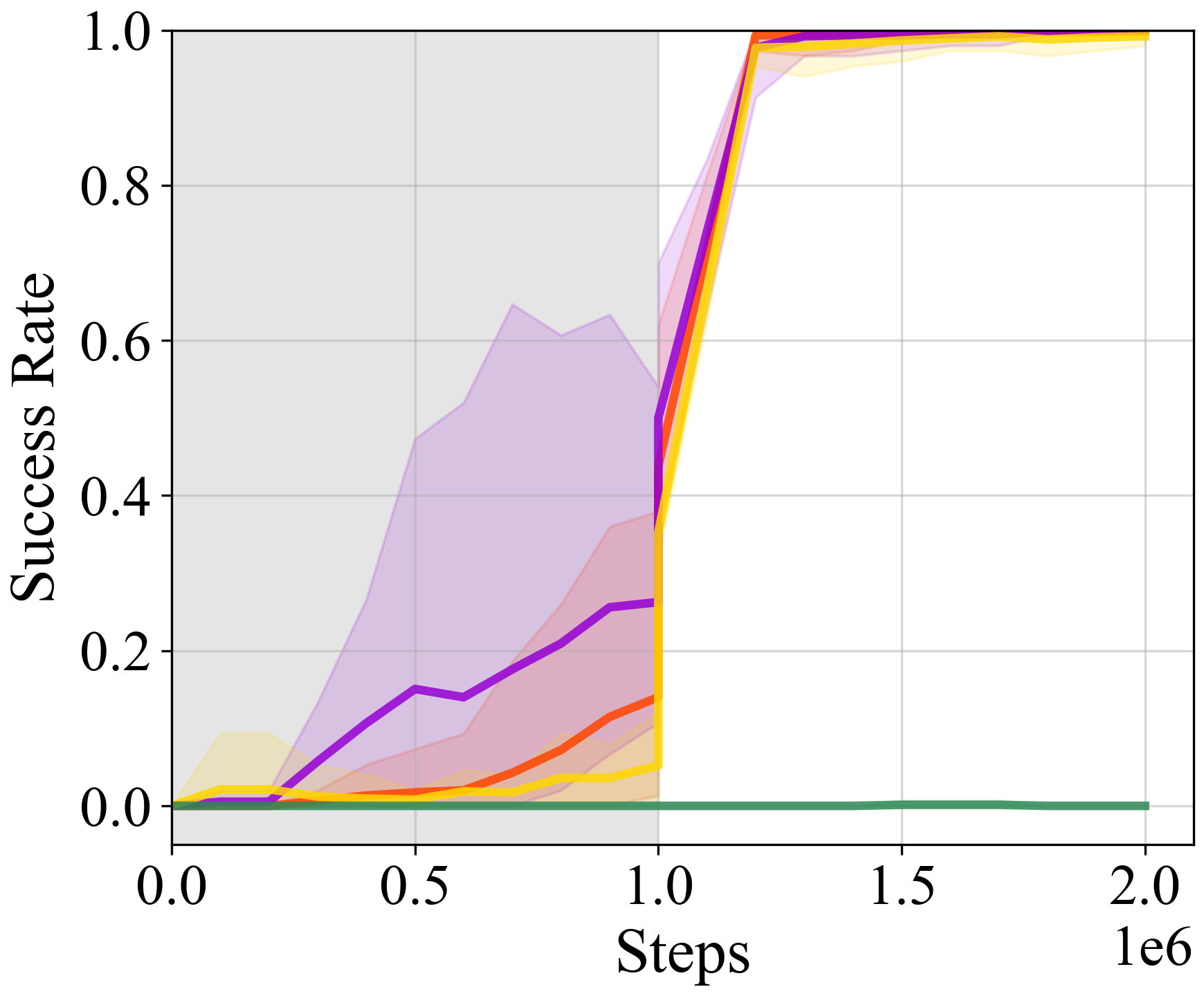}%
        \label{fig:quadruple-task1}%
    }
    \hfill
    \subfloat[\captionsetup{justification=centering}Cube-Quadruple-Task2]{%
        \includegraphics[width=0.19\textwidth]{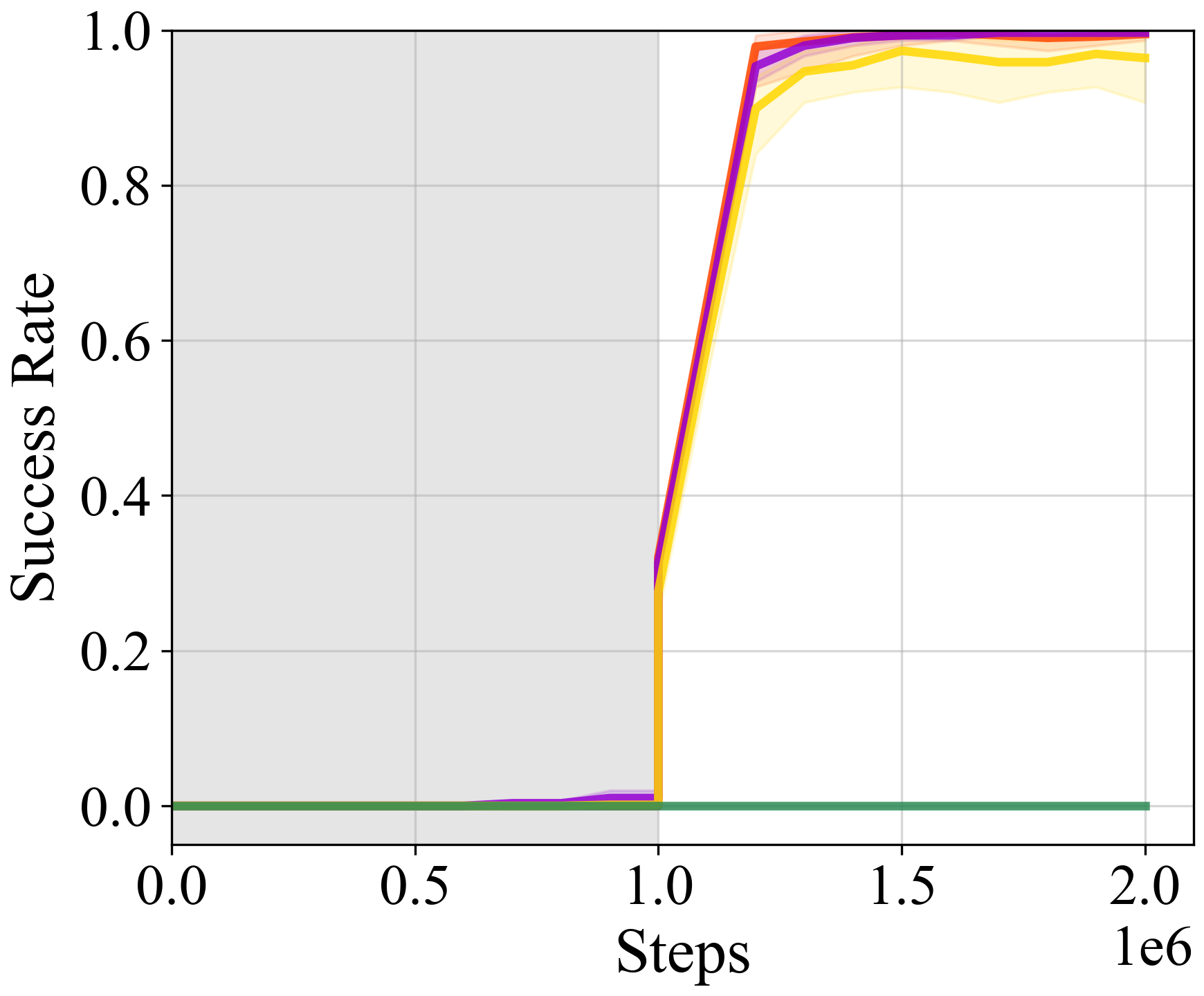}%
        \label{fig:quadruple-task2}%
    }
    \hfill
    \subfloat[\captionsetup{justification=centering}Cube-Quadruple-Task3]{%
        \includegraphics[width=0.19\textwidth]{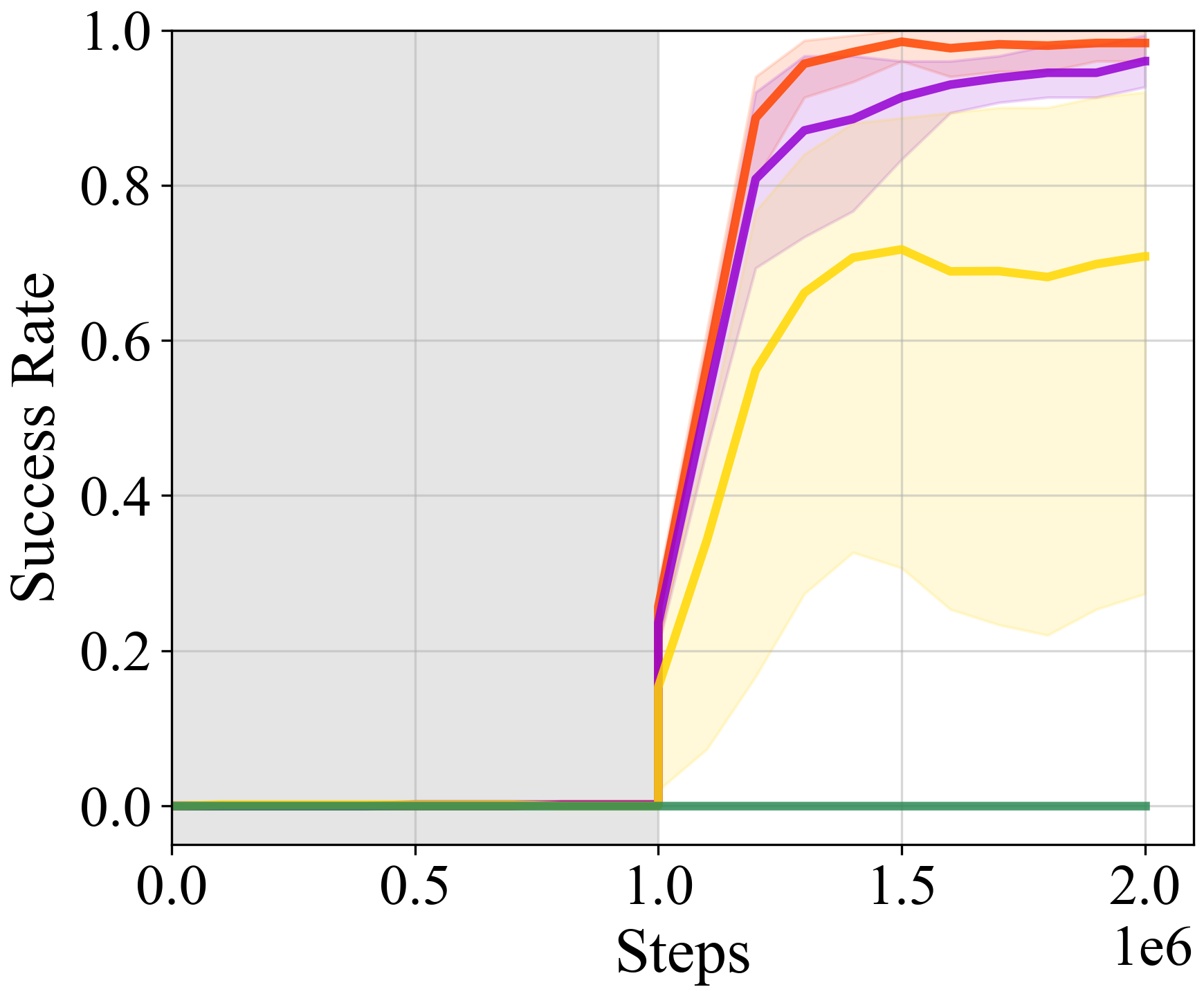}%
        \label{fig:quadruple-task3}%
    }
    \hfill
    \subfloat[\captionsetup{justification=centering}Cube-Quadruple-Task4]{%
        \includegraphics[width=0.19\textwidth]{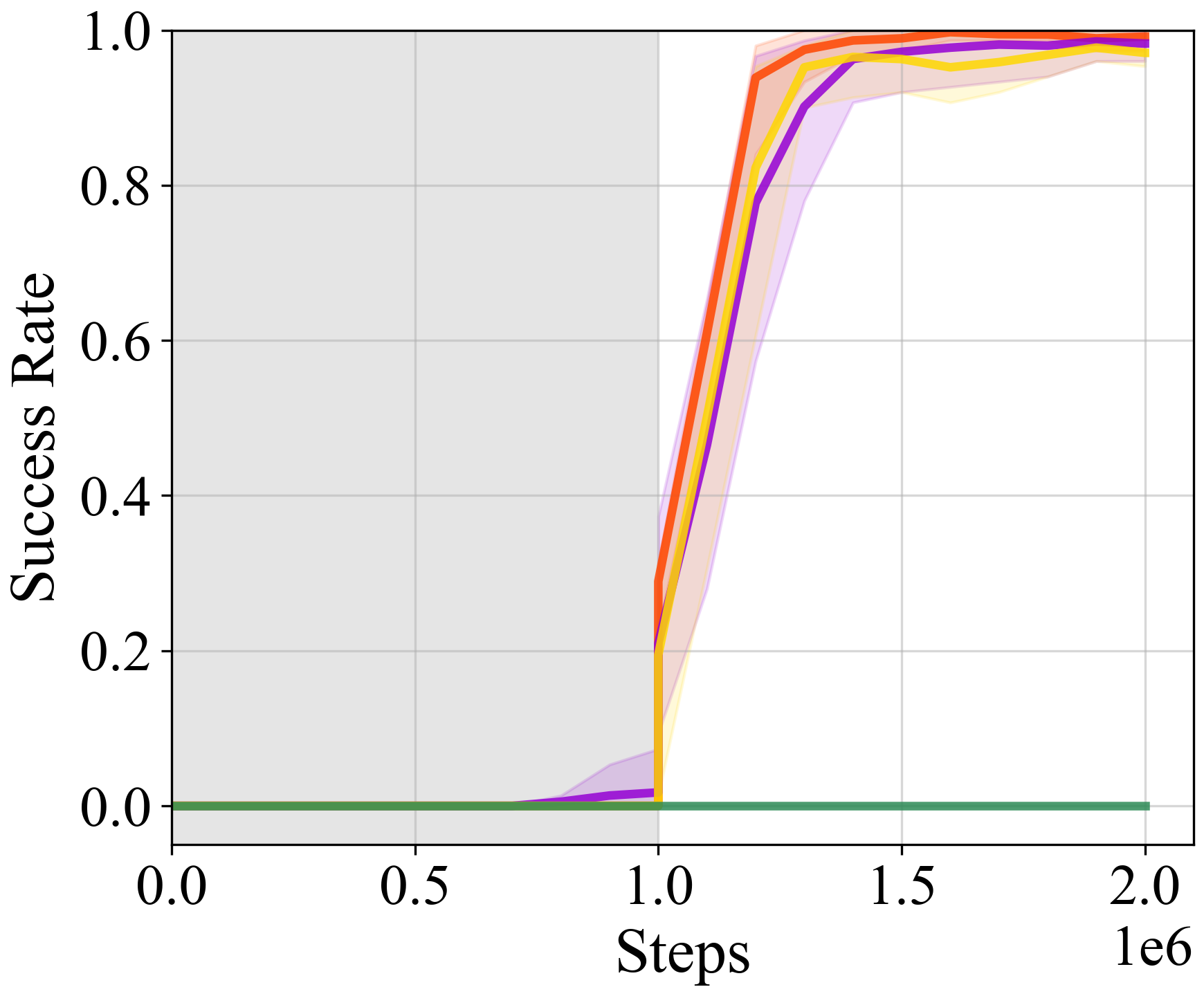}%
        \label{fig:quadruple-task4}%
    }
    \hfill
    \subfloat[\captionsetup{justification=centering}Cube-Quadruple-Task5]{%
        \includegraphics[width=0.19\textwidth]{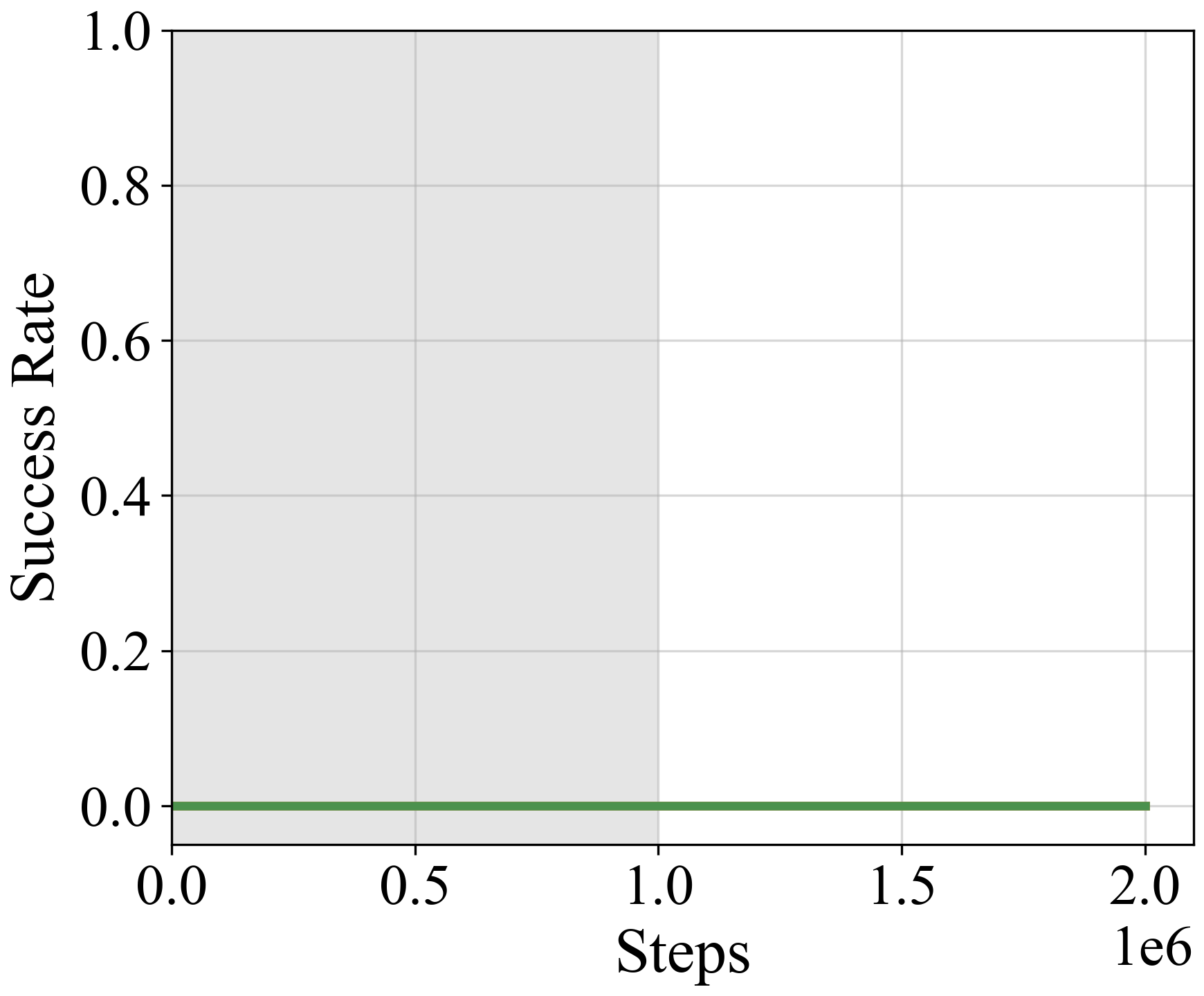}%
        \label{fig:quadruple-task5}%
    }
    \\
    \subfloat[\captionsetup{justification=centering}Cube-Triple-Task1]{%
        \includegraphics[width=0.19\textwidth]{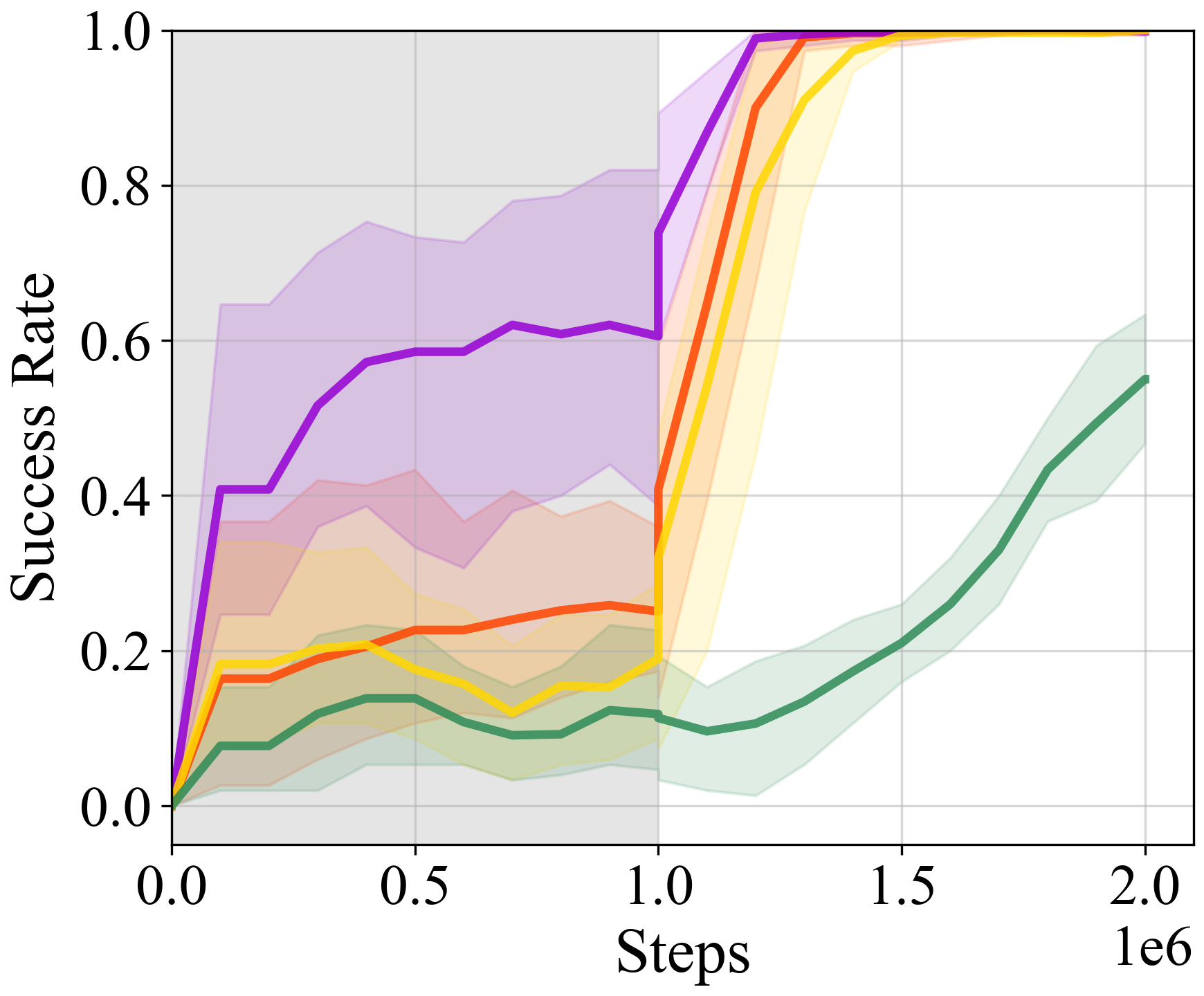}%
        \label{fig:triple-task1}%
    }
    \hfill
    \subfloat[\captionsetup{justification=centering}Cube-Triple-Task2]{%
        \includegraphics[width=0.19\textwidth]{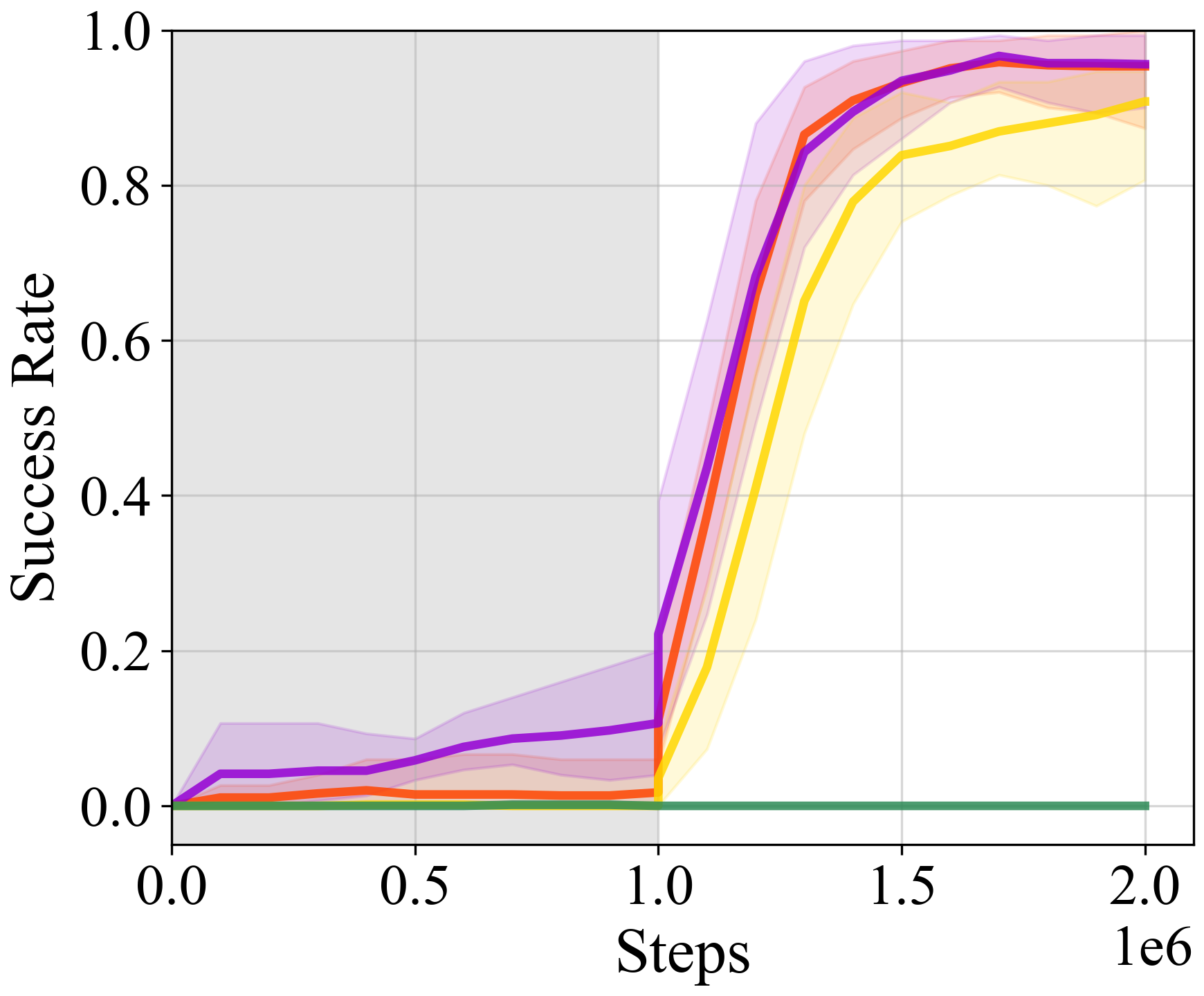}%
        \label{fig:triple-task2}%
    }
    \hfill
    \subfloat[\captionsetup{justification=centering}Cube-Triple-Task3]{%
        \includegraphics[width=0.19\textwidth]{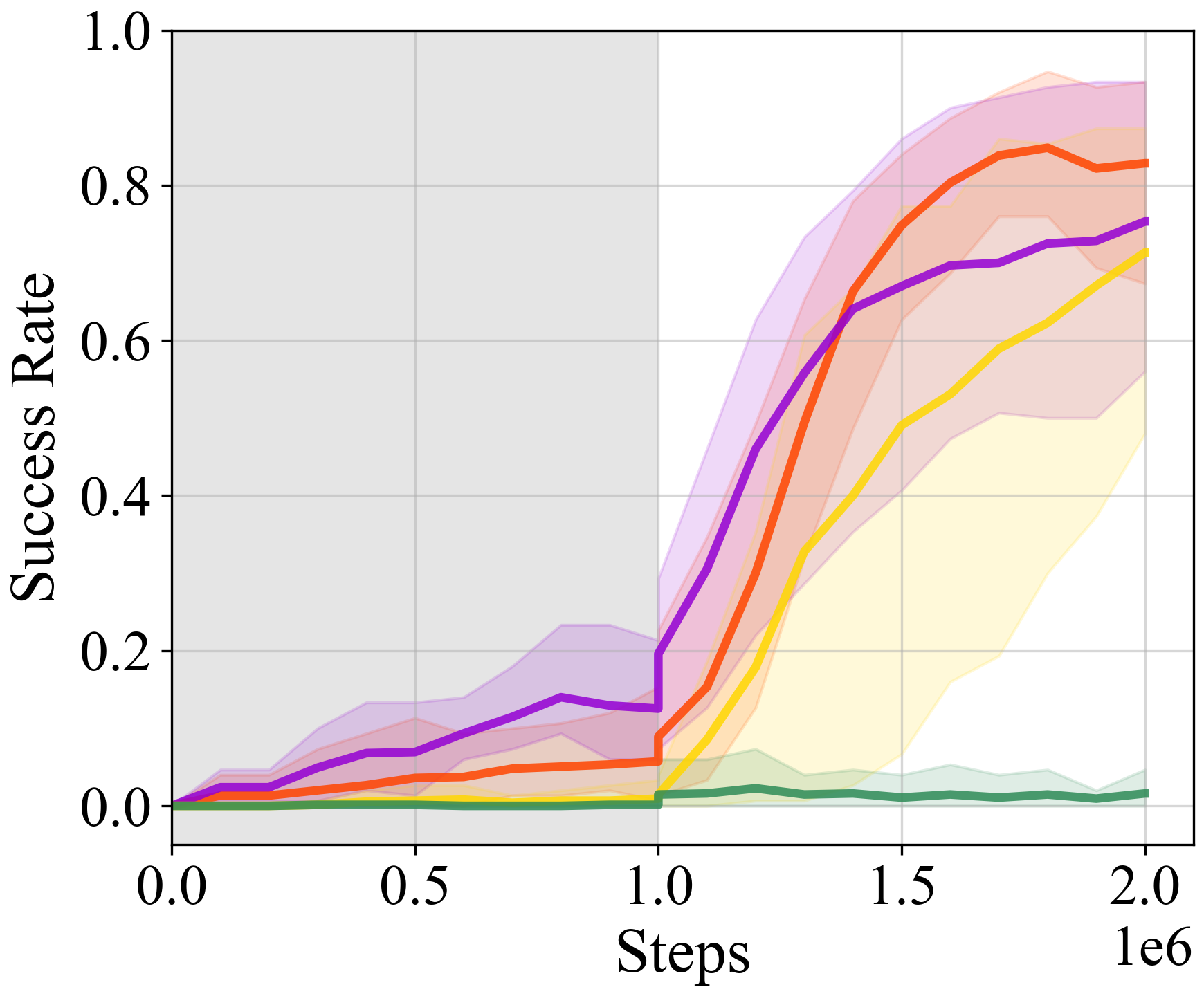}%
        \label{fig:triple-task3}%
    }
    \hfill
    \subfloat[\captionsetup{justification=centering}Cube-Triple-Task4]{%
        \includegraphics[width=0.19\textwidth]{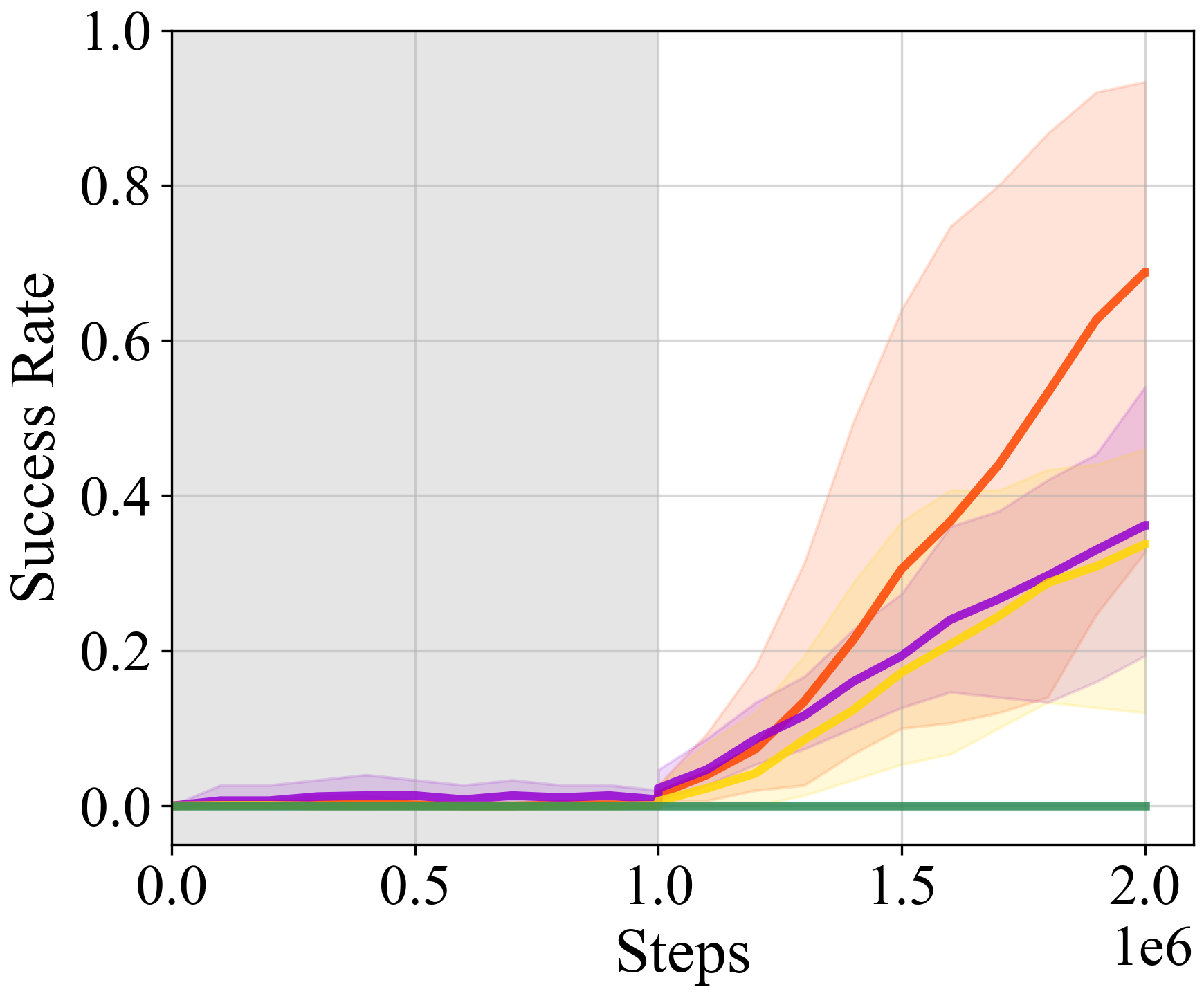}%
        \label{fig:triple-task4}%
    }
    \hfill
    \subfloat[\captionsetup{justification=centering}Cube-Triple-Task5]{%
        \includegraphics[width=0.19\textwidth]{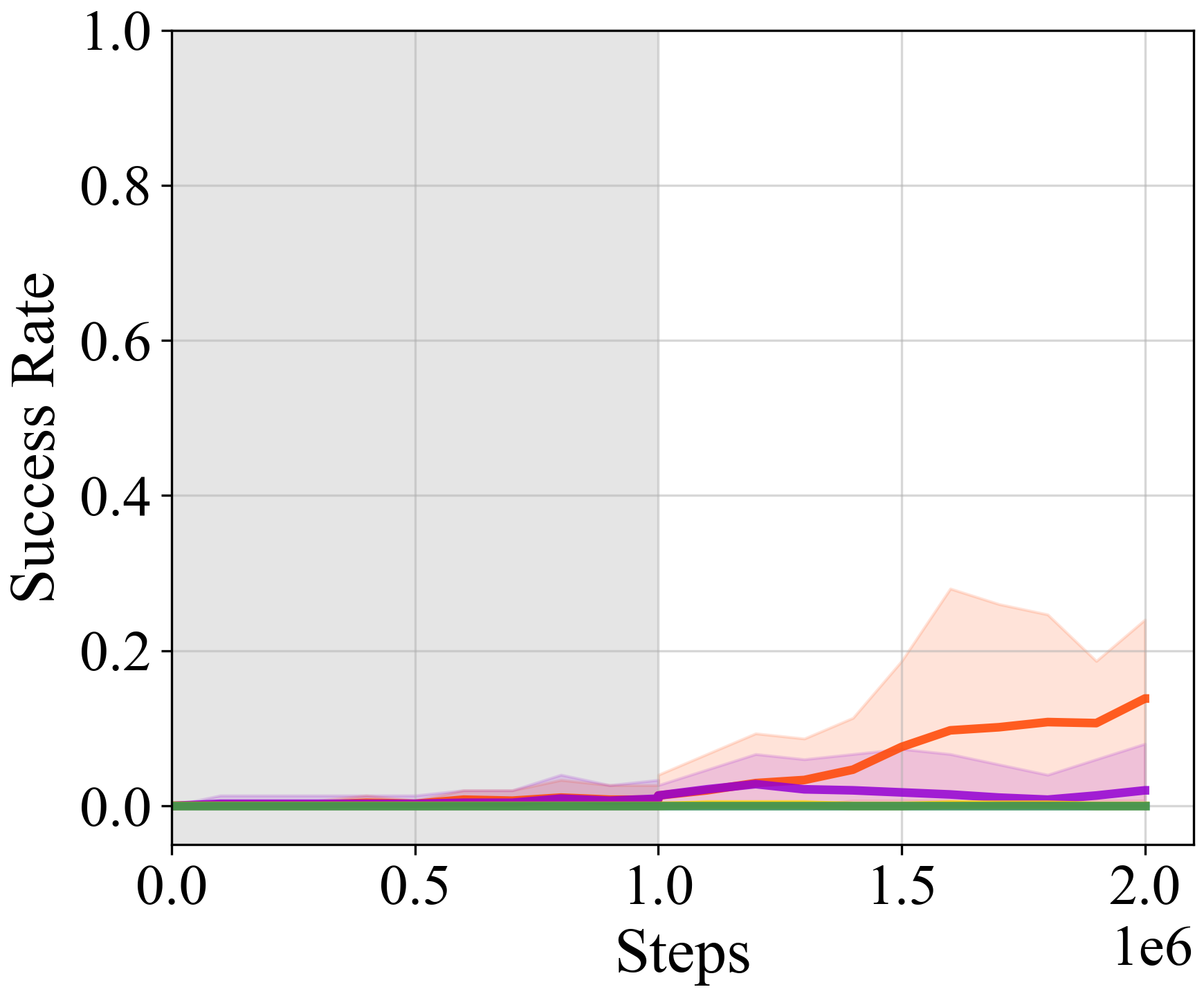}%
        \label{fig:triple-task5}%
    }
    \\
    \subfloat[\captionsetup{justification=centering}Cube-Double-Task1]{%
        \includegraphics[width=0.19\textwidth]{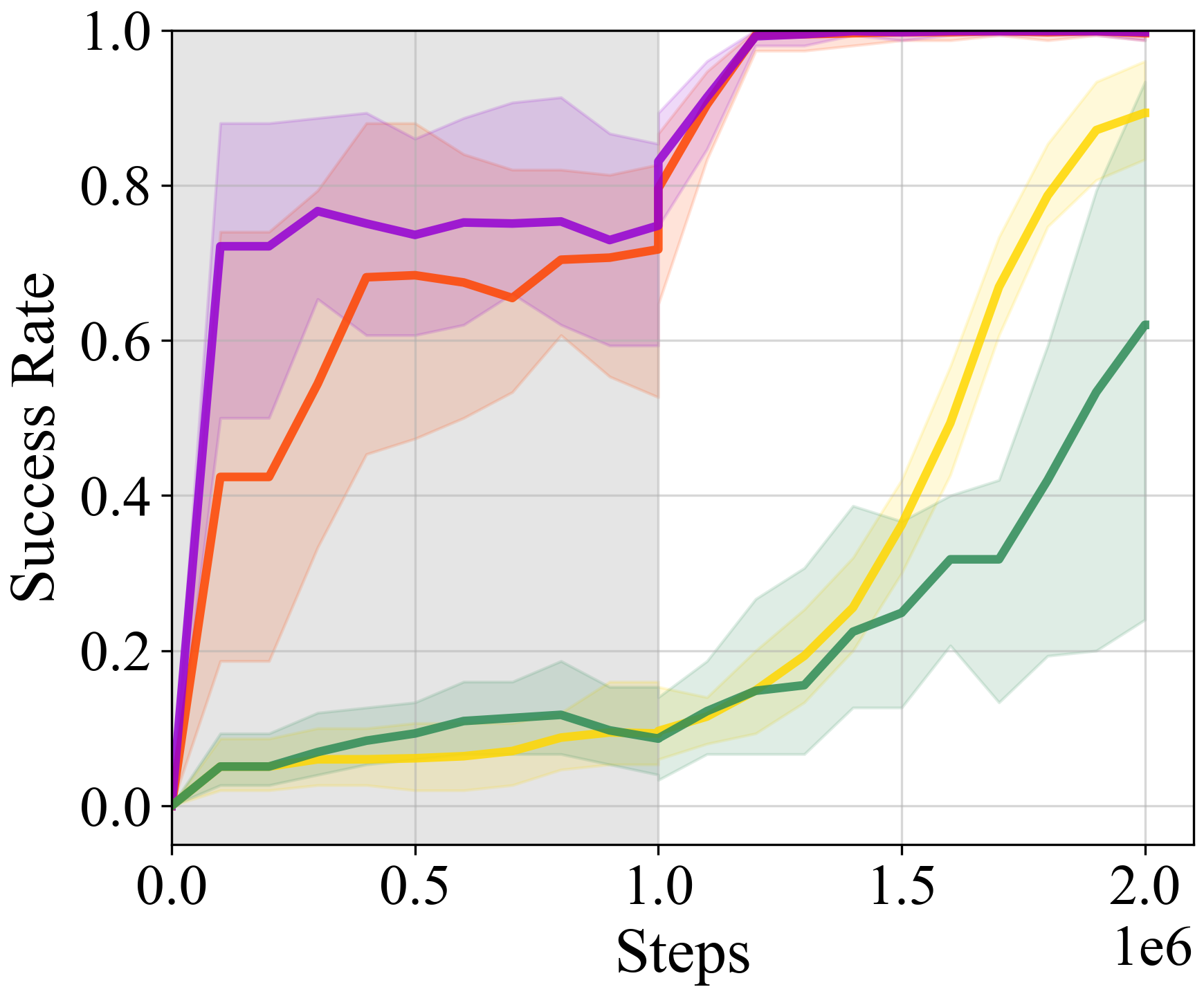}%
        \label{fig:double-task1}%
    }
    \hfill
    \subfloat[\captionsetup{justification=centering}Cube-Double-Task2]{%
        \includegraphics[width=0.19\textwidth]{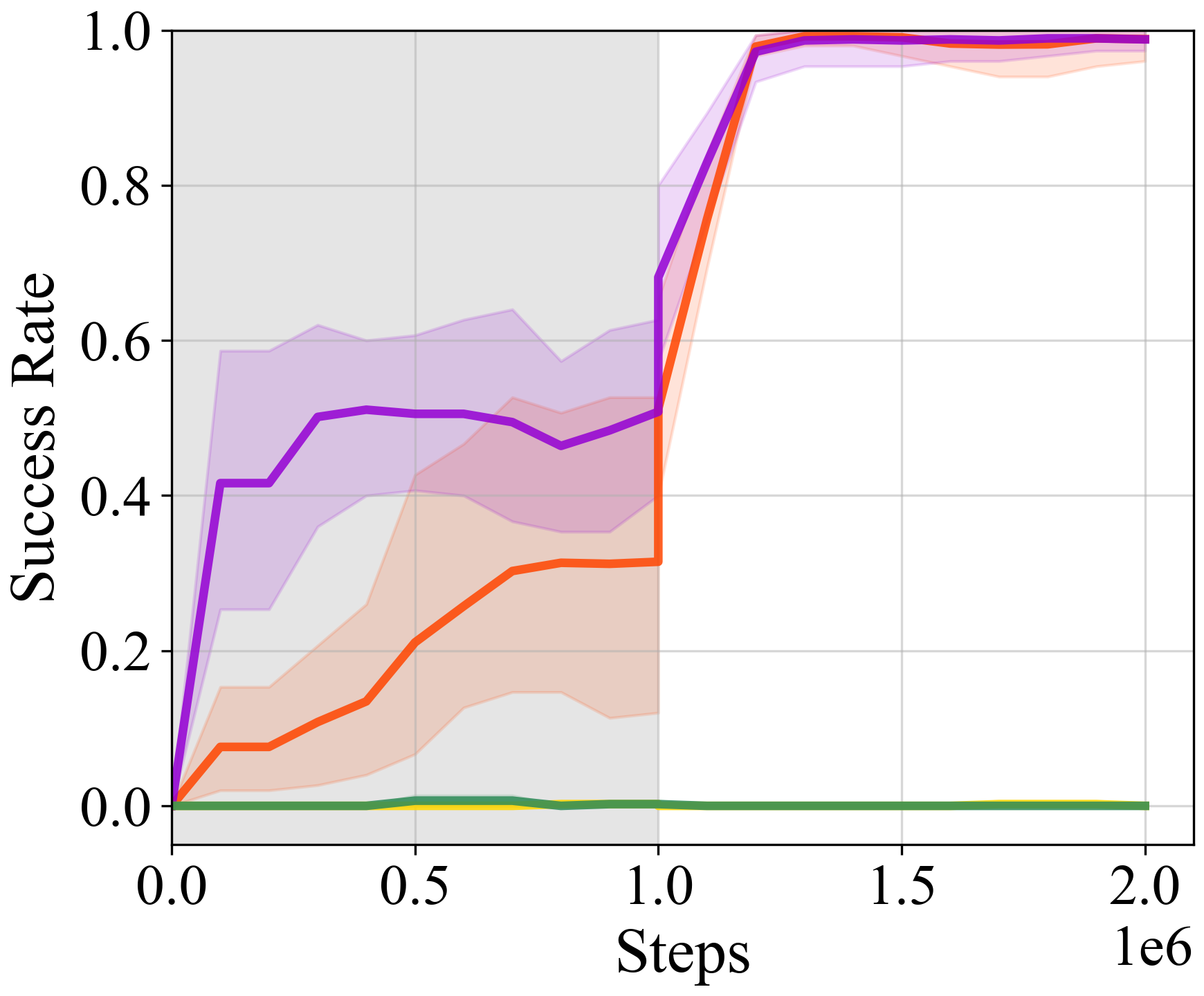}%
        \label{fig:double-task2}%
    }
    \hfill
    \subfloat[\captionsetup{justification=centering}Cube-Double-Task3]{%
        \includegraphics[width=0.19\textwidth]{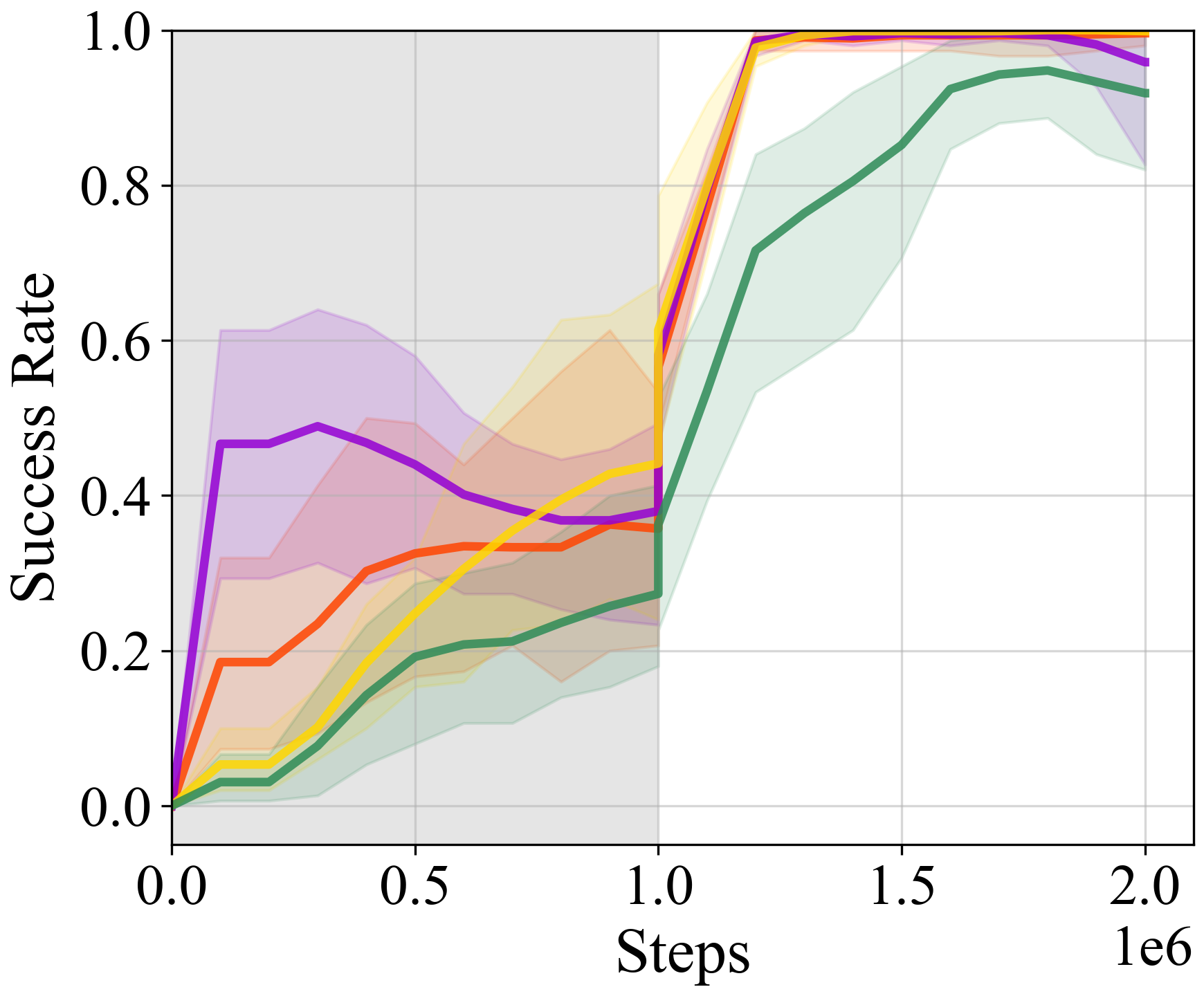}%
        \label{fig:double-task3}%
    }
    \hfill
    \subfloat[\captionsetup{justification=centering}Cube-Double-Task4]{%
        \includegraphics[width=0.19\textwidth]{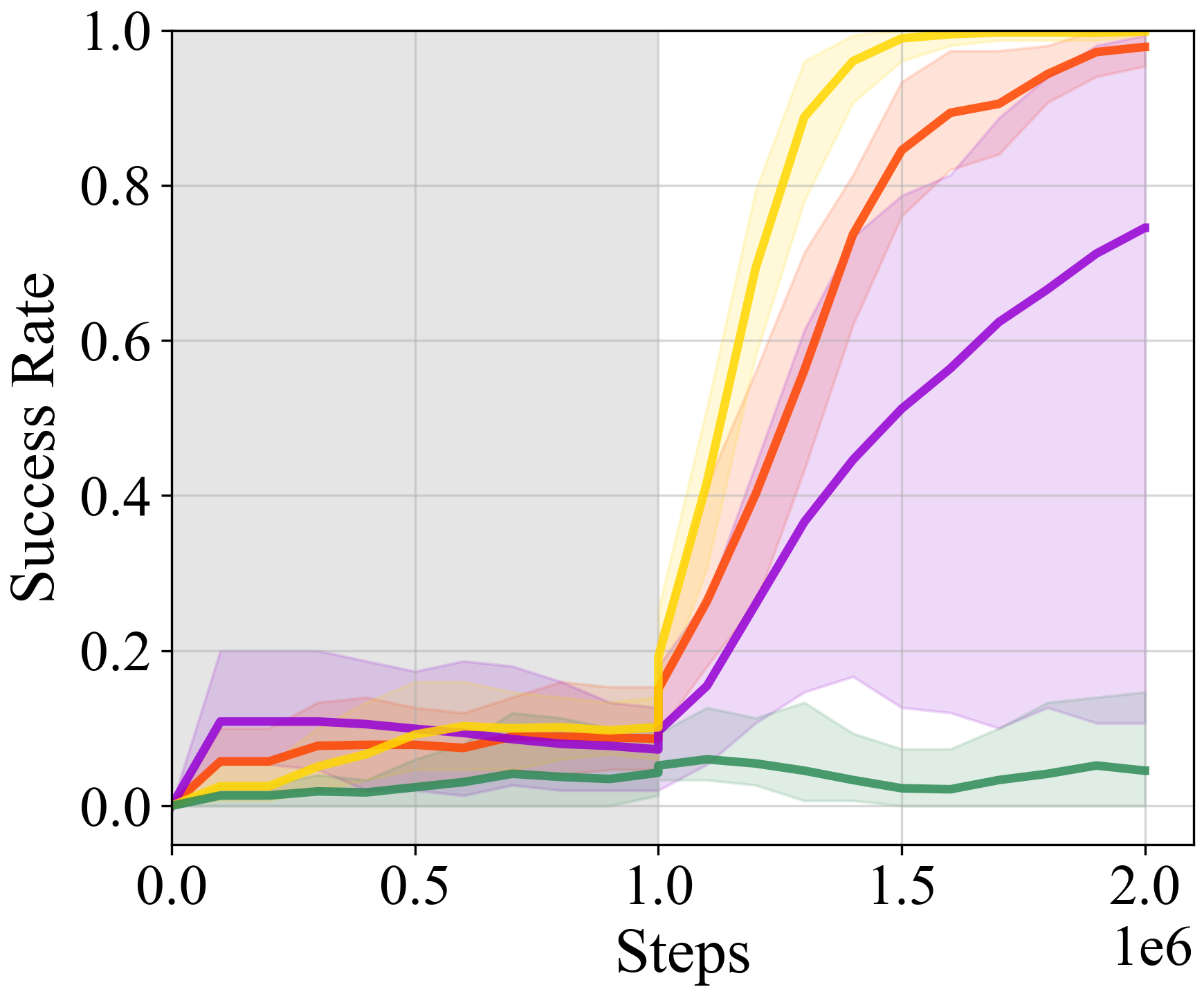}%
        \label{fig:double-task4}%
    }
    \hfill
    \subfloat[\captionsetup{justification=centering}Cube-Double-Task5]{%
        \includegraphics[width=0.19\textwidth]{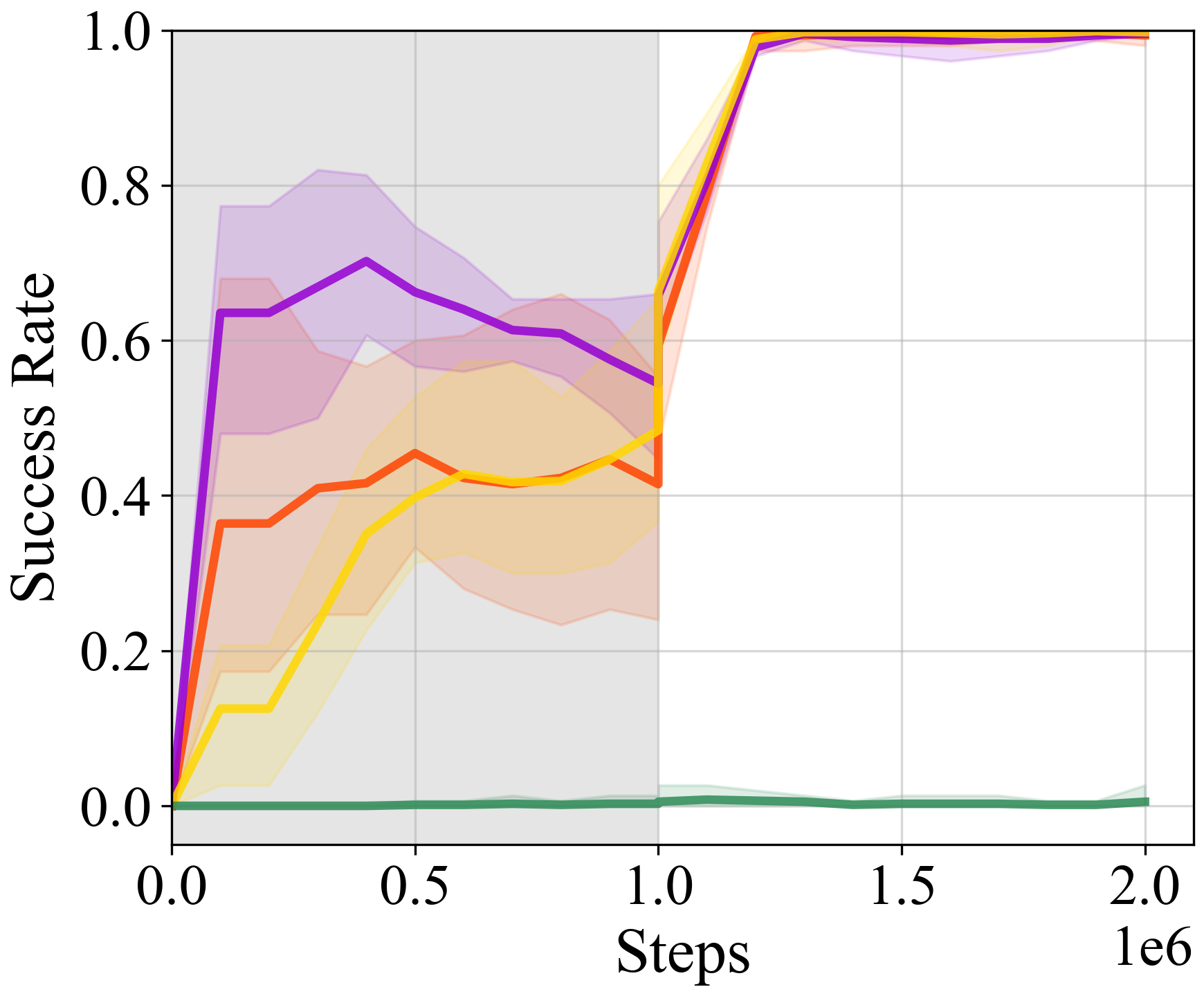}%
        \label{fig:double-task5}%
    }
    \\
    \vspace{-1mm}
    \subfloat{%
        \includegraphics[width=0.95\textwidth]{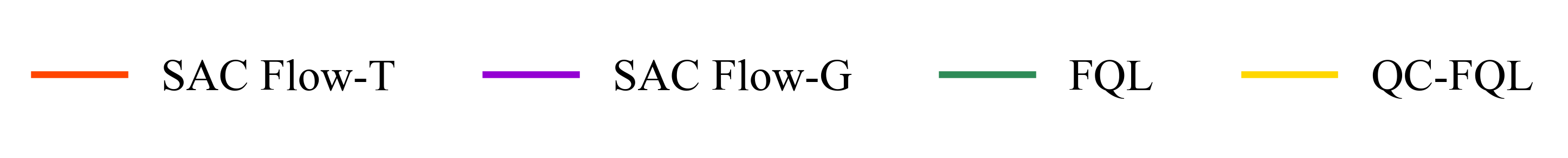}%
    }
    \vspace{-3mm}
    \caption{Complete offline-to-online training performance in OGBench. This figure illustrates the comprehensive training performance across all tasks. All methods are trained on 1M offline updates followed by 1M online interaction steps. Our methods, SAC Flow-T and SAC Flow-G, achieve competitive—often superior—performance across the evaluated benchmarks.}
    \label{fig:Complete_OGBench} 
    \vspace{-2mm}
\end{figure}

\begin{figure}[htbp]
  \centering 

  \subfloat[\centering Robomimic-Lift]{%
    \includegraphics[width=0.33\textwidth]{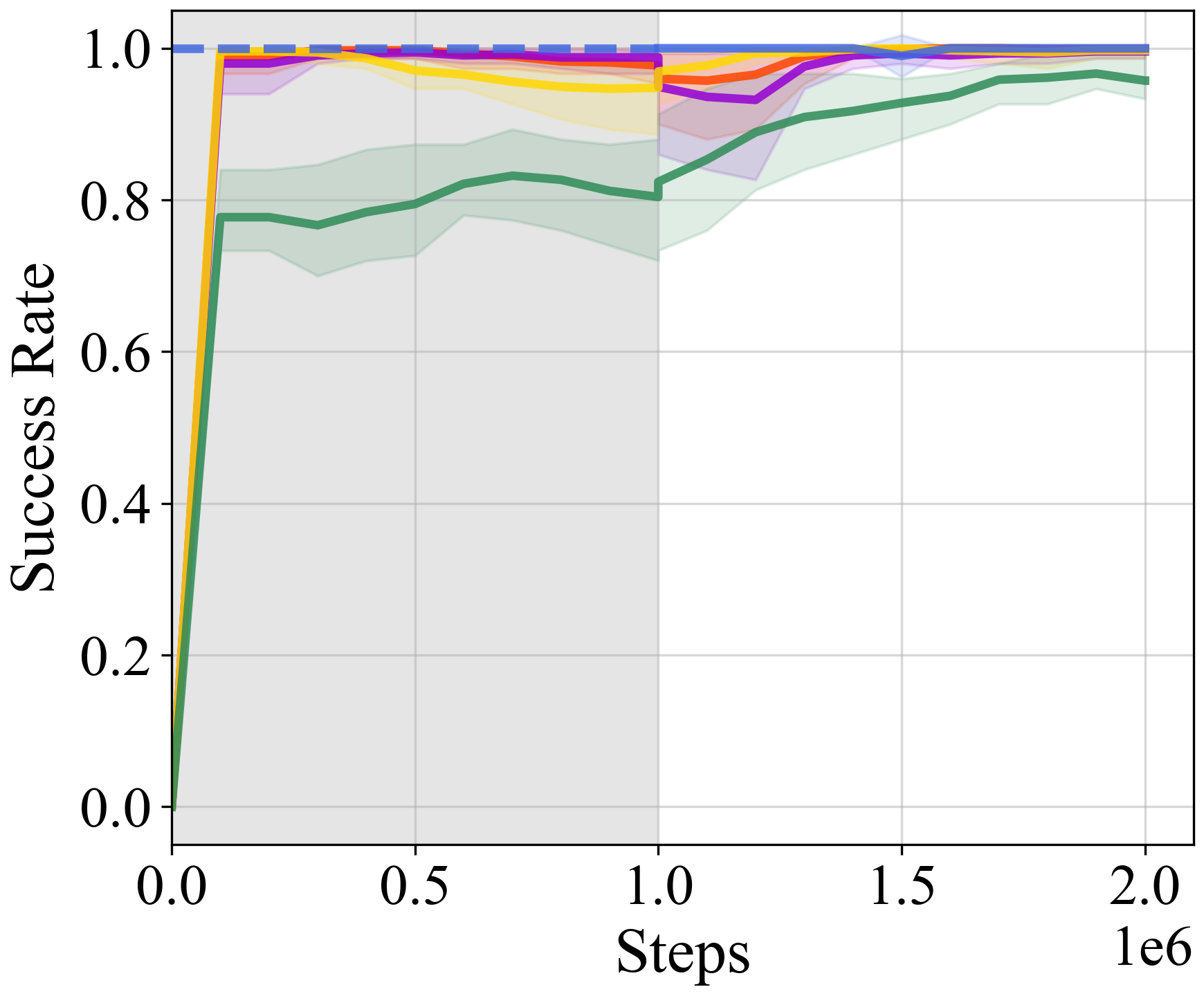}%
  }
  \subfloat[\centering Robomimic-Can]{%
    \includegraphics[width=0.33\textwidth]{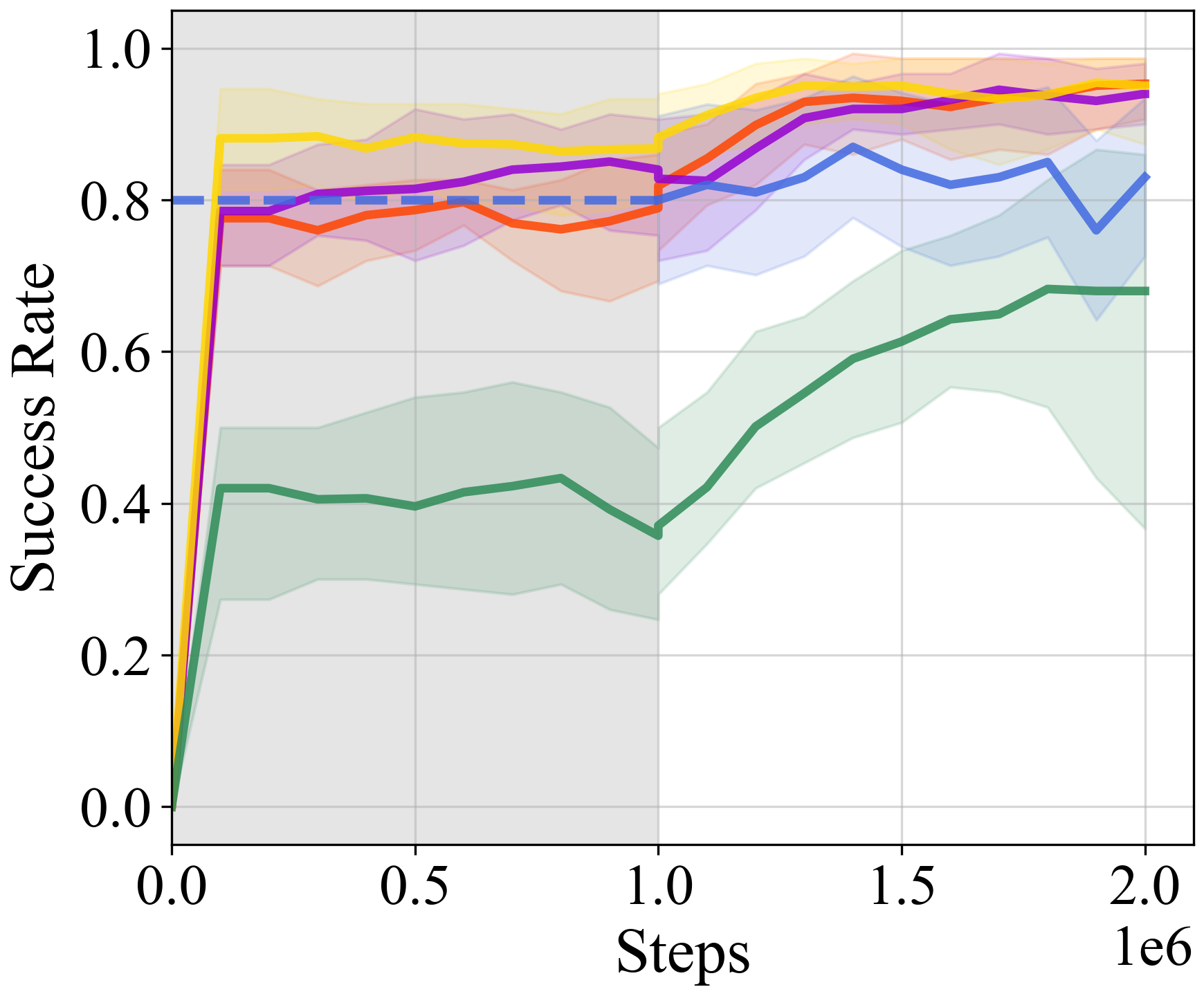}%
  }
  \subfloat[\centering Robomimic-Square]{%
    \includegraphics[width=0.33\textwidth]{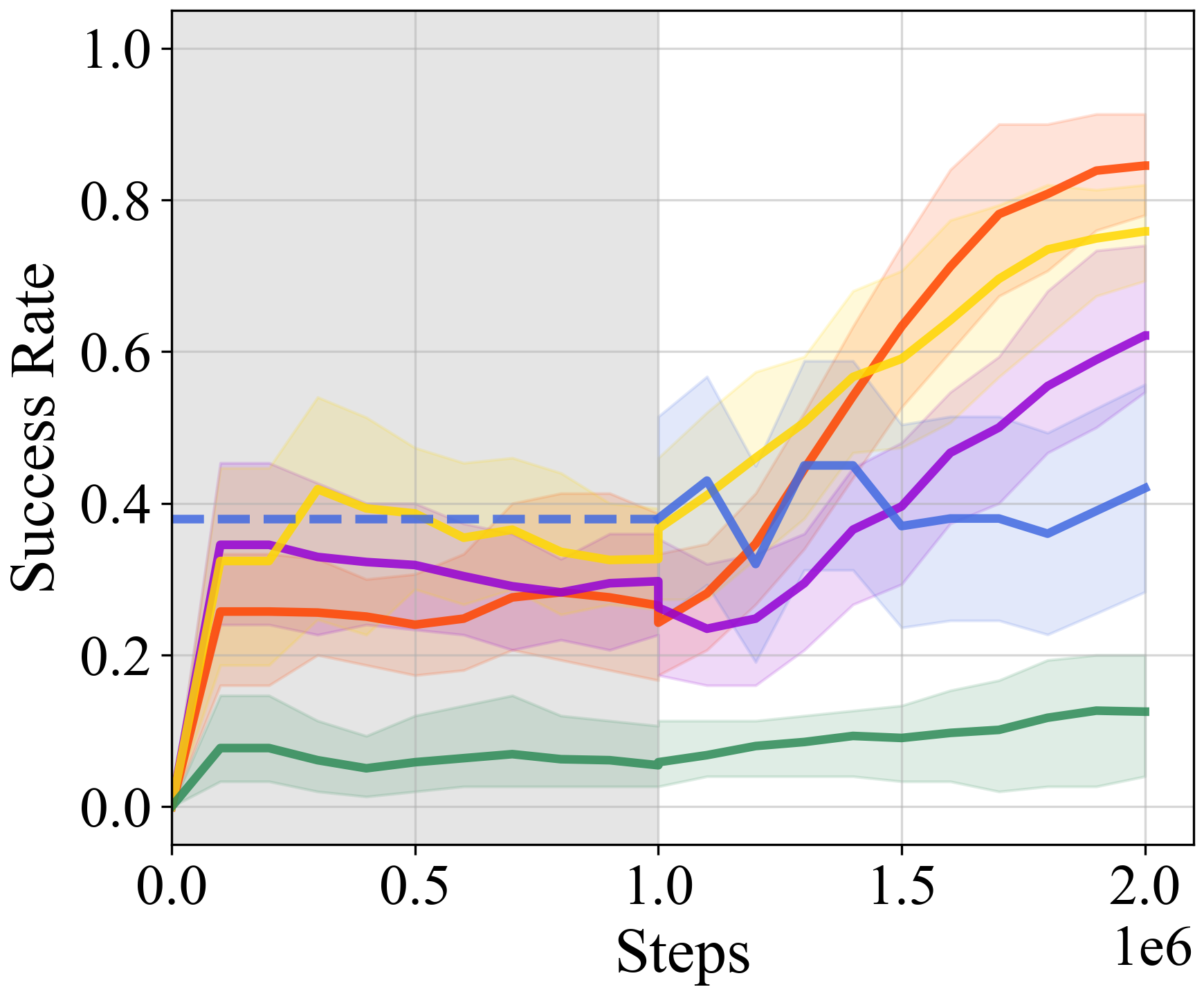}%
  }
  \\
  \vspace{-1mm}
  \subfloat{
  \includegraphics[width = 0.95\textwidth]{figs/o2o_results/legend.png}
  }
  \vspace{-4mm}
  \caption{Complete offline-to-online training performance in Robomimic.
  }
  \label{fig:Complete_Robomimic} 
  \vspace{-2mm}
\end{figure}

\section{The Algorithm Framework for Fine-tuning Setting}
\label{fine-tuning}

In this section, we provide a detailed description of how our Flow-G and Flow-T frameworks can be adapted to fine-tune an arbitrary pre-trained flow-based policy, denoted $v_\theta^{pre}$. We introduce specific formulations and initialization strategies designed to seamlessly integrate $v_\theta^{pre}$ while preserving its initial performance.

\textbf{Flow-G (Adapter):} We utilize a simplified variant of the gated velocity from Equ. (\ref{eq:gru_flow}). Instead of $v_\theta = g_i \odot \left(\hat{v}_\theta - A_{t_i}\right)$, we define the velocity as $v_\theta = g_i \odot \hat{v}_\theta$, where $g_i={Sig}\,\big(z_\theta(t_i,A_{t_i},s)\big)$ acts as a modulating gate. We set the candidate network $\hat{v}_\theta$ to be the frozen, pre-trained network $v_\theta^{pre}$. The new gate network $z_\theta$ is initialized such that its output bias is large (e.g., 5.0) and weights are zero. At the beginning of fine-tuning, this causes $g_i \approx 1$, making the gate act as an identity mapping ($v_\theta \approx v_\theta^{pre}$). The fine-tuning process then only optimizes the lightweight gate network $z_\theta$.

\textbf{Flow-T (Adapter):} For Flow-T, we adapt the decoder architecture from Equ. (\ref{eq:tf_cross_attn}). We replace the final linear projection $W_o$ with the entire pre-trained network $v_\theta^{pre}$. The reformulated velocity becomes $v_\theta(t_i,A_{t_i},s) = v_\theta^{pre}\Big(\mathrm{LN}(\Phi_{A_i}^{(L)})\Big)$. The newly introduced cross-attention and feed-forward blocks within the Transformer stack are zero-initialized. Due to the residual connections, the output $\mathrm{LN}(\Phi_{A_i}^{(L)})$ initially passes through the unchanged action-time embedding, effectively preserving the original input structure expected by $v_\theta^{pre}$. This ensures $v_\theta$ behaves identically to $v_\theta^{pre}$ at the start of fine-tuning.

In most fine-tuning scenarios, the critic $Q_\psi$ is initialized from scratch. To allow the critic to learn meaningful values before the policy is updated with the RL objective, and to let the adapter networks settle, we introduce a "warm-up" phase. During this phase, the critic is trained normally via Equ. (\ref{critic_main}), but the actor is trained only with a behavioral cloning loss:

\begin{equation}
\label{warm_actor}
L_{\text{actor}}^{warm}(\theta)
= \beta\, \|a_h^\theta - a_h\|_2^2,
\quad (s_h,a_h)\sim \mathcal{B}, \quad
a_h^\theta=\tanh(A_{t_K}^\theta) \sim \pi_\theta(\cdot\mid s_h).
\end{equation}

\begin{figure}[t]
    \centering 

    \subfloat[\captionsetup{justification=centering}Cube-Double-Task1]{%
        \includegraphics[width=0.19\textwidth]{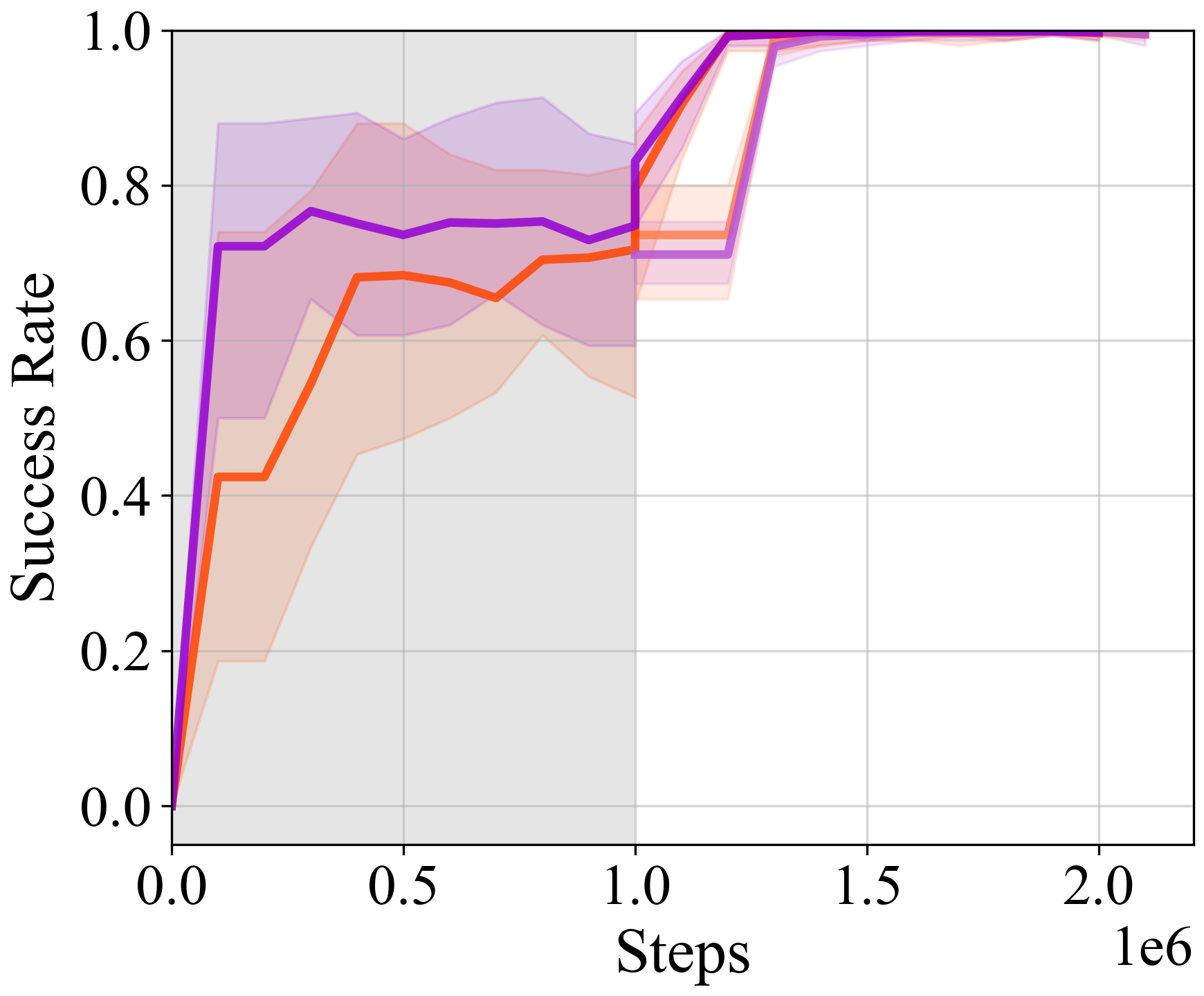}%
        \label{fig:double-task1}%
    }
    \hfill
    \subfloat[\captionsetup{justification=centering}Cube-Double-Task2]{%
        \includegraphics[width=0.19\textwidth]{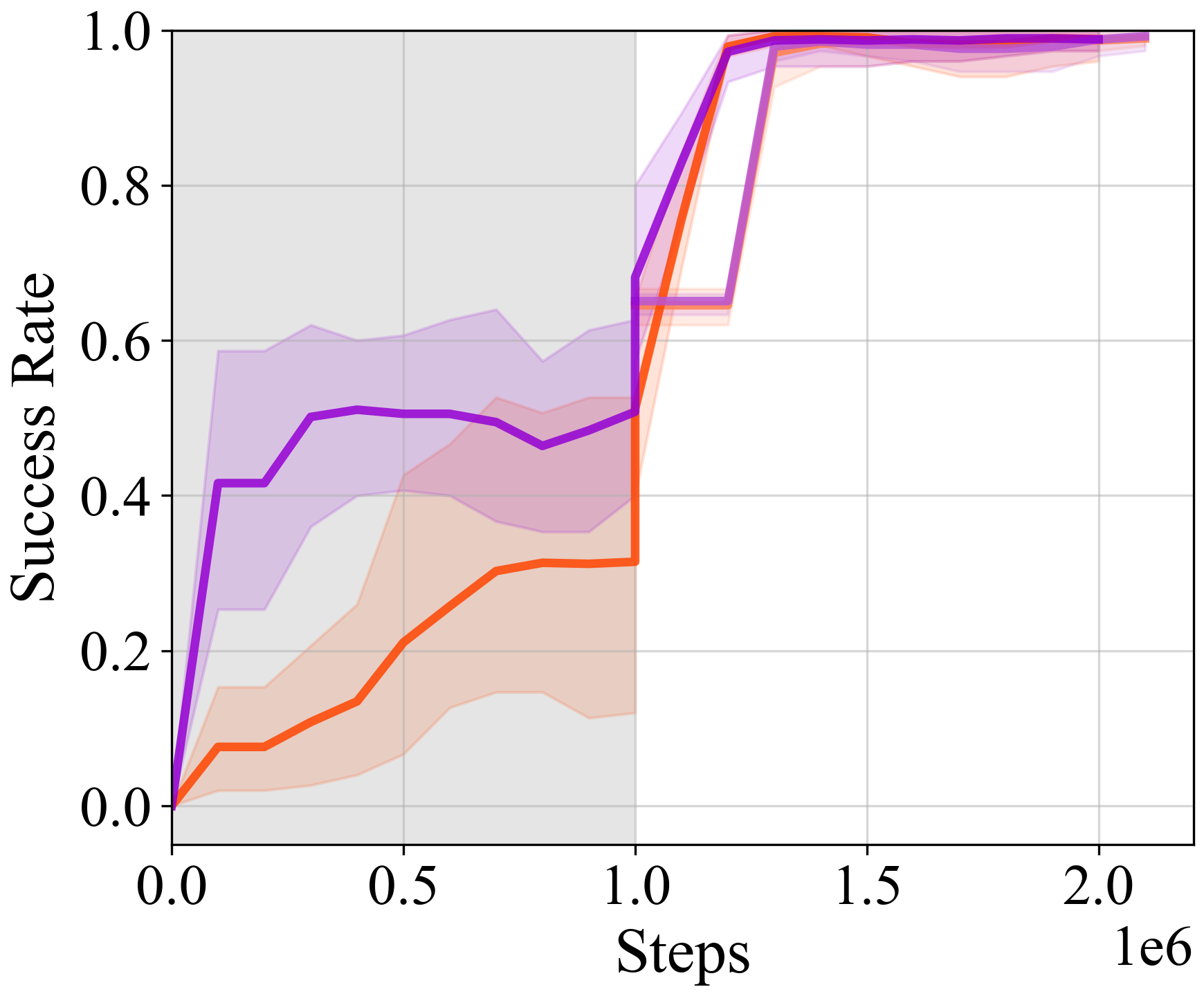}%
        \label{fig:double-task2}%
    }
    \hfill
    \subfloat[\captionsetup{justification=centering}Cube-Double-Task3]{%
        \includegraphics[width=0.19\textwidth]{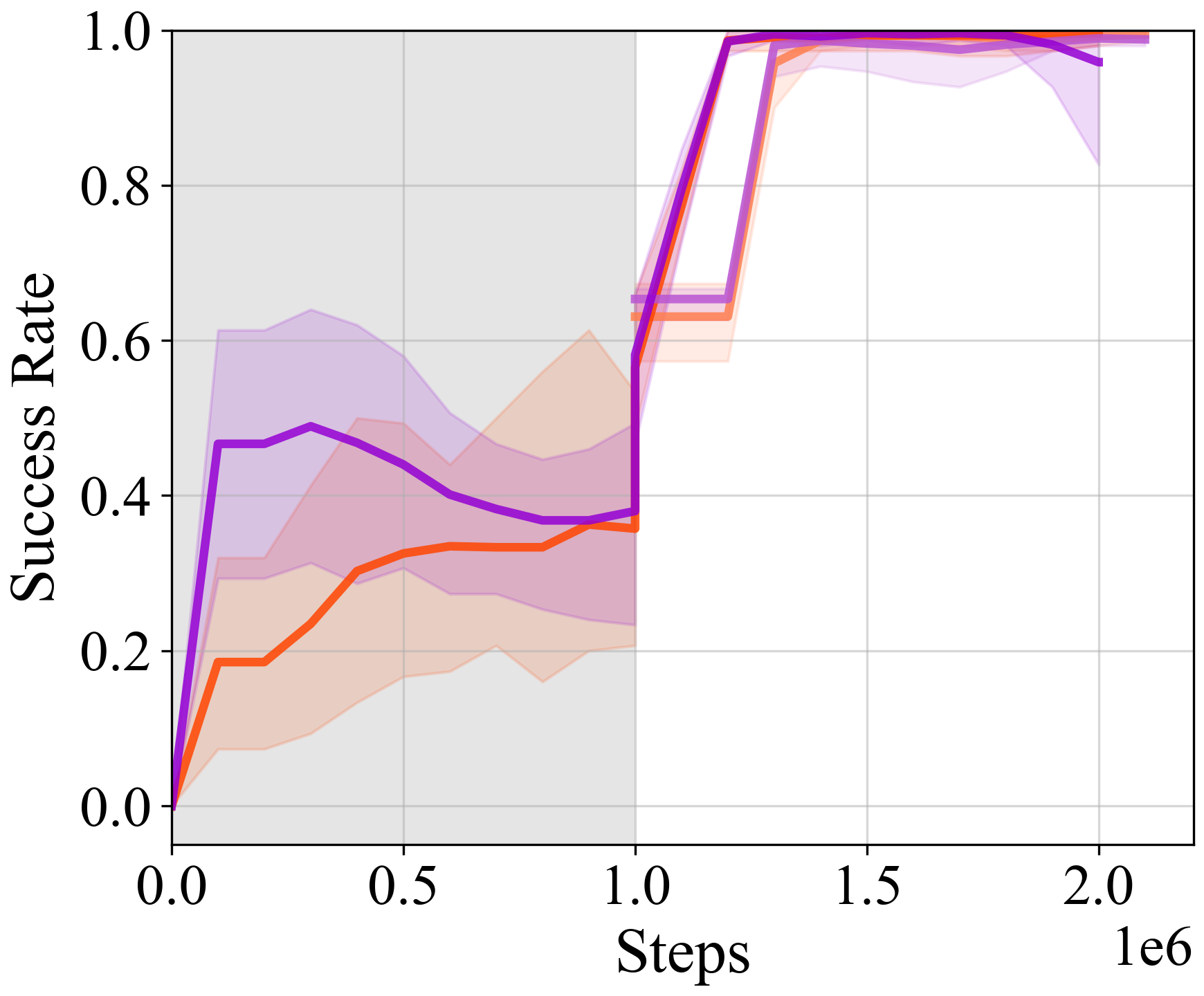}%
        \label{fig:double-task3}%
    }
    \hfill
    \subfloat[\captionsetup{justification=centering}Cube-Double-Task4]{%
        \includegraphics[width=0.19\textwidth]{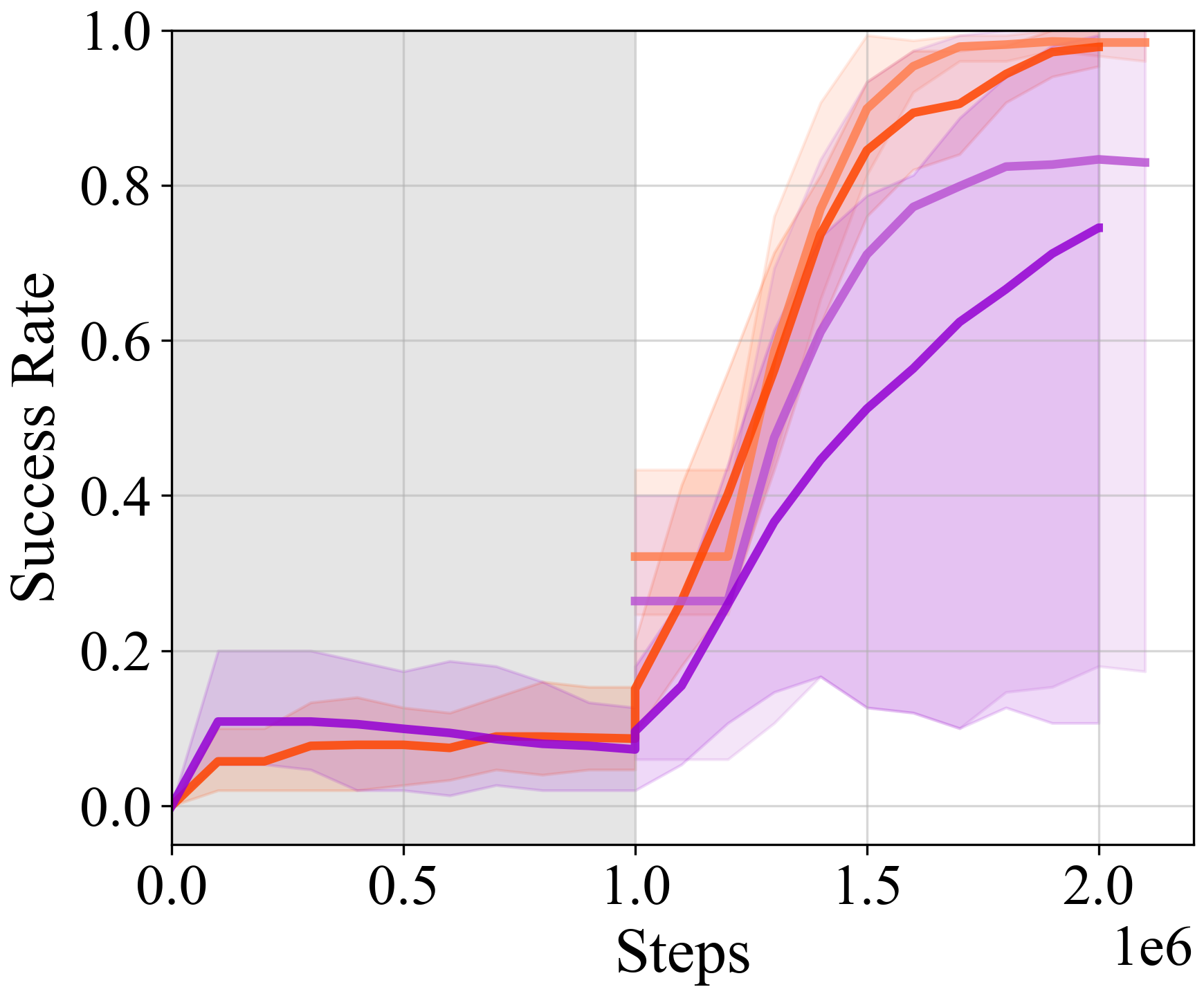}%
        \label{fig:double-task4}%
    }
    \hfill
    \subfloat[\captionsetup{justification=centering}Cube-Double-Task5]{%
        \includegraphics[width=0.19\textwidth]{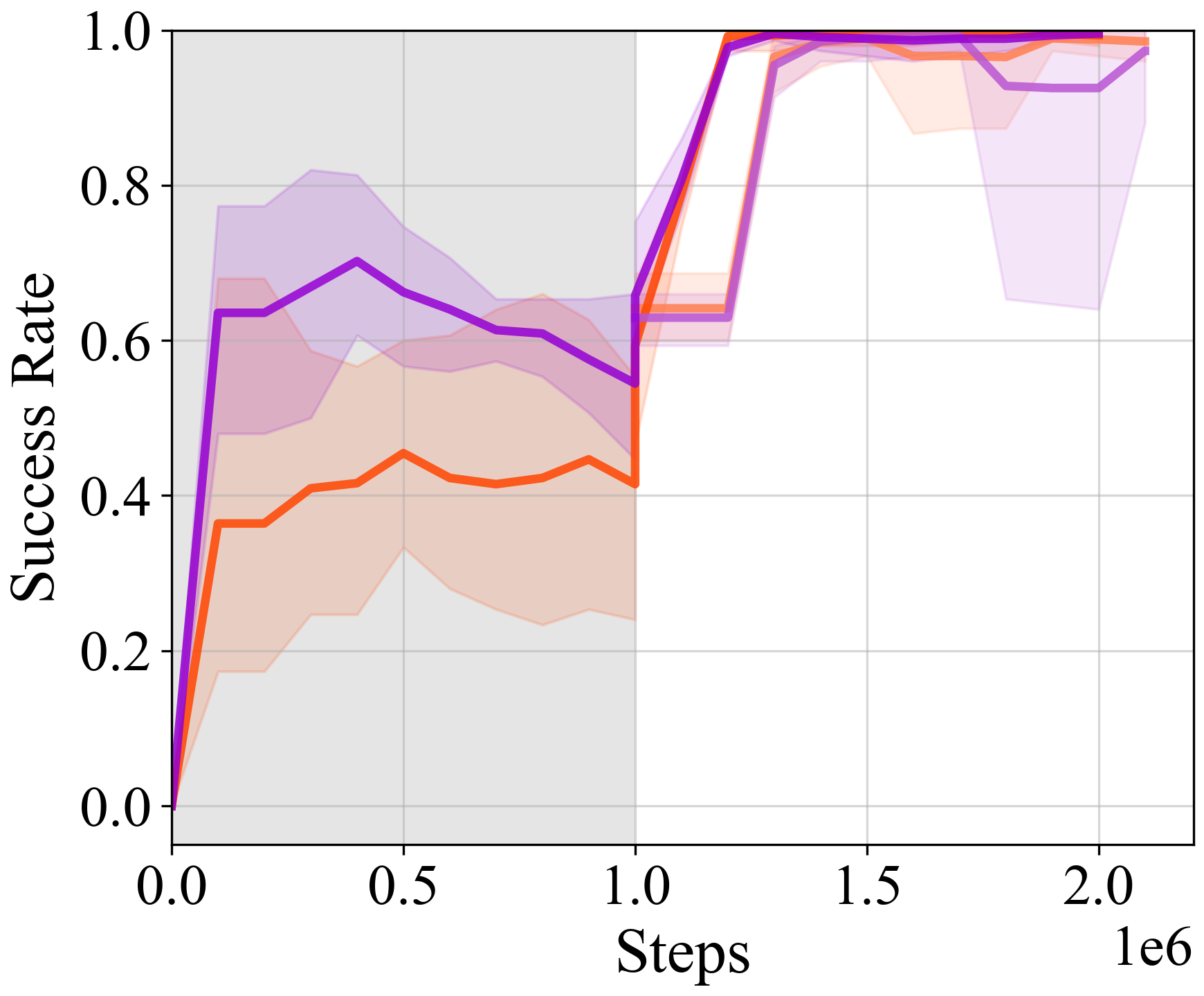}%
        \label{fig:double-task5}%
    }
    \\
    \subfloat[\captionsetup{justification=centering}Cube-Triple-Task1]{%
        \includegraphics[width=0.19\textwidth]{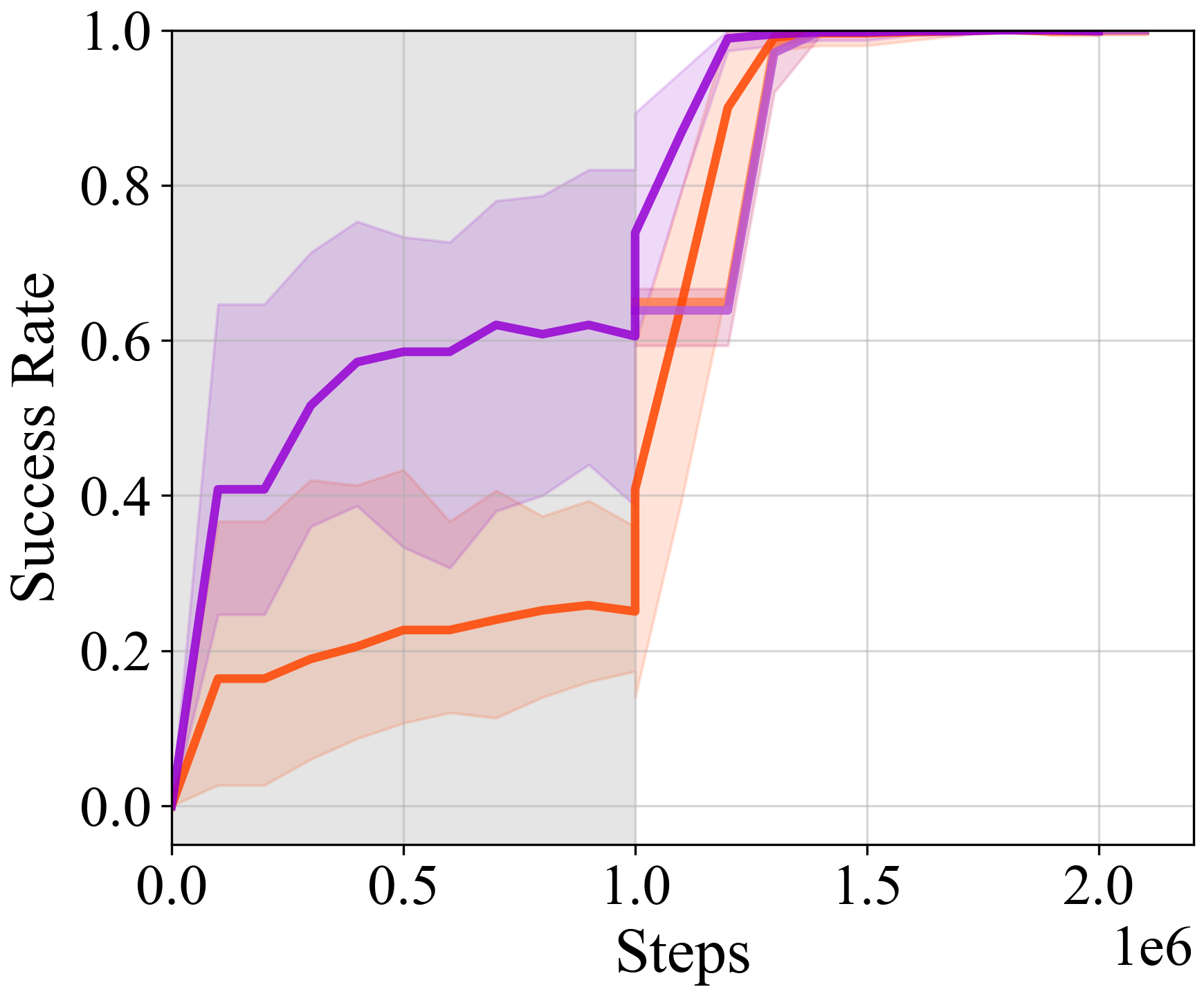}%
        \label{fig:triple-task1}%
    }
    \hfill
    \subfloat[\captionsetup{justification=centering}Cube-Triple-Task2]{%
        \includegraphics[width=0.19\textwidth]{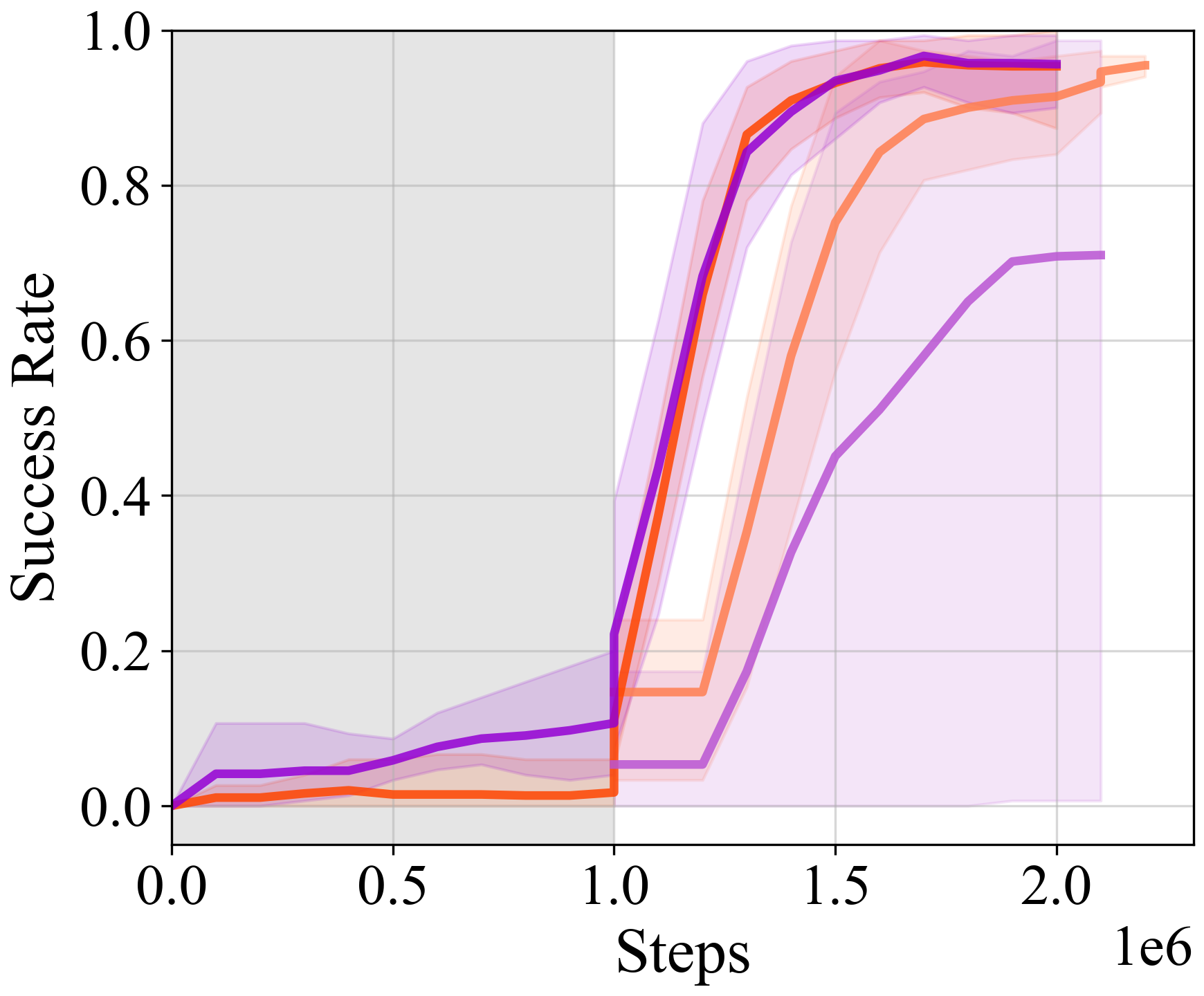}%
        \label{fig:triple-task2}%
    }
    \hfill
    \subfloat[\captionsetup{justification=centering}Cube-Triple-Task3]{%
        \includegraphics[width=0.19\textwidth]{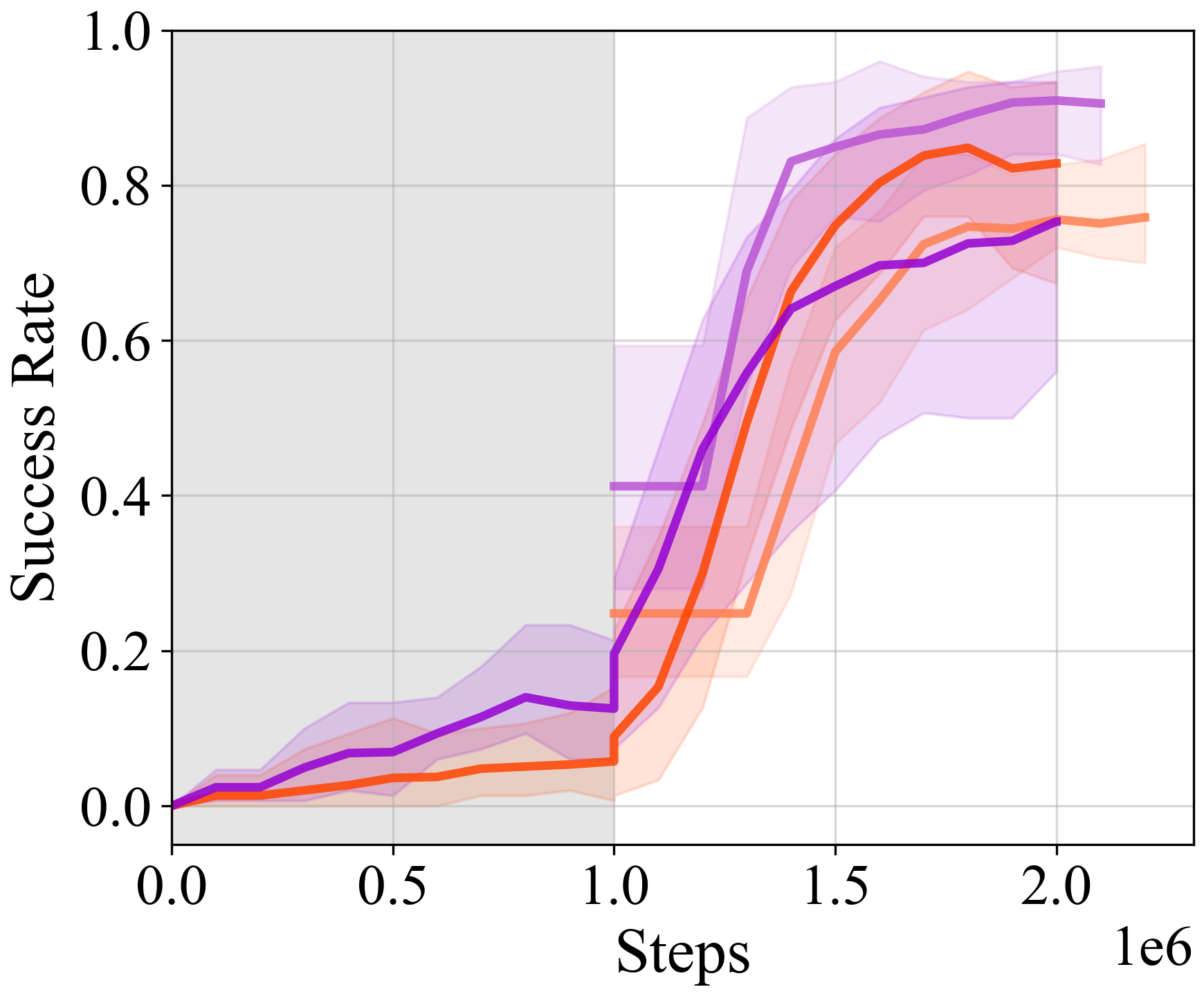}%
        \label{fig:triple-task3}%
    }
    \hfill
    \subfloat[\captionsetup{justification=centering}Cube-Triple-Task4]{%
        \includegraphics[width=0.19\textwidth]{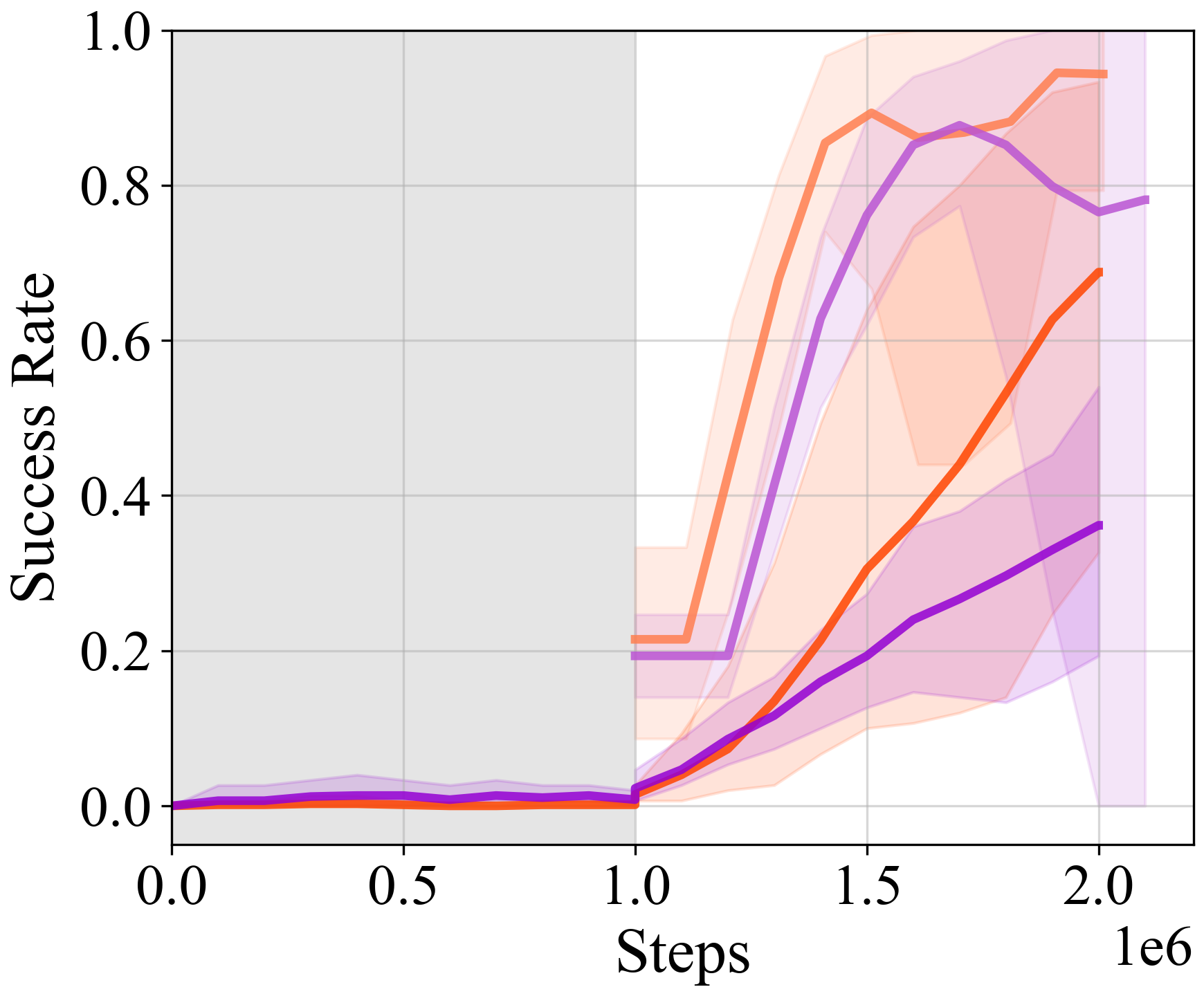}%
        \label{fig:triple-task4}%
    }
    \hfill
    \subfloat[\captionsetup{justification=centering}Cube-Triple-Task5]{%
        \includegraphics[width=0.19\textwidth]{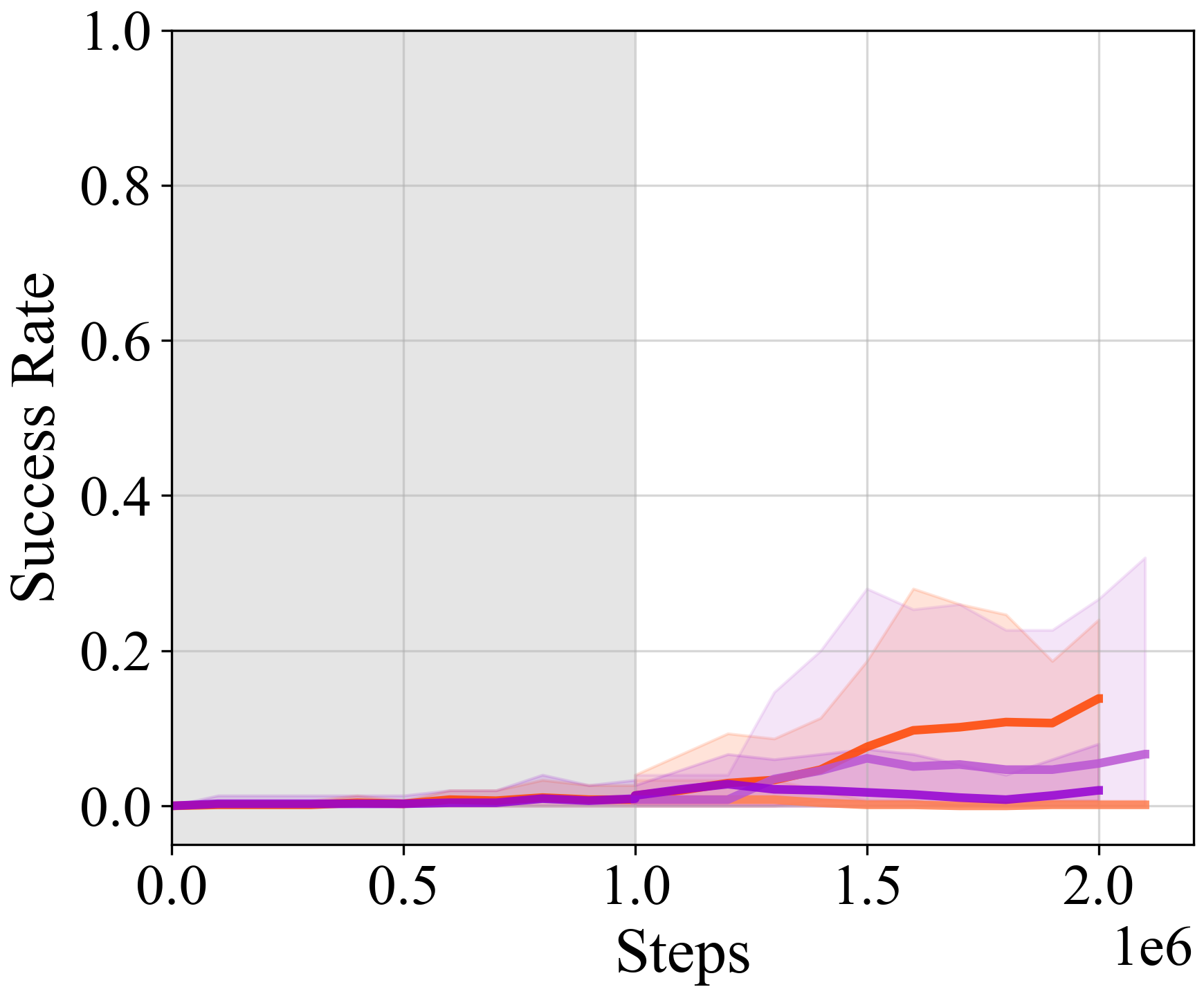}%
        \label{fig:triple-task5}%
    }
    \\
    \vspace{-1mm}
    \subfloat{%
        \includegraphics[width=0.95\textwidth]{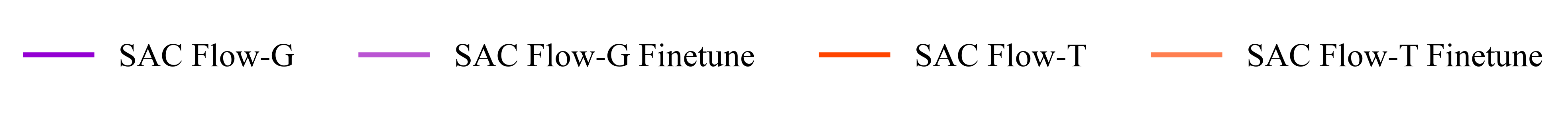}%
    }
    \vspace{-3mm}
    \caption{Complete fine-tuning performance on OGBench. All methods are pre-trained for 1M steps using flow matching with a standard policy. The pre-trained policy is then adapted using our Flow-G or Flow-T formulation and fine-tuned, starting with 0.1M warm-up steps followed by 1M online interaction steps (see Algo. \ref{alg:fine-tune}).}
    \label{fig:Complete_finetune_OGBench} 
    \vspace{-2mm}
\end{figure}

\begin{figure}[htbp]
  \centering 

  \subfloat[\centering Robomimic-Lift]{%
    \includegraphics[width=0.33\textwidth]{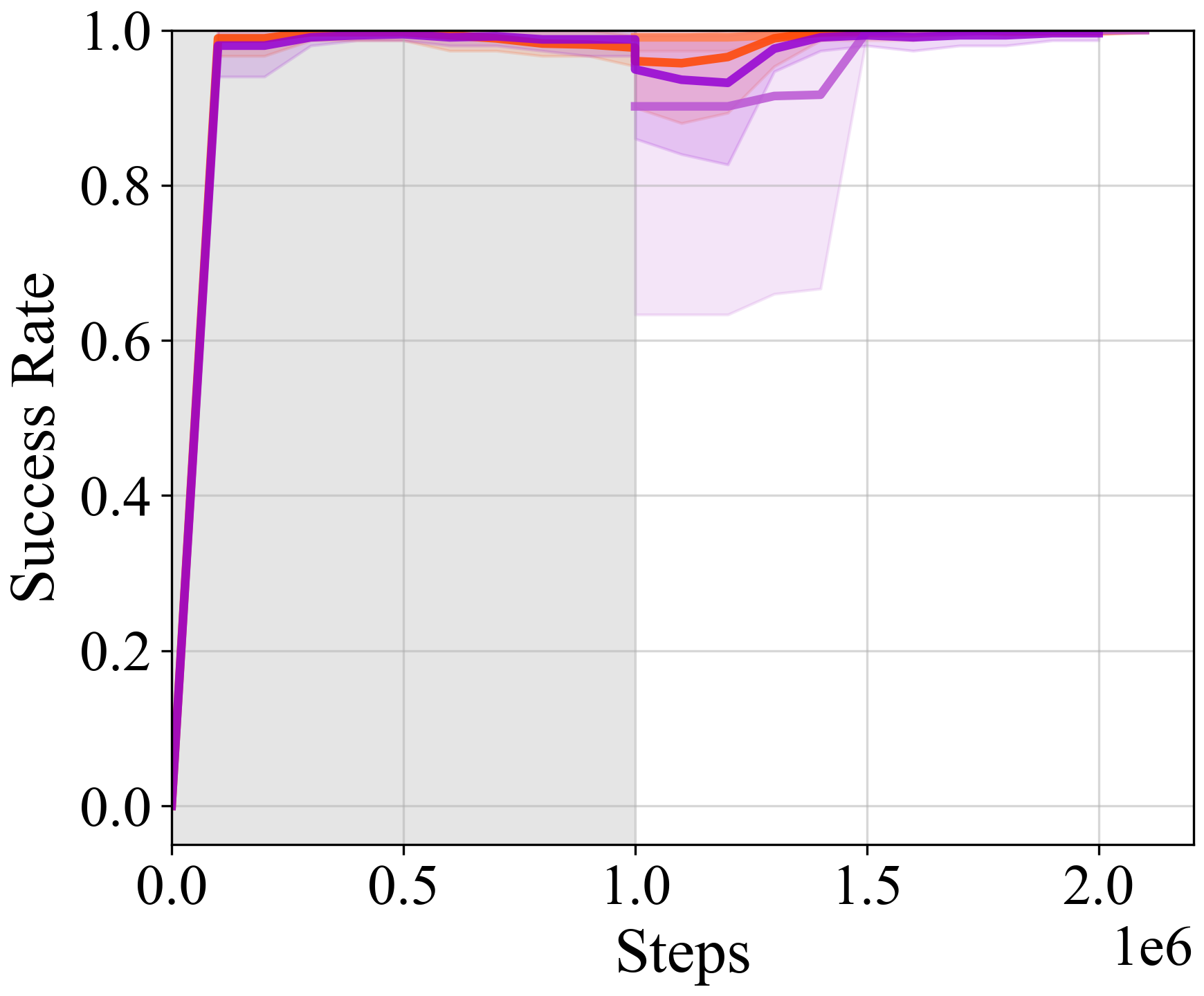}%
  }
  \subfloat[\centering Robomimic-Can]{%
    \includegraphics[width=0.33\textwidth]{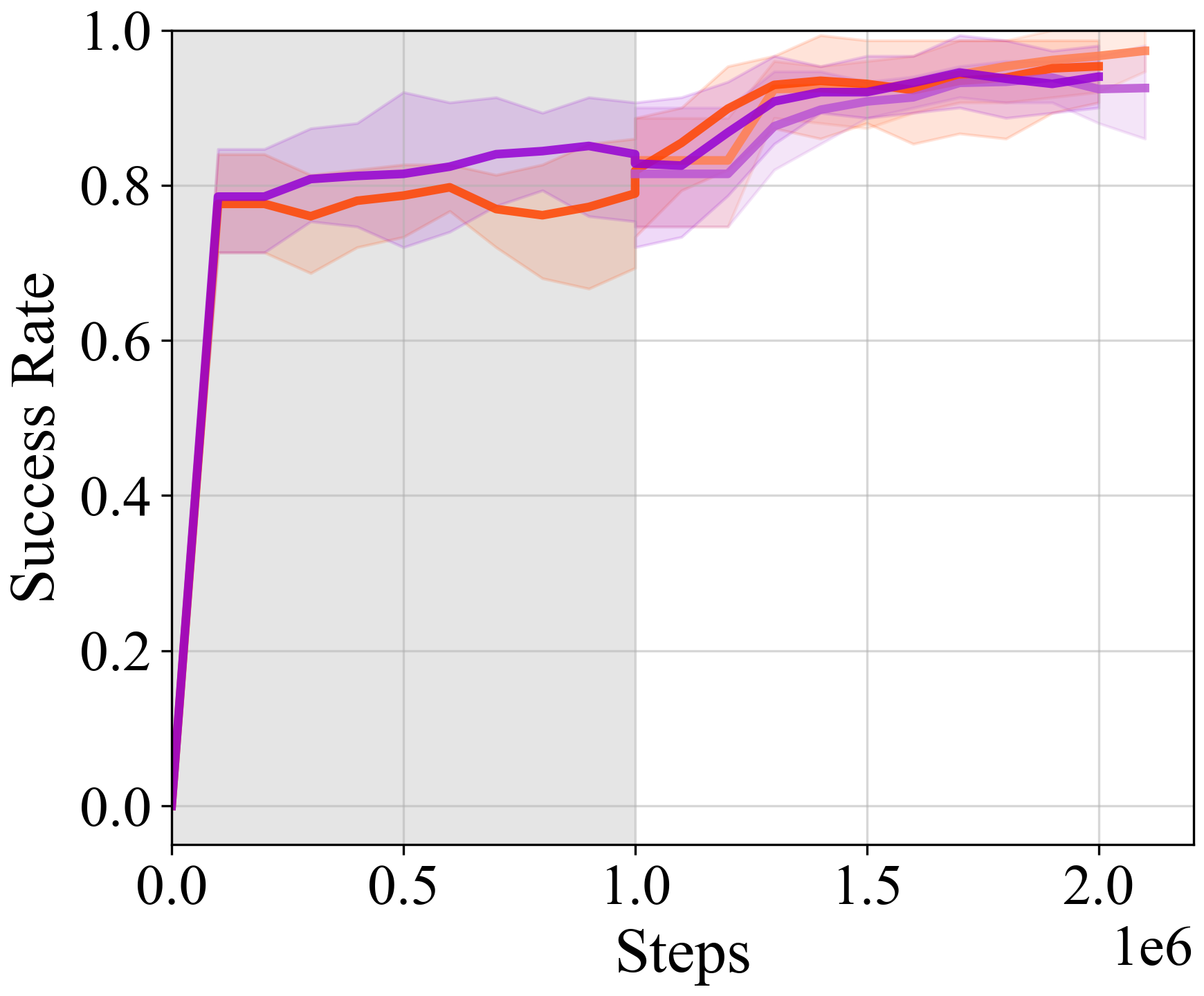}%
  }
  \subfloat[\centering Robomimic-Square]{%
    \includegraphics[width=0.33\textwidth]{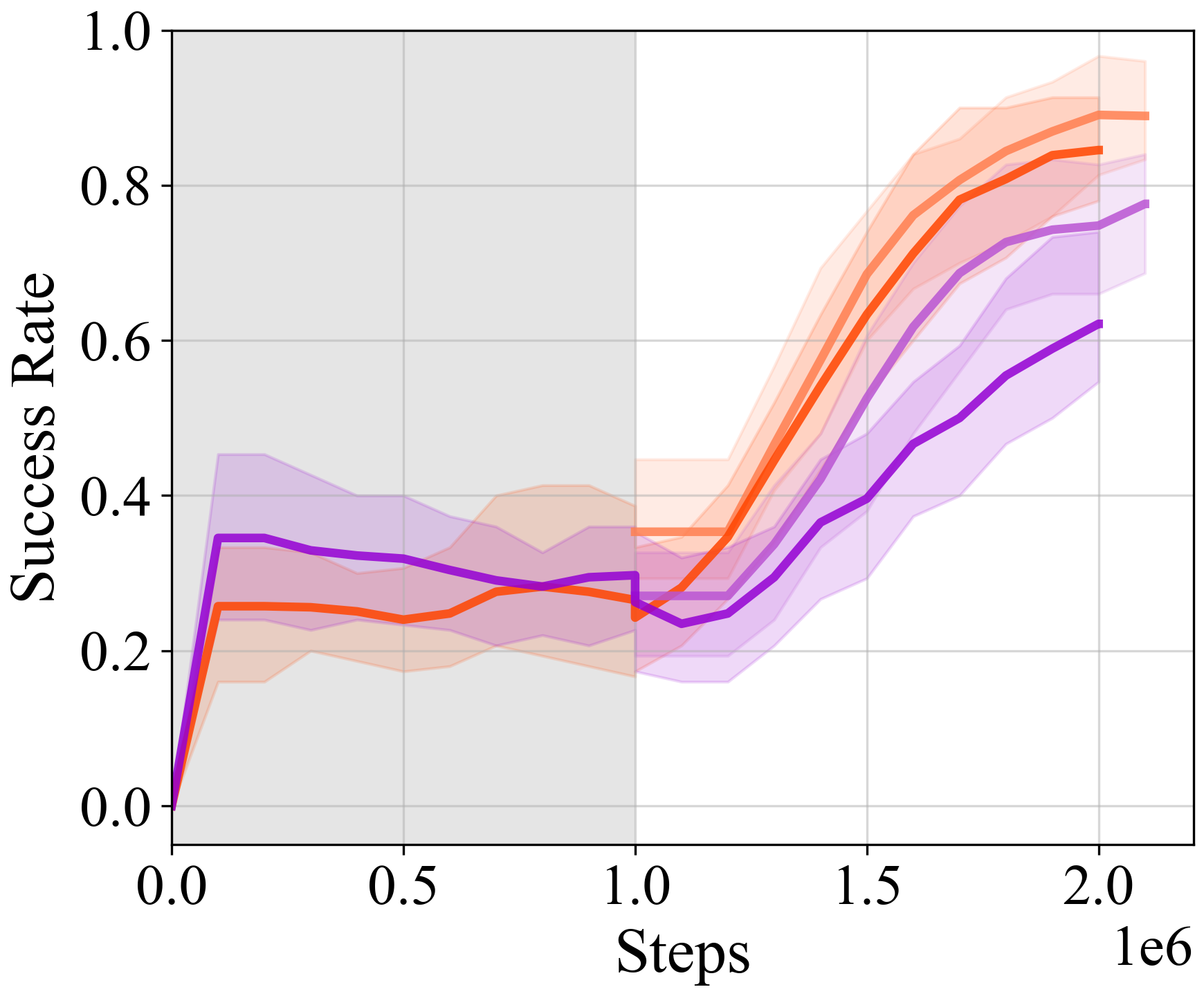}%
  }
  \\
  \vspace{-1mm}
  \subfloat{
  \includegraphics[width = 0.95\textwidth]{figs/finetun/legend.png}
  }
  \vspace{-4mm}
  \caption{Complete fine-tuning performance on Robomimic. The training protocol is identical to the OGBench experiments (Fig. \ref{fig:Complete_finetune_OGBench}): 1M steps of flow matching pre-training, followed by 0.1M warm-up steps and 1M online fine-tuning steps.
  }
  \label{fig:Complete_finetune_Robomimic} 
  \vspace{-2mm}
\end{figure}

Using these specific adapter formulations, we introduce the fine-tuning algorithm in Algo. \ref{alg:fine-tune}. The reformulated policy is first trained for $L_{\text{warmup}}$ steps, followed by $L_{\text{online}}$ steps of full online training. The experimental results are shown in Fig. \ref{fig:Complete_finetune_OGBench} and Fig. \ref{fig:Complete_finetune_Robomimic}. The fine-tuning methods achieve performance comparable to the offline-to-online methods. It is surprising that the fine-tuning methods achieve better performance in Cube-Triple-Task3, Task4, and Robomimic-Square. This suggests that a randomly initialized critic might be easier to optimize during the online phase than one pre-trained via offline RL.

\begin{algorithm}
\caption{SAC Flow (fine-tuning)}
\label{alg:fine-tune}
\begin{algorithmic}[1]
\STATE Initialize critic $Q_\psi$, target $Q_{\bar\psi}$. Set $\mathcal{B} \leftarrow \mathcal{D}_{\text{expert}}$.
\STATE Load pre-trained $v_\theta^{pre}$ and construct $\pi_\theta$ using the Flow-G or Flow-T adapter formulation.
\FOR{$\ell=1$ to $L_{\text{warmup}}$}
\STATE Interact with the environment using $\pi_\theta$; append to $\mathcal{B}$.
\STATE Sample $\{(s_h,a_h,r_h,s_{h+1})\}_{h=1}^N \sim \mathcal{B}$.
\STATE Actor: minimize Equ. (\ref{warm_actor}) with $a_h^\theta$ from the noisy rollout.
\STATE Critic: minimize Equ. (\ref{critic_main}); update the target network.
\ENDFOR
\FOR{$\ell=L_{\text{warmup}}+1$ to $L_{\text{warmup}} + L_{\text{online}}$}
\STATE Interact with the environment using $\pi_\theta$; append to $\mathcal{B}$.
\STATE Sample $\{(s_h,a_h,r_h,s_{h+1})\}_{h=1}^N \sim \mathcal{B}$.
\STATE Actor: minimize Equ. (\ref{actor_o2o}) with $a_h^\theta$ from the noisy rollout.
\STATE Critic: minimize Equ. (\ref{critic_main}); update the target network.
\ENDFOR
\end{algorithmic}
\end{algorithm}

\end{document}